\documentclass[twoside,11pt]{article}
\pdfoutput=1
\usepackage{jmlr2e}

\usepackage{amssymb}
\usepackage{graphicx}
\usepackage{amsmath}
\usepackage{multirow}
\usepackage{multicol}
\usepackage{algorithmic}
\usepackage{algorithm}

\def\ga{\gamma_{A}}
\def\gi{\gamma_{I}}
\def\Bu{\mbox{\boldmath $1$}}
\def\Bk{\mbox{\boldmath $k$}}
\def\Bw{\mbox{\boldmath $w$}}
\def\Bf{\mbox{\boldmath $f$}}
\def\Br{\mbox{\boldmath $r$}}
\def\Bd{\mbox{\boldmath $d$}}
\def\Bx{\mbox{\boldmath $x$}}
\def\Bz{\mbox{\boldmath $z$}}

\def\Ba{\mbox{\boldmath $\alpha$}}
\def\Bb{\mbox{\boldmath $\beta$}}
\def\Bv{\mbox{\boldmath $\varsigma$}}
\def\Bp{\mbox{\boldmath $\xi$}}
\def\Bc{\mbox{\boldmath $c$}}
\def\By{\mbox{\boldmath $y$}}
\def\Bc{\mbox{\boldmath $c$}}

\def\Bw{\mbox{\boldmath $w$}}

\def\Bo{\mbox{\boldmath $0$}}

\newcommand{\subjectto}{\operatornamewithlimits{subject\ to:\ }}

\newcommand{\argmin}{\operatornamewithlimits{arg\,min}}

\def\RealSet{\mbox{$I\! \! R$}}

\def\equationname{Eq.}
\def\sectionname{Section}

\firstpageno{1}

\begin{document}

\title{Laplacian Support Vector Machines\\Trained in the Primal}

\author{\name Stefano Melacci \email mela@dii.unisi.it \\
       \addr Department of Information Engineering\\
       University of Siena\\
       53100, Siena, ITALY
       \AND
       \name Mikhail Belkin \email mbelkin@cse.ohio-state.edu \\
       \addr Department of Computer Science and Engineering\\
       Ohio State University\\
       Columbus, OH 43210, USA}

\editor{}

\maketitle

\begin{abstract}
In the last few years, due to the growing ubiquity of unlabeled data, much effort has been spent by the machine 
learning community to develop better understanding and improve the quality of classifiers exploiting unlabeled data. 
Following the manifold regularization approach, Laplacian Support Vector Machines (LapSVMs) have shown the state of 
the art performance in semi--supervised classification. 
In this paper we present two strategies to solve the \textit{primal} LapSVM problem, in order to overcome some issues 
of the original \textit{dual}  
formulation. Whereas training a LapSVM in the dual requires two steps, using the primal form allows us to collapse 
training to a single step. Moreover, 
the computational complexity of the training algorithm is reduced from $O(n^3)$ to $O(n^2)$ using preconditioned 
conjugate gradient, where $n$ is the 
combined number of labeled and unlabeled examples. We speed up training by using an early stopping strategy based on 
the prediction on unlabeled data or, if available, on labeled validation examples. This allows the algorithm to quickly compute 
approximate solutions 
with roughly the same classification accuracy as the 
optimal ones, considerably reducing the training time. 
Due to its simplicity, training LapSVM in the primal can be the starting point for  additional enhancements of the 
original LapSVM formulation, such as those for dealing with large datasets.
We present an extensive experimental evaluation on real world data showing the benefits of the proposed approach. 
\end{abstract}

\begin{keywords}
  Laplacian Support Vector Machines, Manifold Regularization, Semi--Supervised Learning, Classification, Optimization.
\end{keywords}

\section{Introduction}
\label{sec:intro}
In semi-supervised learning one estimates a target classification/regression function from a few labeled examples together with a large collection of unlabeled data.
In the last few years  there has been a growing interest in the semi--supervised learning in the scientific community. 
Many algorithms for exploiting unlabeled data in order to enhance the quality of classifiers have been recently proposed, see, e.g., \citep{chapelle2006ssl} and \citep{zhu}. The general principle underlying semi-supervised learning  is that the marginal distribution, which can be  estimated from data alone,  may suggest a suitable way to adjust the target function. 
The two commons assumption on such distribution that, 
explicitly or implicitly, are made by many of semi--supervised learning algorithms are the \textit{cluster assumption} \citep{chapelle2003cks} and the \textit{manifold assumption} \citep{lapsvm}. The cluster assumption states that two points are likely to have the same class label 
if they can be connected by a curve  through a high density region. Consequently, the separation boundary between classes should lie in the lower density region of the space. For example, this intuition underlies the Transductive Support Vector Machines~\citep{vapnik2000nature}  and in its different implementations, such as TSVM in \citep{joachims1999tit} or S$^3$VM \citep{demiriz2000oas,chapelle2008ots}. The manifold assumption  states that the marginal probability distribution underlying the data is supported on or near a low--dimensional manifold, and that the target function should change smoothly along the tangent direction. Many graph based methods have been proposed in this direction, but the most of them only perform transductive inference \citep{joachims2003tlv,belkin2003ums,zhu2003}, that is classify the unlabeled data given in training. Laplacian Vector Machines (LapSVM) \citep{lapsvm}   provide a natural out--of--sample extension, so that they can classify data that becomes available after the training process, without having to retrain the classifier or resort to various heuristics.

In this paper, we focus on the LapSVM algorithm, that has shown to achieve the state of the art performances in semi--supervised classification. The original approach used to train LapSVM in \cite{lapsvm} is based on the dual formulation of the problem, in a traditional SVM--like fashion. This dual problem is defined only on a number of dual variables equal to $l$, the number of labeled points, and the the relationship between the $l$ variables and the final $n$ coefficients is given by a linear system of $n$ equations and variables, where $n$ is the total number of training points, both labeled and unlabeled. The overall cost of this ``two step'' process is $O(n^3)$.

Motivated by the recent interest in solving the SVM problem in the primal \citep{chap,joachims2006training,pegasos}, we present a way to solve the primal LapSVM problem that can significantly reduce training times and overcome some issues of the original training algorithm. Specifically, the contributions of this paper are the following:
\begin{enumerate}
\item We propose two methods for solving the LapSVM problem in the primal form (not limited to the linear case), following the ideas presented in \citep{chap} for SVMs. Our Matlab library can be downloaded from \texttt{http://www.dii.unisi.it/\~{\hspace{3mm}melacci}/lapsvmp/}. The solution can now be compactly computed in a ``single step'' on the whole variable set. We show how to solve the problem by Newton's method, comparing it with the supervised case. From this comparison it turns out that the real advantages of the Newton's method for the SVM problem are lost in LapSVM due to the intrisic norm regularizer, and the complexity of this solution is still  $O(n^3)$, same as in the original dual formulation. On the other hand, preconditioned conjugate gradient can be directly applied. Preconditioning by the kernel matrix come at no additional cost, and convergence can be achieved  with only a small number of $O(n^2)$ iterations. Complexity can be further reduced if the kernel matrix is sparse, increasing the scalability of the algorithm.  
\item An approximate solution of the dual form and the resulting approximation of the target optimal function are not directly related due to the change of variables while switching to the dual problem. Training LapSVMs in the primal overcomes this issue, and it allows us to directly compute approximate solutions by controlling the number conjugate gradient iterations. 
\item An approximation of the target function with roughly the same classification accuracy as the optimal one can be achieved 
with a small number of iterations due to the effects of the intrinsic norm regularizer of LapSVMs on the training process. We investigate those effects, showing that they make common stopping conditions for iterative gradient based algorithms hard to tune, often 
 leading to either  a premature  stopping  of the  algorithm or to the execution of a  large amount of iterations without  improvements to the classification accuracy. We suggest to use a criterion built upon the \textit{output} of the classifier on the available training data for terminating the iteration of   the algorithm. Specifically, the stability of the prediction on the unlabeled data, or the classification accuracy on validation data (if available) can be exploited. A number of experiments on several datasets support these types of criteria, showing  that  accuracy similar to that of the optimal solution can be obtained in  with significatly reduced training time.
\item The primal solution of the LapSVM problem is based on an $L_2$ hinge loss, that establishes a direct connection to the Laplacian Regularized Least Square Classifier (LapRLSC) \citep{lapsvm}. We discuss the similarities between primal LapSVM and LapRLSC and we show that the proposed fast solution can be trivially applied also to LapRLSC. 
\end{enumerate}

The rest of the paper is organized as follows. In \sectionname~\ref{sec:mr} the basic principles behind manifold regularization are resumed. \sectionname~\ref{sec:lapsvm} describes the LapSVM algorithm in its original formulation whereas \sectionname~\ref{sec:trainp} discusses the proposed solutions of the primal form and their details. The quality of an approximate solution and the data based early stopping criterion are the key contents of \sectionname~\ref{sec:approx}. In  \sectionname~\ref{sec:laprlsc} a parallel with the primal solution of LapSVM and the one of LapRLSC is drawn, describing some possible future work. An extensive experimental analysis is presented in \sectionname~\ref{sec:results}, and, finally,  \sectionname~\ref{sec:concl} concludes the paper.

\section{Manifold Regularization}
\label{sec:mr}
First, we introduce some notation that will be used in this \sectionname\ and in the rest of the paper. We take $n=l+u$ to be the number of $m$ dimensional training examples $\Bx_i\in X \subset \RealSet^m$, collected in $\mathcal{S}=\{\Bx_i,i=1,\ldots,n\}$. Examples are ordered so that the first $l$ ones are labeled, with label $y_i\in\{-1,1\}$, and the remaining $u$ points are unlabeled. We put  $\mathcal{S}=\mathcal{L}\cup\mathcal{U}$, where $\mathcal{L}=\{(\Bx_i,y_i),i=1,\ldots,l\}$ is the labeled data set and $\mathcal{U}=\{\Bx_i,i=l+1,\ldots,n\}$ is the unlabeled data set. Labeled examples are generated accordingly to the distribution $P$ on $X \times \RealSet$, whereas unlabeled examples are drawn according to the marginal distribution $P_X$ of $P$. Labels are obtained from the conditional probability distribution $P(y | \Bx)$. 
$L$ is the graph Laplacian associated to $\mathcal{S}$, given by $L=D-W$, where $W$ is the adjacency matrix of the data graph (the entry in position $i,j$ is indicated with $w_{ij}$) and $D$ is the diagonal matrix with the degree of each node (i.e. the element $d_{ii}$ from $D$ is $d_{ii}=\sum_{j=1}^{n}{w_{ij}}$). Laplacian can be expressed in the normalized form, $L=D^{-\frac{1}{2}}LD^{-\frac{1}{2}}$, and iterated to a degree $p$ greater that one. By $K\in\RealSet^{n,n}$ we denote  the Gram matrix associated to the $n$ points of $\mathcal{S}$ and the $i,j$--th entry of such matrix is the evaluation of the kernel function $k(\Bx_i,\Bx_j)$, $k: X \times X \to \RealSet$. The unknown target function that the learning algorithm must estimate is indicated with $f: X \to \RealSet$, where $\Bf$ is the vector of the $n$ values of $f$ on training data, $\Bf=[f(\Bx_i),\Bx_i\in\mathcal{S}]^T$. In a classification problem, the decision function that discriminates between classes is indicated with $y(\Bx)=g(f(\Bx))$, where we overloaded the use of $y$ to denote such function.

Manifold regularization approach \citep{lapsvm} exploits the geometry of the marginal distribution $P_X$.
The support of the  probability distribution of data is assumed to have the geometric structure of a Riemannian manifold $\mathcal{M}$. The labels of two points that are close in the intrinsic geometry of $P_X$ (i.e. with respect to geodesic distances on $\mathcal{M}$) should be the same or similar in sense that  the conditional probability distribution $P(y | \Bx)$ should change little between two such points.
This constraint is enforced in the learning process by an intrinsic regularizer $\| f \|^2_I$ that is empirically estimated from the point cloud of labeled and unlabeled data using the graph Laplacian associated to them, since $\mathcal{M}$ is truly unknown. In particular, choosing exponential weights for the adjacency matrix leads to convergence of the graph Laplacian to the Laplace--Beltrami operator on the manifold \citep{belkin2008ttf}. As a result, we have
\begin{equation}
\| f \|^2_I=\sum_{i=1}^{n}\sum_{j=i}^{n}w_{ij}(f(\Bx_i)-f(\Bx_j))^2=\Bf^T L \Bf.
\label{eq:norm}
\end{equation}
Consider that, in general, several natural choices of $\| \|_I$ exist \citep{lapsvm}.

In the established regularization framework for function learning, given a kernel function $k(\cdot,\cdot)$, its associated Reproducing Kernel Hilber Space (RKHS) $\mathcal{H}_k$ of functions $X\to \RealSet$ with corresponding norm $\| \|_A$, we estimate the target function by minimizing
\begin{equation}
f^*=\argmin_{f\in\mathcal{H}_k}\sum_{i=1}^{l}{V(\Bx_i,y_i,f)+\ga\|f\|_A^2+\gi\| f \|^2_I}
\label{eq:gen}
\end{equation}
where $V$ is some loss function and $\ga$ is the weight of the norm of the function in the RKHS (or \textit{ambient} norm), that enforces a smoothness condition on the possible solutions, and $\gi$ is the weight of the norm of the function in the low dimensional manifold (or \textit{intrinsic} norm), that enforces smoothness along the sampled $\mathcal{M}$. For simplicity, we removed every normalization factor of the weights of each term in the summation. The ambient regularizer makes the problem well--posed, and its presence can be really helpful from a practical point of view when the manifold assumption holds at a lesser degree. 

\begin{figure}[ht]
	\centering
	  \hspace{-2mm}	
		\begin{minipage}{0.305\textwidth}	
	  \centering
		\includegraphics[width=1.0\textwidth]{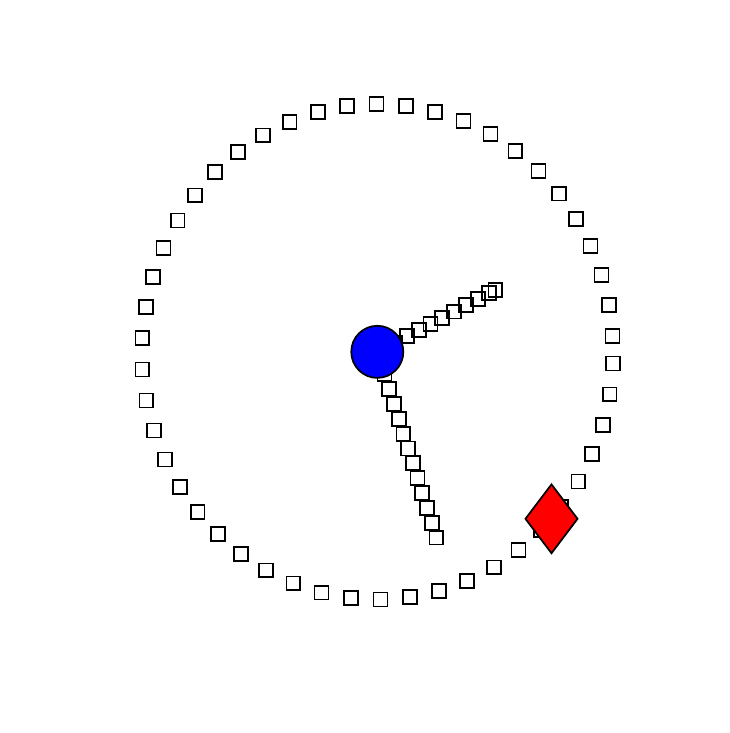}\\(a)
		\end{minipage}
		\hspace{4mm}
		\begin{minipage}{0.305\textwidth}	
	  \centering
		\includegraphics[width=1.0\textwidth]{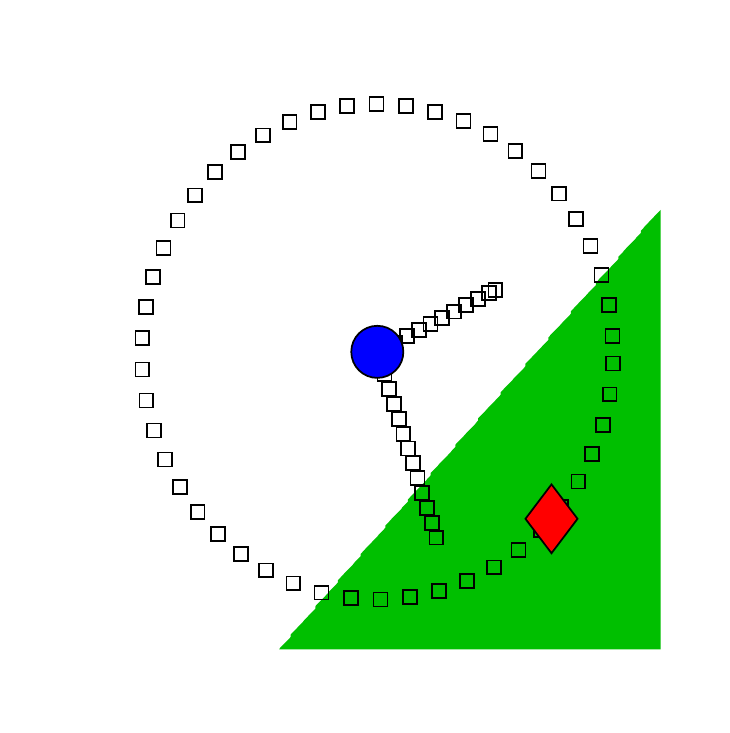}\\(b) 
		\end{minipage}
		\hspace{4mm}	
		\begin{minipage}{0.305\textwidth}	
	  \centering
		\includegraphics[width=1.0\textwidth]{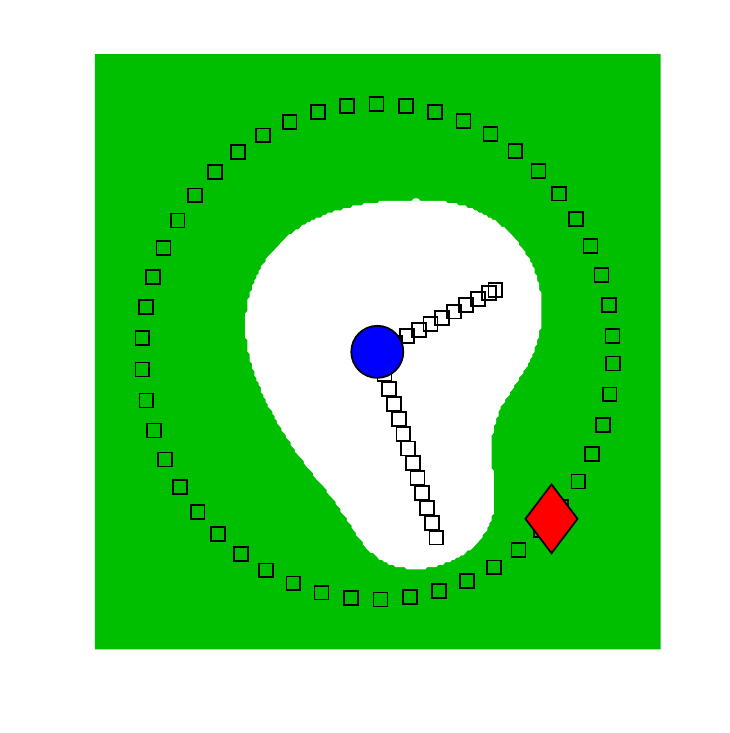}\\(c)
		\end{minipage}\\					
	\caption{(a) The two class ``clock'' dataset. One class is the circular border of the clock, the other one is the hour/minute hands. A large set of unlabeled examples (black squares) and only one labeled example per class (red diamond, blue circle) are selected. - (b) The result of a maximum margin supervised classification - (c) The result of a semi--supervised classification with intrinsic norm from manifold regularization.}	
	\label{fig:2circles}	
\end{figure}

It has been shown in \citet{lapsvm} that $f^*$ admits an expansion in terms of the $n$ points of $\mathcal{S}$,
\begin{equation}
f^*(\Bx)=\sum_{i=1}^{n}\alpha_i^*k(\Bx_i,\Bx).
\label{eq:rt}
\end{equation}
The decision function that discriminates between class $+1$ and $-1$ is $y(\Bx)=sign(f^*(\Bx))$. \figurename~\ref{fig:2circles} shows the effect of the intrinsic regularizer on the ``clock'' toy dataset. The supervised approach defines the classification hyperplane just by considering the two labeled examples, and it does not benefit from unlabeled data (\figurename~\ref{fig:2circles}(b)). With manifold regularization, the classification appears more natural with respect to the geometry of the marginal distribution (\figurename~\ref{fig:2circles}(c)).

The intrinsic norm of \equationname~\ref{eq:norm} actually performs a transduction along the manifold that enforces the values of $f$ in nearby points with respect to geodesic distances on $\mathcal{M}$ to be the ``same''. From a merely practical point of view, the intrinsic regularizer can be excessively strict in some situations. Since the decision function $y(\Bx)$ relies only on the sign of the target function $f(\Bx)$, if $f$ has the same sign on nearby points along $\mathcal{M}$ then the graph transduction is actually complete. Requiring that $f$ assumes exactly the same value on a pair of nearby points could be considered as over constraining the problem. 

This intuition is closely related to the  ideas explored in \citet{vikassphd,coreg,witch}. In particular, in some restricted function spaces the intrinsic regularizer could degenerate to the ambient one as  it is not able to model some underlying geometries of the given data. The Manifold Co-Regularization (MCR) framework \citep{coreg} has been proposed to overcome such issue using multi--view learning. It has been shown that MCR corresponds to adding some extra slack variables in the objective function of \equationname~\ref{eq:gen} to better fit the intrinsic regularizer. The slack variables of MCR could be seen as a way to relax the regularizer. Similarly, \cite{witch} uses a slack based formulation to improve the flexibility of the graph regularizer of their spam detector. This problem has been addressed also by \citet{tsang2007lss}, where the intrinsic regularizer is an $\epsilon$--insensitive loss. We will use these considerations in \sectionname~\ref{sec:approx} to early stop the training algorithm.

\subsection{Laplacian Support Vector Machines}
\label{sec:lapsvm}
LapSVMs follow the principles behind manifold regularization (\equationname~\ref{eq:gen}), where the loss function $V(\Bx,y,f)$ is the linear hinge loss \citep{vapnik2000nature}, or $L_1$ loss. The interesting property of such function is that well classified labeled examples are not penalized by $V(\Bx,y,f)$, independently by the value of $f$. 

In order to train a LapSVM classifier, the following problem must be solved
\begin{equation}
\min_{f\in\mathcal{H}_k}\sum_{i=1}^{l}{\max(1-y_if(\Bx_i),0)+\ga\|f\|_A^2+\gi\| f \|^2_I}.
\label{eq:lapsvm}
\end{equation}
The function $f(\Bx)$ admits the expansion of \equationname~\ref{eq:rt}, where an unregularized bias term $b$ can be added as in many SVM formulations.

The solution of LapSVM problem proposed by \citet{lapsvm} is based on the dual form. By introducing the slack variables $\xi_i$, the unconstrained primal problem can be written as a constrained one:
$$
\begin{array}{cl}
\multicolumn{2}{c}{\min_{\Ba\in \RealSet^n,\Bp\in \RealSet^l}\sum_{i=1}^l{\xi_i}+\ga\Ba^TK\Ba+\gi\Ba^TKLK\Ba}\\
& \\
\subjectto & y_i(\sum_{j=1}^n\alpha_ik(\Bx_i,\Bx_j)+b)\geq 1-\xi_i,\ \ \ i=1,\ldots,l\\
& \\
& \xi_i\geq0,\ \ \ i=1,\ldots,l\\
\end{array}
$$

After the introduction of two sets of $n$ multipliers $\Bb$, $\Bv$, the Lagrangian $L_{g}$ associated to the problem is:
\begin{eqnarray*}
L_{g}(\Ba,\Bp,b,\Bb,\Bv)&=&\sum_{i=1}^l{\xi_i}+\frac{1}{2}\Ba^T(2\ga K+2\gi KLK)\Ba-\\
&&-\sum_{i=1}^l\beta_i(y_i(\sum_{j=1}^n\alpha_ik(\Bx_i,\Bx_j)+b)-1+\xi_i)-\sum_{i=1}^l\varsigma_i\xi_i
\end{eqnarray*}

In order to recover the dual representation we need to set:
\begin{eqnarray*}
\frac{\partial L_{g}}{\partial b}=0 &\Longrightarrow & \sum_{i=1}^l\beta_iy_i=0  \\
\frac{\partial L_{g}}{\partial \xi_i}=0 &\Longrightarrow & 1-\beta_i-\varsigma_i=0\ \ \Longrightarrow \ \  0\leq\beta_i\leq1\
\end{eqnarray*}
where the bounds on $\beta_i$ consider that $\beta_i,\varsigma_i\geq 0$, since they are Lagrange multipliers.
Using the above identities, we can rewrite the Lagrangian as a function of $\Ba$ and $\Bb$ only. Assuming (as stated in \sectionname~\ref{sec:mr}) that the points in $\mathcal{S}$ are ordered such that the first $l$ are labeled and the remaining $u$ are unlabeled, we define with $J_\mathcal{L}\in\RealSet^{l,n}$ the matrix $[I\ 0]$ where $I\in\RealSet^{l,l}$ is the identity matrix and $0\in\RealSet^{l,u}$ is a rectangular matrix with all zeros. Moreover, $Y\in\RealSet^{l,l}$ is a diagonal matrix composed by the labels $y_i,i=1,\ldots,l$. The Lagrangian becomes
\begin{eqnarray*}
L_{g}(\Ba,\Bb)&=&\frac{1}{2}\Ba^T(2\ga K+2\gi KLK)\Ba -\sum_{i=1}^l\beta_i(y_i(\sum_{j=1}^n\alpha_ik(\Bx_i,\Bx_j)+b)-1)=\\
&=&\frac{1}{2}\Ba^T(2\ga K+2\gi KLK)\Ba-\Ba^TKJ^T_\mathcal{L}Y\Bb+\sum_{i=1}^l\beta_i.
\end{eqnarray*}

Setting to zero the derivative with respect to $\Ba$ establishes a direct relationships between the $\Bb$ coefficients and the $\Ba$ ones:
\begin{equation*}
\frac{\partial L_{g}}{\partial \Ba}=0 \ \ \Longrightarrow \ \ (2\ga K+2\gi KLK)\Ba-KJ^T_\mathcal{L}Y\Bb=0
\end{equation*}
\begin{equation}\ \ \ \ \ \ \ \Longrightarrow \ \ \Ba=(2\ga I+2\gi KL)^{-1}J^T_\mathcal{L}Y\Bb
\label{eq:ab}
\end{equation}

After substituting back in the Lagrangian expression, we get the dual problem whose solution leads to the optimal $\Bb^*$:
$$
\begin{array}{cl}
\multicolumn{2}{c}{\max_{\Bb\in\RealSet^l}\sum_{i=1}^l\beta_i-\frac{1}{2}\Bb^TQ\Bb}\\
& \\
\subjectto & \sum_{i=1}^l\beta_iy_i=0\\
& \\
& 0\leq\beta_i\leq 1,\ \ \ i=1,\ldots,l\\
\end{array}
$$
where
\begin{equation}
Q=YJ^T_\mathcal{L}K(2\ga I+2\gi KL)^{-1}J^T_\mathcal{L}Y.
\label{eq:q}
\end{equation}

Training the LapSVM classifier requires to optimize this $l$ variable problem, for example using a standard quadratic SVM solver, and then to solve the linear system of $n$ equations and $n$ variables of \equationname~\ref{eq:ab} in order to get the coefficients $\Ba^*$ that define the target function $f^*$.

The overall complexity of this ``two step'' solution is $O(n^3)$, due to the matrix inversion of \equationname~\ref{eq:ab} (and \ref{eq:q}). Even if the $l$ coefficients $\Bb^*$  are sparse, since they come from a SVM--like dual problem, the expansion of $f^*$ will generally involves all  $n$ coefficients $\Ba^*$.

\section{Training in the Primal}
\label{sec:trainp}
In this \sectionname\ we analyze the optimization of the primal form of the non linear LapSVM problem, following the growing interest in training SVMs in the primal of the last few years \citep{joachims2006training,chap,pegasos}. 
Primal optimization of a SVM has strong similarities with the dual strategy \citep{chap}, and its implementation does not require any particularly complex optimization libraries. The focus of researchers has been mainly  on the solution of the linear SVM primal problem, showing how it can be solved  fast and efficiently \citep{joachims2006training,pegasos}. Most of the existing results can be directly extended to the non linear case by reparametrizing the linear output function $f(x)=\langle \Bw,\Bx\rangle+b$ with $\Bw=\sum_{i=1}^{l}\alpha_i\Bx_i$ and introducing the Gram matrix $K$. However this may result in a loss of efficiency. In \citet{chap,keerthifava} the authors investigated efficient solutions for the non linear SVM case. 

Primal and dual optimization are  two ways different of solving the same problem, neither of which can in general be considered a ``better'' approach.
Therefore why should a solution of the primal problem be useful  in the case of LapSVM? 
There are three primary reasons why such a solution may be preferable. 
First, it allows us to efficiently solve a single problem without the need of a two step solution. Second, it allows us to very quickly compute good \textit{approximate} solutions, 
while the exact relation between  approximate solutions of the dual and original problems
may be involved.  Third, since it allows us to directly ``manipulate'' the $\Ba$ coefficients of $f$ without passing through the $\Bb$ ones, greedy techniques for incremental building of the LapSVM classifier are easier to manage \citep{vikassphd}. We believe that studying the primal LapSVM problem is the basis for future investigations and improvements of this classifier.

\begin{figure}[ht]
	  \centering
		\includegraphics[width=0.5\textwidth]{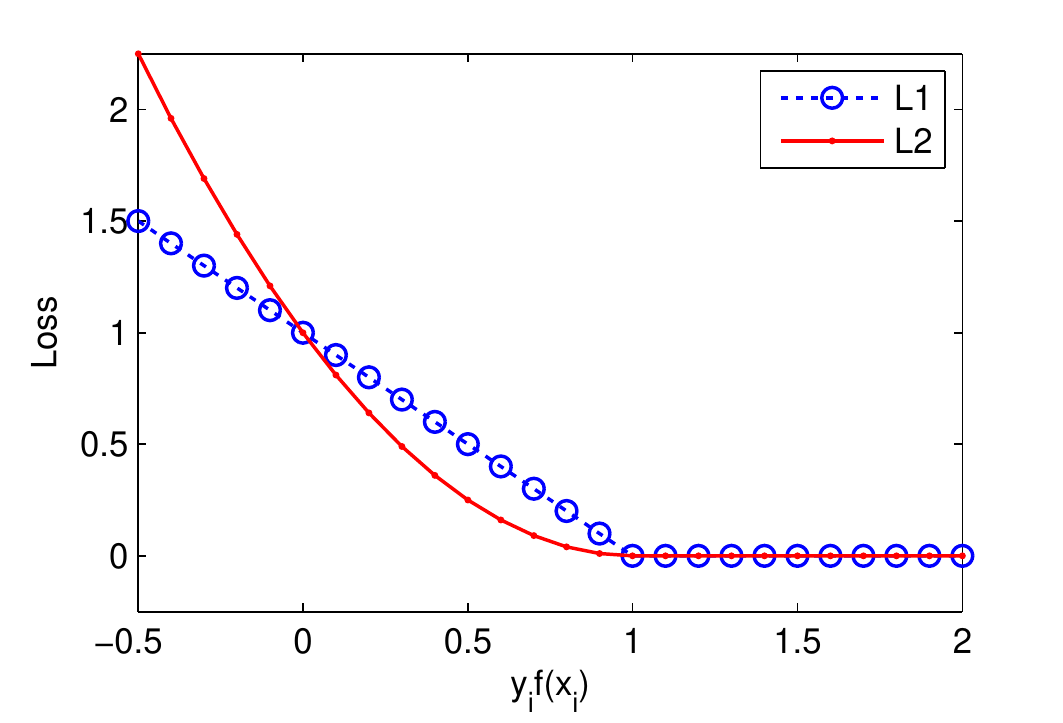}\\			
	\caption{$L_1$ hinge loss, piecewise linear, continuous and non differentiable in $y_if_(\Bx_i)=1$. $L_2$ hinge loss, continuous and differentiable.}	
	\label{fig:l1l2}	
\end{figure}

We rewrite the primal LapSVM problem of \equationname~\ref{eq:lapsvm} by considering the representation of $f$ of \equationname~\ref{eq:rt}, the intrinsic regularized of \equationname~\ref{eq:norm}, and by indicating with $\Bk_i$ the $i$-th column of the matrix $K$
\begin{equation*}
\min_{\Ba\in\RealSet^n,b\in\RealSet}\sum_{i=1}^{l}{V(x_i,y_i,\Bk_i^T\Ba+b)+\ga \Ba^TK\Ba +\gi\Ba^TKLK\Ba}.
\end{equation*}
Note that, for completeness, we included the bias $b$ in the expansion of $f$. Such bias does not affect the intrinsic norm that is actually a sum of squared differences of $f$ evaluated on pair of points\footnote{If the Laplacian is normalized then the expression of the intrinsic norm changes. This must be taken into account when computing the bias.}. We use a squared hinge loss, or $L_2$ loss, for the labeled examples, following \citet{chap} (see \figurename~\ref{fig:l1l2}). $L_2$ loss makes the LapSVM problem continuous and differentiable in $f$ and so in $\Ba$. The optimization problem after adding the scaling constant $\frac{1}{2}$ becomes
\begin{equation}
\min_{\Ba\in\RealSet^n,b\in\RealSet}\frac{1}{2}(\sum_{i=1}^{l}{\max(1-y_i(\Bk_i^T\Ba+b),0)^2+\ga \Ba^TK\Ba +\gi\Ba^TKLK\Ba}).
\label{eq:lapsvmp}
\end{equation}

We solved such convex problem by Newton's method and by preconditioned conjugate gradient, comparing their complexities and the complexity of the original LapSVM solution, and showing a parallel with the SVM case. The two solution strategies are analyzed in the following Subsections, while a large set of experimental results are collected in \sectionname~\ref{sec:results}.

\subsection{Newton's Method}
\label{sec:newton}
The problem of \equationname~\ref{eq:lapsvmp} is piecewise quadratic and the Newton's method appears a natural choice for an efficient minimization, since it builds a quadratic approximation of the function. After indicating with $\Bz$ the vector $\Bz=[b,\Ba^T]^T$, each Newton's step consists of the following update
\begin{equation}
\Bz^{t}=\Bz^{t-1}-sH^{-1}\nabla
\label{eq:newton}
\end{equation}
where $t$ is the iteration number, $s$ is the step size, and $\nabla$ and $H$ are the gradient and the Hessian of \equationname~\ref{eq:lapsvmp} with respect to $\Bz$. We will use the symbols $\nabla_{\Ba}$ and $\nabla_{b}$ to indicate the gradient with respect to $\Ba$ and to $b$. 

Before continuing, we introduce the further concept of \textit{error vectors} \citep{chap}. The set of error vectors $\mathcal{E}$ is the subset of $\mathcal{L}$ with the points that generate a $L_2$ hinge loss value greater than zero. The classifier does not penalize all the remaining labeled points, since the $f$ function on that points produces outputs with the same sign of the corresponding label and with absolute value greater then or equal to it. In the classic SVM framework, error vectors correspond to support vectors at the optimal solution. In the case of LapSVM, all points are support vectors at optimum in the sense that they all generally contribute to the expansion of $f$. 

We have
\begin{equation}
\begin{array}{cl}
\nabla=\left[
\begin{array}{c}
\nabla_b\\
\nabla_{\Ba}\\
\end{array}
\right]
&=\left(
\begin{array}{c}
\sum_{i=1}^ly_i(y_i(\Bk_i\Ba+b)-1)\\
\sum_{i=1}^l\Bk_iy_i(y_i(\Bk_i\Ba+b)-1)+\ga K\Ba+\gi KLK\Ba
\end{array}
\right)=\\
& \\
&=\left(
\begin{array}{c}
\Bu^TI_{\mathcal{E}}(K\Ba+\Bu b)-\Bu I_{\mathcal{E}}\By\\
KI_{\mathcal{E}}(K\Ba+\Bu b)-KI_{\mathcal{E}}\By+\ga K\Ba+\gi KLK\Ba
\end{array}
\right)\\
\end{array}
\label{eq:grad}
\end{equation}
where $\Bu$ is the vector on $n$ elements equal to $1$ and $\By\in\{-1,0,1\}^n$ is the vector that collects the $l$ labels $y_i$ of the labeled training points and a set of $u$ zeros. 
The matrix $I_{\mathcal{E}}\in\RealSet^{n,n}$ is a diagonal matrix where the only elements different from 0 (and equal to 1) along the main diagonal are in positions corresponding to points of $\mathcal{S}$ that belong to $\mathcal{E}$ at the current iteration.

The Hessian $H$ is
\begin{equation*}
\begin{array}{lcl}
H=\left(
\begin{array}{cc}
\nabla^2_{b}&\nabla_{b}(\nabla_{\Ba})\\
\nabla_{\Ba}(\nabla_{b})&\nabla^2_{\Ba}
\end{array}
\right)&=&
\left(
\begin{array}{cc}
\Bu^TI_{\mathcal{E}}\Bu&\Bu^TI_{\mathcal{E}}K\\
KI_{\mathcal{E}}\Bu&KI_{\mathcal{E}}K+\ga K+\gi KLK
\end{array}
\right)=\\
& & \\
&=&\left(
\begin{array}{cc}
-\ga&\Bu^T\\
0&K
\end{array}
\right)
\left(
\begin{array}{cc}
0&\Bu^T\\
I_{\mathcal{E}}\Bu&I_{\mathcal{E}}K+\ga I+\gi LK
\end{array}
\right)
\end{array}
\end{equation*}
Note that the criterion function of \equationname~\ref{eq:lapsvmp} is not twice differentiable everywhere, so that $H$ is the generalized Hessian where the subdifferential in the breakpoint of the hinge function is set to 0. This leaves intact the least square nature of the problem, as in the Modified Newton's method proposed by \citet{keerthi2006modified} for linear SVMs. In other words, the contribute to the Hessian of the $L_2$ hinge loss is the same as the one of a squared loss $(y_i-f(\Bx_i))^2$ applied to error vectors only.

Combining the last two expressions we can write $\nabla$ as
\begin{equation}
\nabla=H\Bz-\left(
\begin{array}{c}
\Bu^T\\
K
\end{array}
\right)I_{\mathcal{E}}\By.
\label{eq:gradh}
\end{equation}

From \equationname~\ref{eq:gradh}, the Newton's update of \equationname~\ref{eq:newton} becomes
\begin{equation}
\begin{array}{lcl}
\Bz^{t}&=&\Bz^{t-1}-s\Bz^{t-1}+sH^{-1}
\left(
\begin{array}{c}
\Bu^T\\
K
\end{array}
\right)I_{\mathcal{E}}\By=\\
& & \\
&=&(1-s)\Bz^{t-1}+s\left(
\begin{array}{cc}
0&\Bu^T\\
I_{\mathcal{E}}\Bu&I_{\mathcal{E}}K+\ga I+\gi LK
\end{array}
\right)^{-1}
\left(
\begin{array}{cc}
-\ga&\Bu^T\\
0&K
\end{array}
\right)^{-1}
\left(
\begin{array}{c}
\Bu^T\\
K
\end{array}
\right)I_{\mathcal{E}}\By=\\
& & \\
&=&(1-s)\Bz^{t-1}+s\left(
\begin{array}{cc}
0&\Bu^T\\
I_{\mathcal{E}}\Bu&I_{\mathcal{E}}K+\ga I+\gi LK
\end{array}
\right)^{-1}
\left(
\begin{array}{c}
0\\
I_{\mathcal{E}}\By
\end{array}
\right).
\label{eq:step}
\end{array}
\end{equation}

Looking at the update rule of \equationname~\ref{eq:step} the analogies and differences with the solution of the linear system of \equationname~\ref{eq:ab} can be clearly appreciated. In particular, \equationname~\ref{eq:ab} relates the dual variables $\Bb$ with the $\Ba$ ones using the information on the ambient and intrinsic regularizers. The contribute of the labeled data has already been collected in the $\Bb$ variables, by solving the dual problem. Differently, in the update rule of of \equationname~\ref{eq:step} the information of the $L_2$ loss is represented by the $I_{\mathcal{E}}K$ term.

The step size $s$ must be computed by solving the one--dimensional minimization of \equationname~\ref{eq:lapsvmp} restricted on the ray from $\Bz^{t-1}$ to $\Bz^{t}$, with exact line search or backtracking \citep{boyd}. Convergence is declared when the set of error vectors does not change between two consecutive iterations of the algorithm. We can see that when $s=1$, \equationname~\ref{eq:step} shows that the vector $\Bz^{t-1}$ of the previous iteration is not explicitly included in the update rule of $\Bz^{t}$. The only variable element that defines the new $\Bz^{t}$ is $I_{\mathcal{E}}$, i.e. the set of error vectors $\mathcal{E}$. Exactly like in the case of primal SVMs \citep{chap}, in our experiments setting $s=1$ did not result in any convergence problems.

\subsubsection{Complexity Analysis}
\label{sec:cn}
Updating the $\Ba$ coefficients with the Newton's method costs $O(n^3)$, due to the matrix inversion in the update rule. Convergence is usually achieved in a tiny number of iterations, no more than 5 in our experiments (see \sectionname~\ref{sec:results}). In order to reduce the cost of each iteration, a Cholesky factorization of the Hessian can be computed before performing the first matrix inversion, and it can be updated using a rank--1 scheme during the following iterations, with cost $O(n^2)$ for each update \citep{seeger-low}. On the other hand,  this does not allow us to simplify $K$ in \equationname~\ref{eq:step}, otherwise the resulting matrix to be inverted will not be symmetric. Since a lot of time is wasted in the product by $K$ (that is usually dense), using the update of Cholesky factorization may not necessarily lead to a reduction of the overall training time.

Solving the primal problem using the Newton's method has the same complexity of the original LapSVM solution based on the dual problem discussed in \sectionname~\ref{eq:lapsvmp}. The only benefit of solving the primal problem with Netwon's method relies on the compact and simple formulation that does not requires the ``two step'' approach and a quadratic SVM solver as in the original dual formulation.

It is interesting to compare the training of SVMs in the primal with the one of LapSVMs for a better insight in the Newton's method based solution. SVMs can benefit from the inversion of only a portion of the whole Hessian matrix, that reduces the complexity of each iteration to $O(|\mathcal{E}|)$. Exploiting this useful aspect, the training algorithm can be run incrementally, reducing the complexity of the whole training process. In detail, an initial run on small portion of the available data is used to compute an approximate solution. Then the remaining training points, or some of them, are added. Due to the hinge loss and the currently estimated separating hyperplane, many of them will probably not belong to $\mathcal{E}$ so that its maximum cardinality during the whole training process will reasonably be smaller than $n$. Moreover, if we fix the step size $s=1$ the components of $\Ba$ that are not associated to an error vector will become zeros after the update, so that the Newton's method encourages sparser solutions. 

In the case of LapSVM those benefits are lost due to the presence of the intrinsic norm $\Bf^T L\Bf$. As a matter of fact and independently by the set $\mathcal{E}$, the constraints $w_{ij}(f(\Bx_i)-f(\Bx_j))^2$ make the Hessian a full matrix, avoiding the described useful block inversion of SVMs. If the classifier is build incrementally, the addiction of a new non--error vector point makes the current solution no more optimal. Following the considerations of \sectionname~\ref{sec:mr} on the $\| f \|^2_I$ norm, this suggests that a different regularizer may help the LapSVM solution with Newton's method to gain the benefits of the SVM one. Some steps in this direction has been moved by \cite{tsang2007lss}, and we will investigate a similar approach, but based on the primal problem, in future work.

Finally, we are assuming that $K$ and the matrix to invert on \equationname~\ref{eq:step} are non singular, otherwise the final expansion of $f$ will not be unique, even if the optimal value of the criterion function of \equationname~\ref{eq:lapsvmp} will be. 

\subsection{Preconditioned Conjugate Gradient}
\label{sec:pcg}
Instead of performing a costly Newton's step, the solution of the system $\nabla=0$ can be computed by conjugate gradient descent. In particular if we look at \equationname~\ref{eq:grad}, we can write the system $\nabla=0$ as as $H\Bz=\Bc$,
\begin{equation}
H\Bz=\Bc \Longrightarrow \left(
\begin{array}{cc}
\Bu^TI_{\mathcal{E}}\Bu&\Bu^TI_{\mathcal{E}}K\\
KI_{\mathcal{E}}\Bu&KI_{\mathcal{E}}K+\ga K+\gi KLK
\end{array}
\right)\Bz=
\left(
\begin{array}{c}
\Bu^T I_{\mathcal{E}}\By\\
KI_{\mathcal{E}}\By\\
\end{array}
\right).
\label{eq:g0}
\end{equation}
The convergence rate of conjugate gradient is related to the condition number of $H$ \citep{pain}. In the most general case, the presence of the terms $KI_{\mathcal{E}}K$ and $KLK$ leads to a not so well conditioned system and to a slow convergence rate. Fortunately the general fix investigated by \citet{chap} can be applied also in the case of LapSVMs, due to the quadratic form of the intrinsic regularizer. \equationname~\ref{eq:g0} can be factorized as
\begin{equation*}
\left(
\begin{array}{cc}
1&\Bo^T\\
\Bo&K
\end{array}
\right)
\left(
\begin{array}{cc}
\Bu^TI_{\mathcal{E}}\Bu&\Bu^TI_{\mathcal{E}}K\\
I_{\mathcal{E}}\Bu&I_{\mathcal{E}}K+\ga I+\gi LK
\end{array}
\right)\Bz=
\left(
\begin{array}{cc}
1&\Bo^T\\
\Bo&K
\end{array}
\right)
\left(
\begin{array}{c}
\Bu^T I_{\mathcal{E}}\By\\
I_{\mathcal{E}}\By\\
\end{array}
\right).
\end{equation*}

For instance, we can precondition the system of \equationname~\ref{eq:g0} with the symmetic matrix
\begin{equation*}
P=\left(
\begin{array}{cc}
1&\Bo^T\\
\Bo&K
\end{array}
\right)
\end{equation*}
so that the condition number of the original system is sensibly decreased. In the preconditioned gradient $\hat{\nabla}=P^{-1}\nabla$ the two previously described terms are reduced to $I_{\mathcal{E}}K$ and $LK$. Moreover, preconditioning is generally useful when such product can be efficiently computed and in our problem it comes at no additional computational cost. As in the Newton's method, we are assuming that $K$ is non singular, otherwise a small ridge can be added to fix it.

Classic rules for the update of the conjugate direction at each step are resumed by \citet{pain}. After some iterations the conjugacy of the descent directions tends to get lost due to roundoff floating point error, so a restart of the preconditioned conjugate gradient algorithm is required. The Fletcher--Reeves (FR) update is commonly used in linear optimization. Due to the piecewise nature of the problem, defined by the $I_{\mathcal{E}}$ matrix, we exploited the Pollak--Ribier (PR) formula, where restart can be automatically performed when the update term becomes negative \citep{pain}\footnote{Note that in the linear case FR and PR are equivalent.}. We experimentally evaluated that for the LapSVM problem such formula is generally the best choice, both for convergence speed and numerical stability. The iterative solution of LapSVM problem using preconditioned conjugate gradient (PCG) is reported in Algorithm \ref{alg:pcg}. The first iteration is actually a steepest descent one, and so it is after each restart of PCG, i.e. when $\rho$ becomes zero in Algorithm \ref{alg:pcg}.

Convergence is usually declared when the norm of the preconditioned gradient falls below a given threshold \citep{chap}, or when the current preconditioned gradient is roughly orthogonal with the real gradient \citep{pain}. We will investigate these conditions in \sectionname~\ref{sec:approx}.

\begin{algorithm}
\caption{Preconditioned Conjugate Gradient (PCG) for primal LapSVMs.}
\label{alg:pcg}       
\begin{algorithmic}
\STATE Let $t=0$, $\Bz^t=\Bo$, $\mathcal{E}=\mathcal{L}$, $\hat{\nabla}^t=-[\Bu^T\By, -\By^T]^T$, $\Bd^t=\hat{\nabla}^t$
\REPEAT
  \STATE $t=t+1$
	\STATE Find $s^*$ by line search on the line $\Bz^{t-1}+s\Bd^{t-1}$
	\STATE $\Bz^{t}=\Bz^{t-1}+s^*\Bd^{t-1}$
	\vspace{2mm}
	\STATE $\mathcal{E}=\{\Bx_i\in\mathcal{L}\ \ s.t.\ \ (\Bk_i\Ba^{t}+b^{t})y_i<1\}$
	\vspace{2mm}
	\STATE 
			$
			\hat{\nabla}^{t}=\left(
			\begin{array}{c}
			\Bu^TI_{\mathcal{E}}(K\Ba^{t}+\Bu b^{t}-\By)\\
			I_{\mathcal{E}}(K\Ba^{t}+\Bu b^{t}-\By)+\ga\Ba^{t}+\gi LK\Ba^{t}
			\end{array}
			\right)
			$
	\vspace{2mm}
  \STATE $\rho=\max(\frac{\hat{\nabla}^{t^T}P(\hat{\nabla}^{t}-\hat{\nabla}^{t-1})}{\hat{\nabla}^{{t-1}^T}P\hat{\nabla}^{t-1}},0)$
  \vspace{2mm}
	\STATE $\Bd^t=-\hat{\nabla}^{t}+\rho\Bd^{t-1}$
\UNTIL Goal condition
\end{algorithmic}
\end{algorithm}

\subsubsection{Line Search}
\label{sec:ls}
The optimal step length $s^*$ on the current direction of the PCG algorithm must be computed by backtracking or exact line search. At a generic iteration $t$ we have to solve
\begin{equation}
s^*=\argmin_{s\geq0}obj(\Bz^{t-1}+s\Bd^{t-1})
\label{eq:ls}
\end{equation}
where $obj$ is the objective function of \equationname~\ref{eq:lapsvmp}. 

The accuracy of the line search is crucial for the performance of PCG. When minimizing a quadratic form that leads to a linear expression of the gradient, line search can be computed in closed form. In our case, we have to deal with the variations of the set $\mathcal{E}$ (and of $I_{\mathcal{E}}$) for different values of $s$, so that a closed form solution cannot be derived, and we have to compute the optimal $s$ in an iterative way.

Due to the quadratic nature of \equationname~\ref{eq:ls}, the 1--dimensional Newton's method can be directly used, but the average number of line search iterations per PCG step can be very large, even if the cost of each of them is negligible with respect to the $O(n^2)$ of a PCG iteration. We can efficiently solve the line search problem analytically, as suggested by \citet{keerthi2006modified} for SVMs.




In order to simplify the notation, we discard the iteration index $t-1$ in the following description. Given the PCG direction $\Bd$, we compute for each point $\Bx_i\in\mathcal{L}$, being it an error vector or not, the step length $s_i$ for which its state switches. The state of a given error vector switches when it leaves the $\mathcal{E}$ set, whether the state of a point initially not in $\mathcal{E}$ switches when it becomes an error vector. We refer to the set of the former points with $\mathcal{Q}_1$ while the latter is $\mathcal{Q}_2$, with $\mathcal{L}=\mathcal{Q}_1\cup\mathcal{Q}_2$. The derivative of \equationname~\ref{eq:ls}, $\psi(s)={\partial{obj(\Bz+s\Bd)}}/{\partial{s}}$, is piecewise linear, and $s_i$ are the break points of such function. 

Let us consider, for simplicity, that $s_i$ are in a non decreasing order, discarding the negative ones. Starting from $s=0$, they define a set of intervals where $\psi(s)$ is linear and the $\mathcal{E}$ set does not change. We indicate with $\psi_j(s)$ the linear portion of $\psi(s)$ in the $j$--th interval. Starting with $j=1$, if the value $s\geq0$ for which $\psi_j(s)$ crosses zero is within such interval, then it is the optimal step size $s^*$, otherwise the following interval must be checked. The convergence of the process is guaranteed by the convexity of the function $obj$.

The zero crossing of $\psi_j(s)$ is given by $s=\frac{\psi_j(0)}{\psi_j(0)-\psi_j(1)}$, where the two points $(0,\psi_j(0))$ and $(1,\psi_j(1))$ determine the line $\psi_j(s)$. We indicate with $f_{d}(\Bx)$ the function $f(\Bx)$ whose coefficients are in $\Bd=[d_b,\Bd_{\alpha}^T]^T$, i.e. $f_{d}(\Bx_i)=\Bk_i^T\Bd_{\alpha}+d_b$, and we have
$$
\begin{array}{l}
\psi_j(0)=\sum_{\Bx_i\in\mathcal{E}_j}{(f(\Bx_i)-y_i)f_{d}(\Bx_i)}+\ga\Ba^TK\Bd_{\alpha}+\gi\Ba^TKLK\Bd_{\alpha}\\
\psi_j(1)=\sum_{\Bx_i\in\mathcal{E}_j}{(f(\Bx_i)+f_{d}(\Bx_i)-y_i)f_{d}(\Bx_i)}+\ga(\Ba+\Bd_{\alpha})^TK\Bd_{\alpha}+\gi(\Ba+\Bd_{\alpha})^TKLK\Bd_{\alpha}\\
\end{array}
$$
where $\mathcal{E}_j$ is the set of error vectors for the $j$--th interval.

Given $\psi_1(0)$ and $\psi_1(1)$, their successive values for increasing $j$ can be easily computed considering that only one point (that we indicate with $\Bx_j$) switches status moving from an interval to the following one. From this consideration we derived the following update rules
$$
\begin{array}{l}
\psi_{j+1}(0)=\psi_{j}(0)+\nu_{j}(f(\Bx_j)-y_{j})f_{d}(\Bx_j)\\
\psi_{j+1}(1)=\psi_{j}(1)+\nu_{j}(f(\Bx_j)+f_{d}(\Bx_j)-y_i)f_{d}(\Bx_j)\\
\end{array}
$$
where $\nu_{j}$ is $-1$ if $\Bx_j\in\mathcal{Q}_1$ and it is $+1$ if $\Br\in\mathcal{Q}_2$.

\subsubsection{Complexity Analysis}
\label{sec:cpcg}
Each PCG iteration requires to compute the $K\Ba$ product, leading to a complexity of $O(n^2)$ to update the $\Ba$ coefficients. The term $LK\Ba$ can then be computed efficiently from $K\Ba$, since the $L$ matrix is generally sparse. Note that, differently from the Newton's method and from the original dual solution of the LapSVM problem, we never explicitly compute the $LK$ product, whereas we always compute matrix by vector products. Even if $L$ is sparse, when the number of training point increases or $L$ is iterated many times, a large amount of time may be wasted in such matrix by matrix product, as we will show in \sectionname~\ref{sec:results}. Moreover, if the kernel matrix is sparse, the complexity drops to $O(n_{nz})$, where $n_{nz}$ is the maximum number of non null elements between $K$ and $L$. 

Convergence of the conjugate gradient algorithm is theoretically declared in $O(n)$ steps, but a solution very close to the optimal one can be computed with far less iterations. The convergence speed is related to the condition number of the Hessian \citep{pain}, that it is composed by a sum of three contributes (\equationname~\ref{eq:g0}). As a consequence, their condition numbers and weighting coefficients ($\ga$, $\gi$) have a direct influence in the convergence speed, and in particular the condition number of the $K$ matrix. For example, using a bandwidth of a Gaussian kernel that lead to a $K$ matrix close to the identity allows the algorithm to converge very quickly, but the accuracy of the classifier may not be sufficient.

Finally, PCG can be efficiently seeded with an initial rough estimate of the solution. This can be crucial for an efficient incremental building of the classifier with reduced complexity, following the one proposed for SVMs by \citet{keerthifava}.

\section{Approximating the Optimal Solution}
\label{sec:approx}
In order to reduce the training times, we want the PCG to converge as fast as possible to a good \textit{approximation} of the optimal solution. 
By appropriately selecting the goal condition of Algorithm \ref{alg:pcg}, we can discard iterations that may not lead to significant improvement in the classifier quality.

The common goal conditions for the PCG algorithm and, more generally, for gradient based iterative algorithms, rely on the norm of the gradient $\| \nabla \|$ \citep{boyd}, of the preconditioned gradient $\| \hat{\nabla} \|$ \citep{chap}, on the mixed product $\sqrt{\hat{\nabla}^T\nabla}$ \citep{pain}. These values are usually normalized by the first estimate of each of them. The value of the objective function $obj$ or its relative decrement between two consecutive iterations can also be checked, requiring some additional computations since the PCG algorithm never explicitly computes it. When one of such ``stopping'' values falls below the chosen threshold $\tau$ associated to it, the algorithm terminates\footnote{Thresholds associated to different conditions are obviously different, but, for simplicity in the description, we will refer to a generic threshold $\tau$.}. Moreover, a maximum number $t_{max}$ of iterations is generally specified. Tuning these parameters is crucial both for the time spent running the algorithm and the quality of the resulting solution.

It is really hard to find a trade--off between good approximation and low number of iterations, since $\tau$ and $t_{max}$ are strictly problem dependent. As an example, consider that the surface of $obj$, the objective function of \equationname~\ref{eq:lapsvmp}, varies among different choices of its parameters. Increasing or decreasing the values of $\ga$ and $\gi$ can lead to a less flat or a more flat region around the optimal point. Fixing in advance the values of $\tau$ and $t_{max}$ may cause an early stop too far from the optimal solution, or it may result in the execution of a large number of iterations without a significant improvement on the classification accuracy. 

The latter situation can be particularly frequent for LapSVMs. As described in \sectionname~\ref{sec:mr} the choice of the intrinsic norm $\Bf^T L \Bf$ introduces the soft constraint $f(\Bx_i)=f(\Bx_j)$ for nearby points $\Bx_i$, $\Bx_j$ along the underlying manifold. This allows the algorithm to perform a graph transduction and diffuse the labels from points in $\mathcal{L}$ to the unlabeled data $\mathcal{U}$. 

When the diffusion is somewhat complete and the classification hyperplane has assumed a quite stable shape around the available training data, similar to the optimal one, the intrinsic norm will keep contributing to the gradient until a balance with respect to the ambient norm (and to the $L_2$ loss on error vectors) is found. Due to the strictness of this constraint, it will still require some iterations (sometimes many) to achieve the optimal solution with $\| \nabla \|=0$, even if the decision function $y(\Bx)=sign( f(\Bx))$ will remain substantially the same. The described common goal conditions do not ``directly'' take into account the decision of the classifier, so that they do not appear appropriate to early stop the PCG algorithm for LapSVMs.

\begin{figure}[ht!]
	\centering
	\footnotesize
		\begin{minipage}{0.49\textwidth}	
	  \centering
		\includegraphics[width=1.0\textwidth]{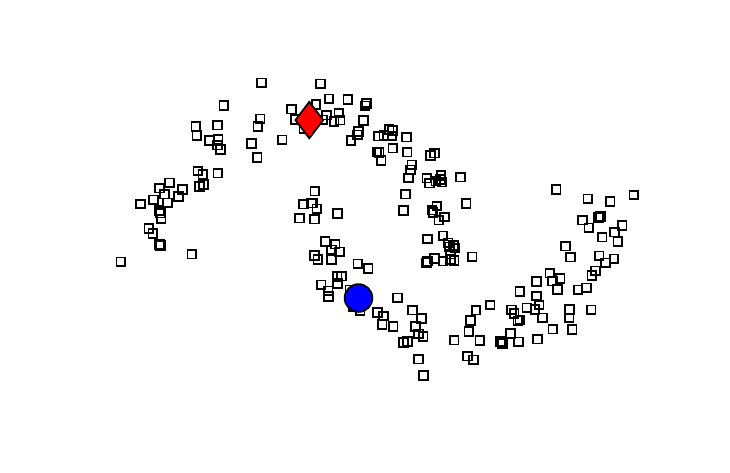}\\(a) The ``two moons'' dataset 
		\end{minipage}				
		\begin{minipage}{0.49\textwidth}	
	  \centering
		\includegraphics[width=1.0\textwidth]{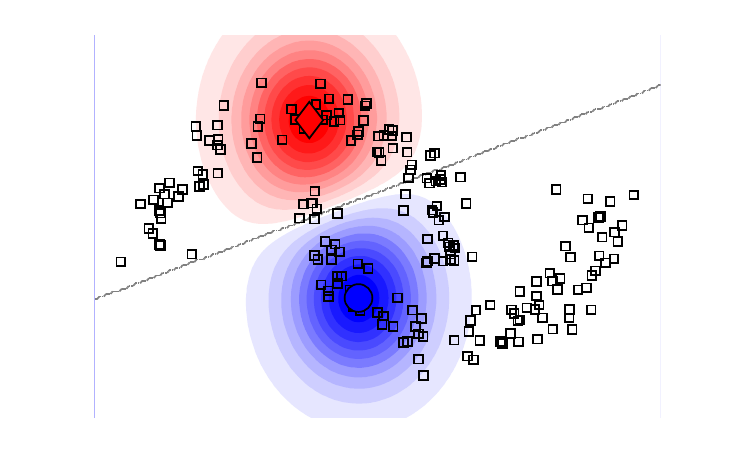}\\(b) 1 PCG iteration 
		\end{minipage}\\
		\begin{minipage}{0.49\textwidth}	
	  \centering
		\includegraphics[width=1.0\textwidth]{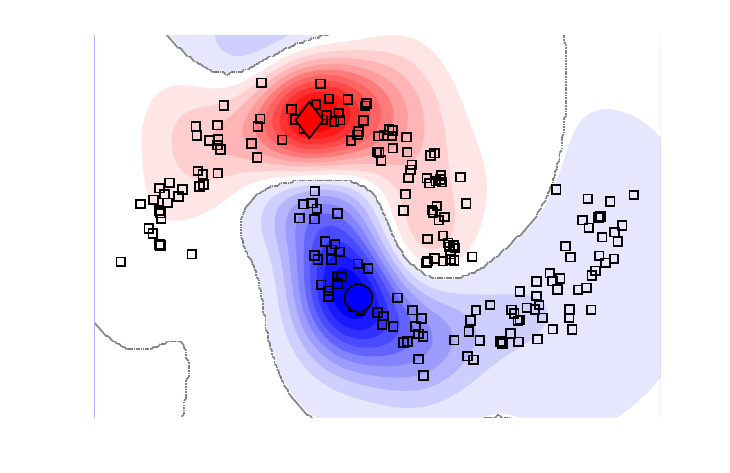}\\(c) 4 PCG iterations (\textbf{0\% error rate})   
		\end{minipage}	
		\begin{minipage}{0.49\textwidth}	
	  \centering
		\includegraphics[width=1.0\textwidth]{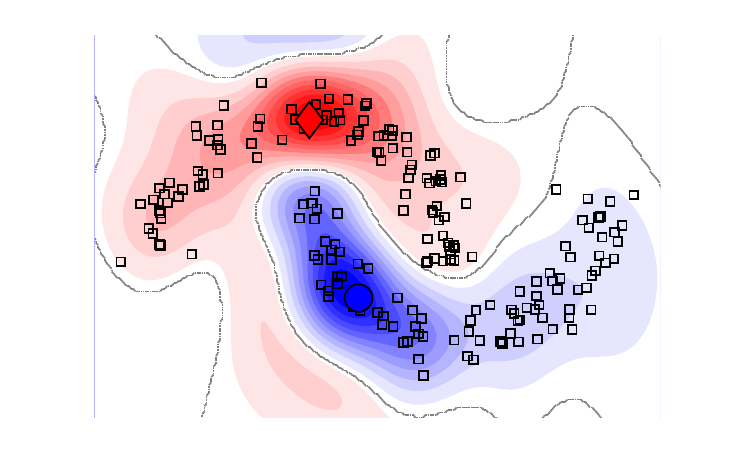}\\(d) 8 PCG iterations   
		\end{minipage}\\
		\begin{minipage}{0.49\textwidth}	
	  \centering
		\includegraphics[width=1.0\textwidth]{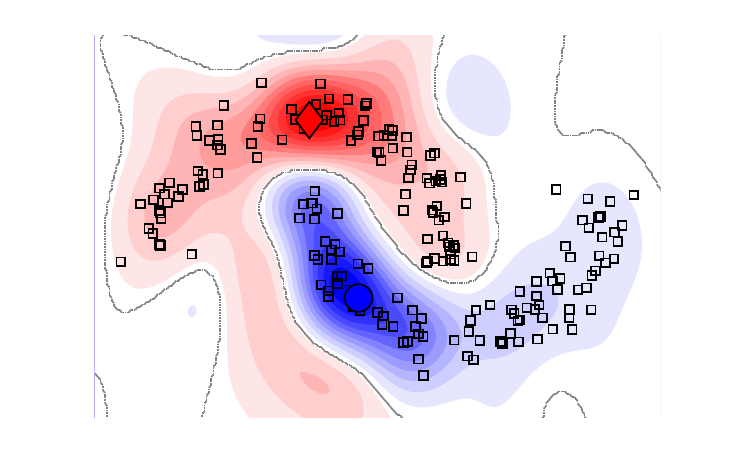}\\(e) 25 PCG iterations 
		\end{minipage}	
		\begin{minipage}{0.49\textwidth}	
	  \centering
		\includegraphics[width=1.0\textwidth]{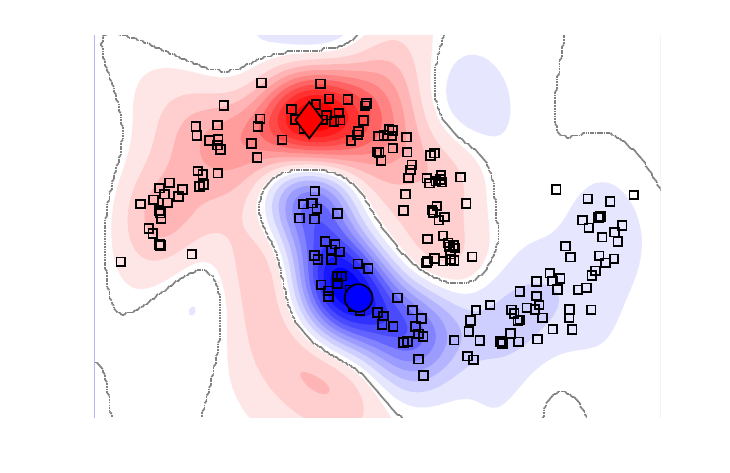}\\(f) Optimal solution
		\end{minipage}\\						
		\normalsize		
	\caption{(a) The ``two moons'' dataset (200 points, 2 classes, 2 labeled points indicated with a red diamond and a blue circle, whereas the remaining points are unlabeled) - (b-e) A LapSVM classifier trained with PCG, showing the result after a fixed number of iterations. The dark continuous line is the decision boundary ($f(\Bx)=0$) and the confidence of the classifier ranges from red ($f(\Bx)\geq 1$) to blue ($f(\Bx)\leq -1$) - (f) The optimal solution of the LapSVM problem computed by means of Newton's method}	
	\label{fig:2moons}	
\end{figure}

We investigate our intuition on the ``two moons'' dataset of \figurename~\ref{fig:2moons}(a), where we compare the decision boundary after each PCG iteration (\figurename~\ref{fig:2moons}(b)-(e)) with the optimal solution (computed by Newton's method, \figurename~\ref{fig:2moons}(f)). Starting with $\Ba=\Bo$, the first iteration exploits only the gradient of the $L_2$ loss on labeled points, since both the regularizing norms are zero. In the following iterations we can observe the label diffusion process along the manifold. After only 4 iterations we get a perfect classification of the dataset and a separating boundary not far from the optimal one. All the remaining iterations until complete convergence are used to slightly asses the coherence along the manifold required by the intrinsic norm and the balancing with the smoothness of the function, as can be observed by looking at the function values after 25 iterations. The most of changes influences regions far from the support of $P_X$, and it is clear that an early stop after 4 PCG steps would be enough to roughly approximate the accuracy of optimal solution. 

\begin{figure}[ht]
	\centering
		\includegraphics[width=0.45\textwidth]{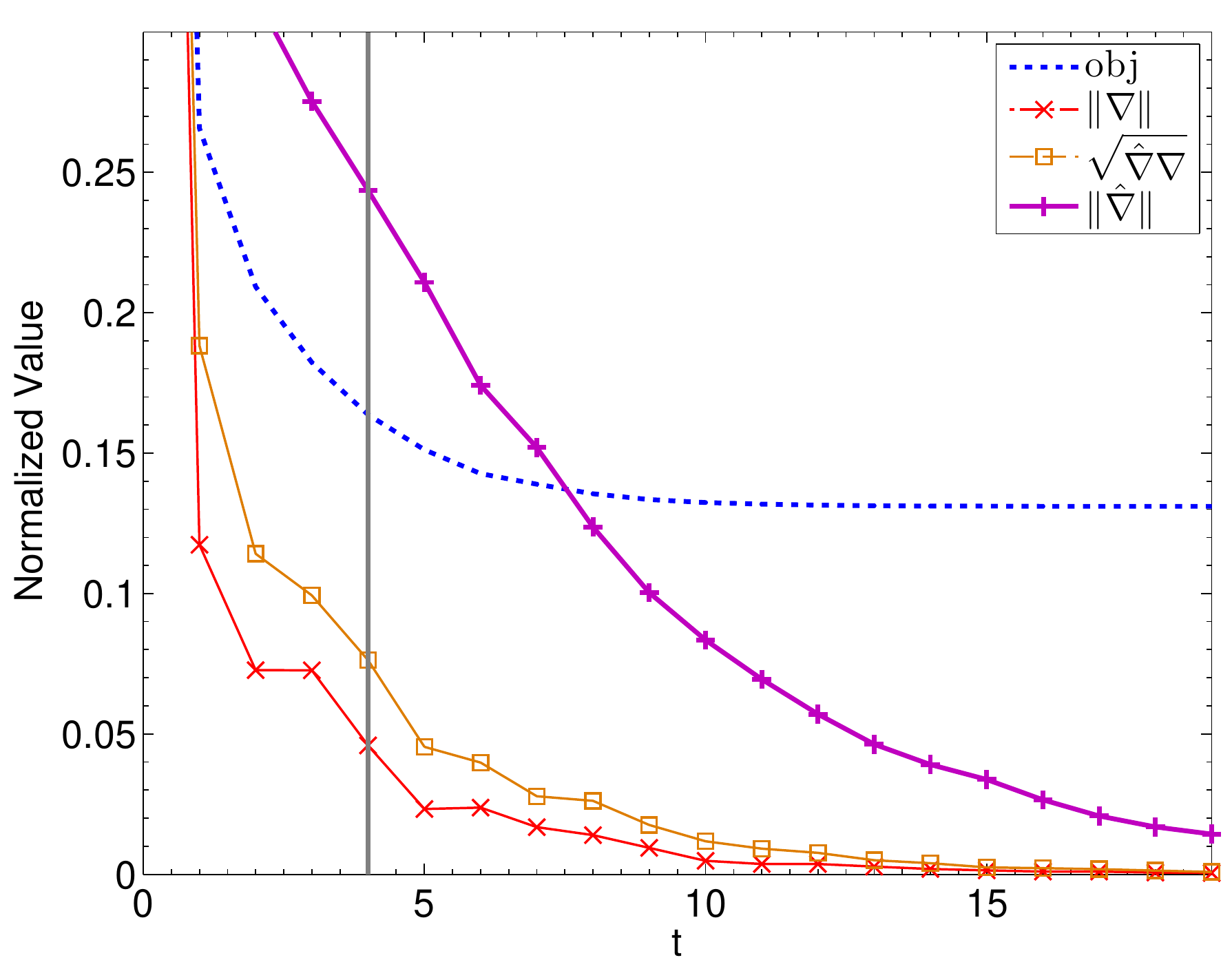}
	\caption{PCG example on the ``two moons'' dataset. The norm of the gradient $\| \nabla \|$, of the preconditioned gradient $\| \hat{\nabla} \|$, the value of the objective function $obj$ and of the mixed product $\sqrt{\hat{\nabla}^T\nabla}$ are displayed in function of the number of PCG iterations. The vertical line represents the number of iterations after which the error rate is $0\%$ and the decision boundary is quite stable.}	
	\label{fig:2moonsgraph}	
\end{figure}

In \figurename~\ref{fig:2moonsgraph} we can observe the values of the previously described general stopping criterion for PCG. After 4 iterations they are still sensibly decreasing, without reflecting real improvements in the classifier quality. The value of the objective function $obj$ starts to become more stable only after, say, 16 iterations, but it is still slightly decreasing even if it appears quite horizontal on the graph, due to its scale. It is clear that fixing in advance the parameters $\tau$ and $t_{max}$ is random guessing and it will probably result in a bad trade--off between training time and accuracy.

\subsection{Early Stopping Conditions}
\label{sec:heur}
Following these considerations, we propose to early stop the PCG algorithm exploiting the predictions of the classifier on the available data. Due to the high amount of unlabeled training points in the semi--supervised learning framework, the stability of the decision $y(\Bx)$, $\Bx \in \mathcal{U}$, can be used as a reference to early stop the gradient descent (\textit{stability check}). Moreover, if labeled validation data (set $\mathcal{V}$) is available for classifier parameters tuning, we can formulate a good stopping condition based on the classification accuracy on it (\textit{validation check}), that can be eventually merged to the previous one (\textit{mixed check}). 

In detail, when $y(\Bx)$ becomes quite stable between consecutive iterations or when $err(\mathcal{V})$, the error rate on $\mathcal{V}$, is not decreasing anymore, then the PCG algorithm should be stopped. Due to their heuristic nature, it is generally better to compare the predictions every $\theta$ iterations and within a certain tolerance $\eta$. As a matter of fact, $y(\Bx)$ may slightly change also when we are very close to the optimal solution, and $err(\mathcal{V})$ is not necessarily an always decreasing function. Moreover, labeled validation data in the semi--supervised setting is usually small with respect to the whole training data, labeled and unlabeled, and it may not be enough to represent the structure of the dataset. 

We propose very simple implementations of such conditions, that we used to achieve the results of \sectionname~\ref{sec:results}. Starting from these, many different and more efficient variants can be formulated, but it goes beyond the scope of this paper.
They are sketched in Algorithms \ref{alg:sta} and \ref{alg:val}.
\begin{algorithm}[!ht]
\caption{The \textit{stability check} for PCG stopping.}
\label{alg:sta}
\begin{algorithmic}
\STATE $d^{old}\leftarrow\Bo\in\RealSet^u$
\STATE $\eta\leftarrow1.5\%$
\STATE $\theta\leftarrow\sqrt{n}/2$
\STATE \textit{Every $\theta$ iterations do the followings:}
\STATE $d=[y(\Bx_j), \Bx_j\in\mathcal{U}, j=1,\ldots,u]^T$
\STATE $\tau=(100\cdot{{\|d-d^{old}\|}_{1}}/u)\%$
\IF {$\tau<\eta $}
	\STATE Stop PCG
\ELSE
  \STATE $d^{old}=d$
\ENDIF
\end{algorithmic}
\end{algorithm}
\begin{algorithm}[!ht]
\caption{The \textit{validation check} for PCG stopping.}
\label{alg:val}       
\begin{algorithmic}
\REQUIRE $\mathcal{V}$
\STATE $err\mathcal{V}^{old}\leftarrow100\%$
\STATE $\eta\leftarrow100\cdot | \mathcal{V} |^{-1}\%$
\STATE $\theta\leftarrow\sqrt{n}/2$
\STATE \textit{Every $\theta$ iterations do the followings:}
\IF {$err({\mathcal{V})}>(err\mathcal{V}^{old} - \eta)$}
	\STATE Stop PCG
\ELSE
  \STATE $err\mathcal{V}^{old}=err({\mathcal{V}})$
\ENDIF
\end{algorithmic}
\end{algorithm}
We computed the classifier decision every $\sqrt{n}/2$ iterations and we required the classifier to improve $err(\mathcal{V})$ by one correctly classifier example at every check, due to the usually small size of $\mathcal{V}$. Sometimes this can also help to avoid a slight overfitting of the classifier. 

Generating the decision $y(\Bx)$ on unlabeled data does not require heavy additional machinery, since the $K\Ba$ product must be necessarily computed to perform every PCG iteration. Its overall cost is $O(u)$. Differently, computing the accuracy on validation data requires the evaluation of the kernel function on validation points against the $n$ training ones, and $O(|\mathcal{V}| \cdot n)$ products, that is negligible with respect to the cost of a PCG iteration.

Finally, please note that even if these are generally early stopping conditions, sometimes they can help in the opposite situation. For instance they can also detect that the classifier needs to move some more steps toward the optimal solution than the ones limited by the selected $t_{max}$.

\section{Laplacian Regularized Least Squares}
\label{sec:laprlsc}
Laplacian Regularized Least Square Classifier (LapRLSC) has many analogies with the proposed $L_2$ hinge loss based LapSVMs. LapRLSC uses a squared loss function to penalize wrongly classified examples, leading to the following objective function
\begin{equation}
\min_{f\in\mathcal{H}_k}\sum_{i=1}^{l}{(y_i-f(\Bx_i))^2+\ga\|f\|_A^2+\gi\| f \|^2_I}.
\label{eq:laprlsc}
\end{equation}

The optimal $\Ba$ coefficients and the optimal bias $b$, collected in the vector $z$, can be obtained by solving the linear system
\begin{equation} 
\left(
\begin{array}{cc}
| \mathcal{L} |&\Bu^TI_{\mathcal{L}}K\\
KI_{\mathcal{L}}\Bu&KI_{\mathcal{L}}K+\ga K+\gi KLK
\end{array}
\right)\Bz=
\left(
\begin{array}{c}
\Bu^T \By\\
K\By\\
\end{array}
\right)
\label{eq:laprlscs}
\end{equation}
where $I_{\mathcal{L}}$ is the diagonal matrix $\in\RealSet^{n,n}$ with the first $l$ elements equal to 1 and the remaining $u$ elements equal to zero. 

Following the notation used for LapSVMs, in LapRLSCs we have a set of error vectors $\mathcal{E}$ that is actually fixed and equal to $\mathcal{L}$. As a matter of fact a LapRLSC requires the estimated function to interpolate the given targets in order to not incur in a penalty. In a hypothetic situation where all the labeled examples always belong to $\mathcal{E}$ during the training of a LapSVM classifier in the primal, then the solution will be the same of LapRLSC. 

Solving the least squares problem of LapRLSC can be performed by matrix inversion, after factoring and simplifying the previously defined matrix $P$ in \equationname~\ref{eq:laprlscs}. Otherwise the proposed PCG approach and the early stopping conditions can be directly used. In this case the classic instruments for linear optimization apply, and the required line search of \equationname~\ref{eq:ls} can be computed in closed form without the need of an iterative process,
\begin{equation*}
s^{*}=-\frac{\nabla^T\Bd}{\Bd^TH\Bd}
\end{equation*}
where $\nabla$ and $H$ are no more functions of $\mathcal{E}$.

As shown by \citet{lapsvm,coreg} and in the experimental Section of this paper, LapRLSC, LapSVM and primal LapSVM allow us to achieve similar classification performances. The interesting property of the LapSVM problem is that the effect of the regularization terms at a given iteration can be decoupled by the one of the loss function on labeled points, since the gradient of the loss function for correctly classified points is zero and do not disturb classifier design. This characteristic can be useful as a starting point for the study of some alternative formulations of the intrinsic norm regularizer.

\section{Experimental results}
\label{sec:results}
We ran a wide set of experiments to analyze the proposed solution strategies of the primal LapSVM problem. In this \sectionname\ we describe the selected datasets, our experimental protocol and the details on the parameter selection strategy. Then we show the main result of the proposed approach, very fast training of the LapSVM classifier with reduced complexity by means of early stopped PCG. We compare the quality of the $L_2$ hinge loss LapSVMs trained in the primal by Newton's method with respect to the $L_1$ hinge loss dual formulation and LapRLSCs. Finally, we describe the convergence speed and the impact on performances of our early stopping conditions.

As a baseline reference for the performances in the supervised setting, we selected two popular regularized classifiers, Support Vector Machines (SVMs) and Regularized Least Square Classifiers (RLSCs). We implemented and tested all the algorithms using Matlab 7.6 on a 2.33Ghz machine with 6GB of memory. The dual problem of LapSVM has been solved using the latest version of Libsvm \citep{libsvm}. Multiclass classification has been performed using the one--against--all approach.

\subsection{Datasets}
We selected eight popular datasets for our experiments. Most of them datasets has been already used in previous works to evaluate several semi--supervised classification algorithms \citep{cloud,lapsvm,coreg}, and all of them are available on the Web. G50C\footnote{It can be downloaded from \ttfamily{http://people.cs.uchicago.edu/\~\ \hspace{-3mm} vikass/manifoldregularization.html}.} is an artificial dataset generated from two unit covariance normal distributions with equal probabilities. The class means are adjusted so that the Bayes error is $5\%$. The COIL20 dataset is a collection of pictures of 20 different objects from the Columbia University. Each object has been placed on a turntable and at every 5 degrees of rotation a 32x32 gray scale image was acquired. The USPST dataset is a collection of handwritten digits form the USPS postal system. Images are acquired at the resolution of 16x16 pixels. USPST refers to the test split of the original dataset. We analyzed the COIL20 and USPST dataset in their original 20 and 10--class versions and also in their 2--class versions, to discard the effects on performances of the selected multiclass strategy. COIL20(B) discriminates between the first 10 and the last 10 objects, whereas USPST(B) from the first 5 digits and the remaining ones. PCMAC is a two--class dataset generated from the famous 20--Newsgroups collection, that collects posts on Windows and Macintosh systems. MNIST3VS8 is the binary version  of the MNIST dataset, a collection of 28x28 gray scale handwritten digit images from NIST. The goal is to separate digit 3 from digit 8. Finally, the FACEMIT dataset of the Center for Biological and Computational Learning at MIT contains 19x19 gray scale, PGM format, images of faces and non--faces. The details of the described datasets are resumed in \tablename~\ref{tabdata}. 

\begin{table}
\footnotesize
\centering
\begin{tabular}{lccc}
\hline\\[-3.8mm]
\textbf{Dataset}&\textbf{Classes}&\textbf{Size}&\textbf{Attributes}\\
\hline\\[-3.4mm]
G50C&2&550&50\\
COIL20(B)&2&1440&1024\\
PCMAC&2&1946&7511\\
USPST(B)&2&2007&256\\
COIL20&20&1440&1024\\
USPST&10&2007&256\\
MNIST3VS8&2&13966&784\\
FACEMIT&2&31022&361\\
\hline
\end{tabular}
\caption{Details of the datasets that have been used in the experiments.}
\label{tabdata} 
\end{table}

\subsection{Experimental protocol}
All presented results has been obtained by averaging them on different splits of the available data. In particular, a 4--fold cross--validation has been performed, randomizing the fold generation process for 3 times, for a total of 12 splits. Each fold contains the same number of per class examples as in the complete dataset. For each split, we have 3 folds that are used for training the classifier and the remaining one that constitutes the test set ($\mathcal{T}$). Training data has been divided in labeled ($\mathcal{L}$), unlabeled ($\mathcal{U}$) and validation sets ($\mathcal{V}$), where the last one is only used to tune the classifier parameters. The labeled and validation sets have been randomly selected from the training data such that at least one example per class is assured to be present on each of them, without any additional balancing constraints. A small number of labeled points has been generally selected, in order to simulate a semi--supervised scenario where labeling data has a large cost. The MNIST3VS8 and FACEMIT dataset are already divided in training and test data, so that the 4--fold generation process was not necessary, and just the random subdivision of training data has been performed. In particular, on the FACEMIT dataset we exchanged the original training and test sets, since, as a matter of fact, the latter is sensibly larger that the former. In this case our goal is just to show how we were able to handle a high amount of training data using the proposed primal solution with PCG, whereas it was not possible to do it with the original dual formulation of LapSVM. Due to the high unbalancing of such dataset, we report the macro error rates for it ($1-TP/2+TN/2$, where $TP$ and $TN$ are the rates of true positives and true negatives). Details are collected in \tablename~\ref{tabdet}.

\begin{table}
\footnotesize
\centering
\begin{tabular}{lcccc}
\hline\\[-3.8mm]
\textbf{Dataset}&$|$\textbf{$\mathcal{L}$}$|$&$|$\textbf{$\mathcal{U}$}$|$&$|$\textbf{$\mathcal{V}$}$|$&$|$\textbf{$\mathcal{T}$}$|$\\
\hline\\[-3.4mm]
G50C&50&314&50&136\\
COIL20(B)&40&1000&40&360\\
PCMAC&50&1358&50&488\\
USPST(B)&50&1409&50&498\\
COIL20&40&1000&40&360\\
USPST&50&1409&50&498\\
MNIST3VS8&80&11822&80&1984\\
FACEMIT&2&23973&50&6997\\
\hline
\end{tabular}
\caption{The number of data points in each split of the selected datasets, where $\mathcal{L}$ and $\mathcal{U}$ are the sets of labeled and unlabeled training points, respectively, $\mathcal{V}$ is the labeled set for cross--validating parameters whereas $\mathcal{T}$ is the out--of--sample test set.}
\label{tabdet}
\vspace{-5mm}
\end{table}

\subsection{Parameters}
We selected a Gaussian kernel function in the form $k(\Bx_i,\Bx_j)=\exp{-\frac{||\Bx_i-\Bx_j||}{2\sigma^2}}$ for each experiment, with the exception of the MNIST3VS8 where a polynomial kernel of degree 9 was used, as suggest by \citet{decoste2002training}. The other parameters were selected by cross--validating them on the $\mathcal{V}$ set. In order to speedup this step, the values of the Gaussian kernel width and of the parameters required to build the graph Laplacian (the number of neighbors, $nn$, and the degree, $p$) for the first six datasets were fixed as specified by \citet{coreg}. For details on the selection of such parameters please refer to \citet{coreg,cloud}. The graph Laplacian was computed by using its normalized expression. The optimal weights of the ambient and intrinsic norms, $\ga$, $\gi$, were determined by varying them on the grid $\{10^{-6},10^{-4},10^{-2},10^{-1},1,10,100\}$ and chosen with respect to validation error. For the FACEMIT dataset also the value $10^{-8}$ was considered, due to the high amount of training points.  The selected parameter values are reported in \tablename~\ref{tabparam} of Appendix A for reproducibility of the experiments. 

\subsection{Results}
Before going further into details, the training times of LapSVMs using the original dual formulation and the primal one are reported in \tablename~\ref{tabquick}, to empathize our main result\footnote{For a fair comparison of the training algorithms, the Gram matrix and the Laplacian were precomputed.}. The last column refers to LapSVMs trained using the best (in terms of accuracy) of the proposed stopping heuristics for each specific dataset. As expected, training in the primal by the Newton's method requires training times similar to the ones of the dual formulation. On the other hand, training by PCG with the proposed early stopping conditions shows an appreciable reduction of them on all datasets. As the size of labeled and unlabeled points increases, the improvement becomes very evident. In the MNIST3VS8 dataset we drop from roughly half an hour to two minutes. Both in the dual formulation of LapSVMs and in the primal one solved by means of Newton's method, a lot of time is spent in computing the $LK$ matrix product. Even if $L$ is sparse, as its size increases or when it is iterated the cost of this product becomes quite high. It is also the case of the PCMAC dataset, where the training time drops from 15 seconds to only 2 seconds when solving with PCG. Finally, also the memory requirements are reduced, since there is no need to explicitly compute, store and invert the Hessian when PCG is used. As an example, we trained the classifier on the FACEMIT dataset only using PCG. The high memory requirements of dual LapSVM and primal LapSVM solved with Newton's method, coupled with the high computational cost and slow training times, made the problem intractable for such techniques on our machine.

\begin{table}
\footnotesize
\centering
\begin{tabular}{l|lll}
\hline\\[-3.8mm]
\multirow{2}{*}{\textbf{Dataset}}&\multicolumn{3}{c}{Laplacian SVMs}\\
&\textbf{Dual (Original)}&\textbf{Primal (Newton)}&\textbf{Primal (PCG)}\\
\hline\\[-3.4mm]
G50C& 0.155 (0.004) & 0.134 (0.006) & \textbf{0.043} (0.006)\\
COIL20(B)& 0.311 (0.012)& 0.367 (0.097) & \textbf{0.097} (0.026)\\
PCMAC& 14.82 (0.104) & 15.756 (0.285) & \textbf{1.967} (0.269)\\
USPST(B)& 1.196 (0.015) & 1.4727 (0.2033) & \textbf{0.300} (0.030)\\
COIL20& 6.321 (0.441) & 7.26 (1.921) & \textbf{3.487} (1.734)\\
USPST& 12.25 (0.2) & 17.74 (2.44) & \textbf{2.032} (0.434)\\
MNIST3VS8& 2064.18 (3.1) & 2824.174 (105.07) & \textbf{114.441} (0.235)\\
FACEMIT&-&-& \textbf{35.728} (0.868)\\
\hline
\end{tabular}
\caption{Our main result. Training times (in seconds) of Laplacian SVMs using different algorithms (standard deviation in brackets). The time required to solve the original dual formulation and the primal solution with Newton's method are comparable, whereas solving the Laplacian SVMs problem in the primal with early stopped preconditioned conjugate gradient (PCG) offers a noticeable speedup.}
\label{tabquick} 
\end{table}

We investigate now the details of the solution of the primal LapSVM problem. In order to compare the effects of the different loss functions of LapRLSCs, LapSVMs trained in the dual, and LapSVMs trained in the primal, in \tablename~\ref{tabres} the classification errors of the described techniques are reported. For this comparison, the optimal solution of primal LapSVMs is computed by means of the Newton's method. The manifold regularization based techniques lead to comparable results, and, as expected, all semi--supervised approaches show a sensible improvement over classic supervised classification algorithms. The error rates of primal LapSVMs and LapRLSCs are really close, due to the described relationship of the $L_2$ hinge loss and the squared loss. We collected the average number of Newton's steps required to compute the optimal solution in \tablename~\ref{tabtnewton}. In all our experiments we always declared convergence in less than 6 steps.

\begin{table}
\footnotesize
\centering
\begin{tabular}{lllll}
\hline\\[-3.8mm]
\textbf{Dataset}&\textbf{Classifier}&\textbf{$\mathcal{U}$}&\textbf{$\mathcal{V}$}&\textbf{$\mathcal{T}$}\\
\hline\\[-3.4mm]
\multirow{5}{*}{\ G50C} &\ SVM&9.33 (2)&9.83 (3.46)&10.06 (2.8)\\
 &\ RLSC&10.43 (5.26)&10.17 (4.86)&11.21 (4.98)\\
 &\ LapRLSC&6.03 (1.32)&6.17 (3.66)&6.54 (2.11)\\ 
 &\ LapSVM Dual (Original)&5.52 (1.15)&5.67 (2.67)&5.51 (1.65)\\
 &\ LapSVM Primal (Newton)&6.16 (1.48)&6.17 (3.46)&7.27 (2.87)\\ 
 \hline\\[-3.1mm]
\multirow{5}{*}{\ COIL20(B)} &\ SVM&16.23 (2.63)&18.54 (6.2)&15.93 (3)\\
 &\ RLSC&16.22 (2.64)&18.54 (6.17)&15.97 (3.02)\\
 &\ LapRLSC&8.067 (2.05)&7.92 (3.96)&8.59 (1.9)\\ 
 &\ LapSVM Dual (Original)&8.31 (2.19)&8.13 (4.01)&8.68 (2.04)\\
 &\ LapSVM Primal (Newton)&8.16 (2.04)&7.92 (3.96)&8.56 (1.9)\\ 
\hline\\[-3.1mm]
\multirow{5}{*}{\ PCMAC} &\ SVM&19.65 (6.91)&20.83 (6.85)&20.09 (6.91)\\
 &\ RLSC&19.63 (6.91)&20.67 (6.95)&20.04 (6.93)\\
 &\ LapRLSC&9.67 (0.74)&7.67 (4.08)&9.34 (1.5)\\ 
 &\ LapSVM Dual (Original)&10.78 (1.83)&9.17 (4.55)&11.05 (2.94)\\
 &\ LapSVM Primal (Newton)&9.68 (0.77)&7.83 (4.04)&9.37 (1.51)\\ 
\hline\\[-3.1mm]
\multirow{5}{*}{\ USPST(B)} &\ SVM&17 (2.74)&18.17 (5.94)&17.1 (3.21)\\
 &\ RLSC&17.21 (3.02)&17.5 (5.13)&17.27 (2.72)\\
 &\ LapRLSC&8.87 (1.88)&10.17 (4.55)&9.42 (2.51)\\ 
 &\ LapSVM Dual (Original)&8.84 (2.2)&8.67 (4.38)&9.68 (2.48)\\
 &\ LapSVM Primal (Newton)&8.72 (2.15)&9.33 (3.85)&9.42 (2.34)\\ 
\hline\\[-3.1mm]
\multirow{5}{*}{\ COIL20} &\ SVM&29.49 (2.24)&31.46 (7.79)&28.98 (2.74)\\
 &\ RLSC&29.51 (2.23)&31.46 (7.79)&28.96 (2.72)\\
 &\ LapRLSC&10.35 (2.3)&9.79 (4.94)&11.3 (2.17)\\ 
 &\ LapSVM Dual (Original)&10.51 (2.06)&9.79 (4.94)&11.44 (2.39)\\
 &\ LapSVM Primal (Newton)&10.54 (2.03)&9.79 (4.94)&11.32 (2.19)\\ 
\hline\\[-3.1mm]
\multirow{5}{*}{\ USPST} &\ SVM&23.84 (3.26)&24.67 (4.54)&23.6 (2.32)\\
 &\ RLSC&23.95 (3.53)&25.33 (4.03)&24.01 (3.43)\\
 &\ LapRLSC&15.12 (2.9)&14.67 (3.94)&16.44 (3.53)\\ 
 &\ LapSVM Dual (Original)&14.36 (2.55)&15.17 (4.04)&14.91 (2.83)\\
 &\ LapSVM Primal (Newton)&14.98 (2.88)&15 (3.57)&15.38 (3.55)\\ 
\hline\\[-3.1mm]
\multirow{5}{*}{\ MNIST3VS8} &\ SVM&8.82 (1.11)&7.92 (4.73)&8.22 (1.36)\\
 &\ RLSC&8.82 (1.11)&7.92 (4.73)&8.22 (1.36)\\
 &\ LapRLSC&1.95 (0.05)&1.67 (1.44)&1.8 (0.3)\\ 
 &\ LapSVM Dual (Original)&2.29 (0.17)&1.67 (1.44)&1.98 (0.15)\\ 
 &\ LapSVM Primal (Newton)&2.2 (0.14)&1.67 (1.44)&2.02 (0.22)\\ 
 \hline\\[-3.1mm]
\multirow{3}{*}{\ FACEMIT} &\ SVM&39.8 (2.34)&38 (1.15)&34.61 (3.96)\\
 &\ RLSC&39.8 (2.34)&38 (1.15)&34.61 (3.96)\\
 &\ LapSVM Primal (PCG)&29.97 (2.51)&36 (3.46)&27.97 (5.38)\\ 
\hline
\end{tabular}
\caption{Comparison of the accuracy of LapSVMs trained by solving the primal (Newton's method) or the dual problem. The average classification error (standard deviation is reported brackets) is reported. Fully supervised classifiers (SVMs, RLSCs) represent the baseline performances. $\mathcal{U}$ is the set of unlabeled examples used to train the semi--supervised classifiers. $\mathcal{V}$ is the labeled set for cross--validating parameters whereas $\mathcal{T}$ is the out--of--sample test set. Results on the labeled training set $\mathcal{L}$ are omitted since all algorithms correctly classify such a few labeled training points.}
\label{tabres}
\end{table}

\begin{table}
\footnotesize
\centering
\begin{tabular}{ll}
\hline\\[-3.8mm]
\textbf{Dataset}&\textbf{Newton's Steps}\\
\hline\\[-3.4mm]
G50C &\ \ \ 1 (0)\\
COIL20(B) &\ \ \ 2.67 (0.78)\\
PCMAC &\ \ \ 2.33 (0.49)\\
USPST(B) &\ \ \ 4.17 (0.58)\\
COIL20 &\ \ \ 2.67 (0.75)\\
USPST &\ \ \ 4.26 (0.76)\\
MNIST3VS8 &\ \ \ 5 (0)\\
\hline
\end{tabular}
\caption{Newton's steps required to compute the optimal solution of the primal Laplacian SVM problem.}
\label{tabtnewton}
\end{table}

\begin{figure}[ht!]
	\centering
	  \hspace{-7mm}	
		\begin{minipage}{0.45\textwidth}	
	  \centering
		\includegraphics[width=1.1\textwidth]{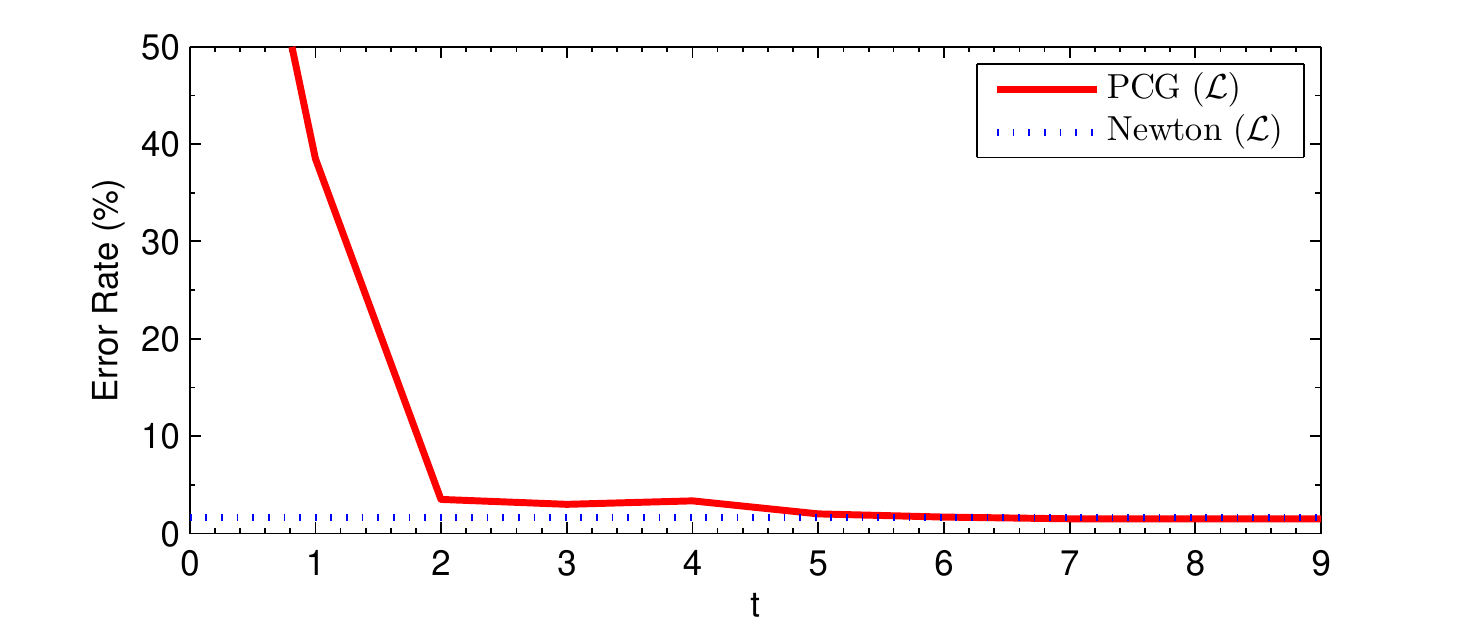}
		\end{minipage}
		\hspace{7mm}
		\begin{minipage}{0.45\textwidth}	
	  \centering
		\includegraphics[width=1.1\textwidth]{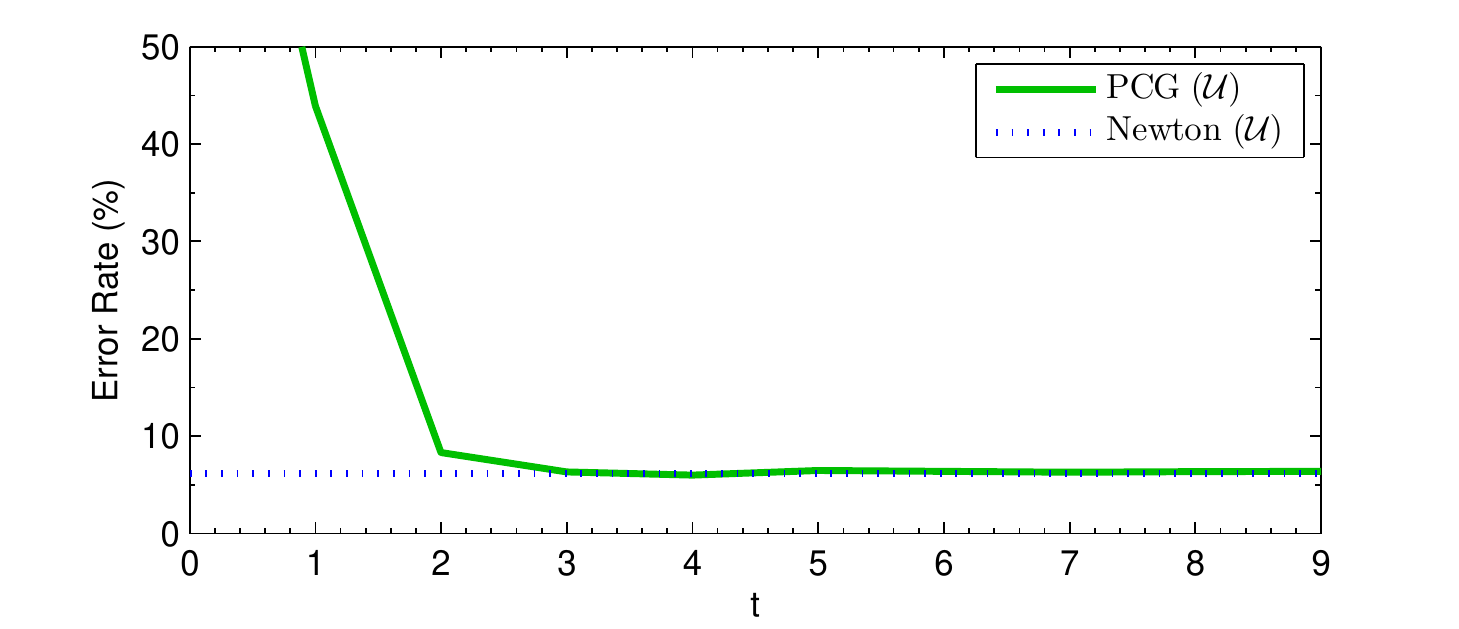}
		\end{minipage}\\	
		\hspace{-7mm}		
		\begin{minipage}{0.45\textwidth}	
	  \centering
		\includegraphics[width=1.1\textwidth]{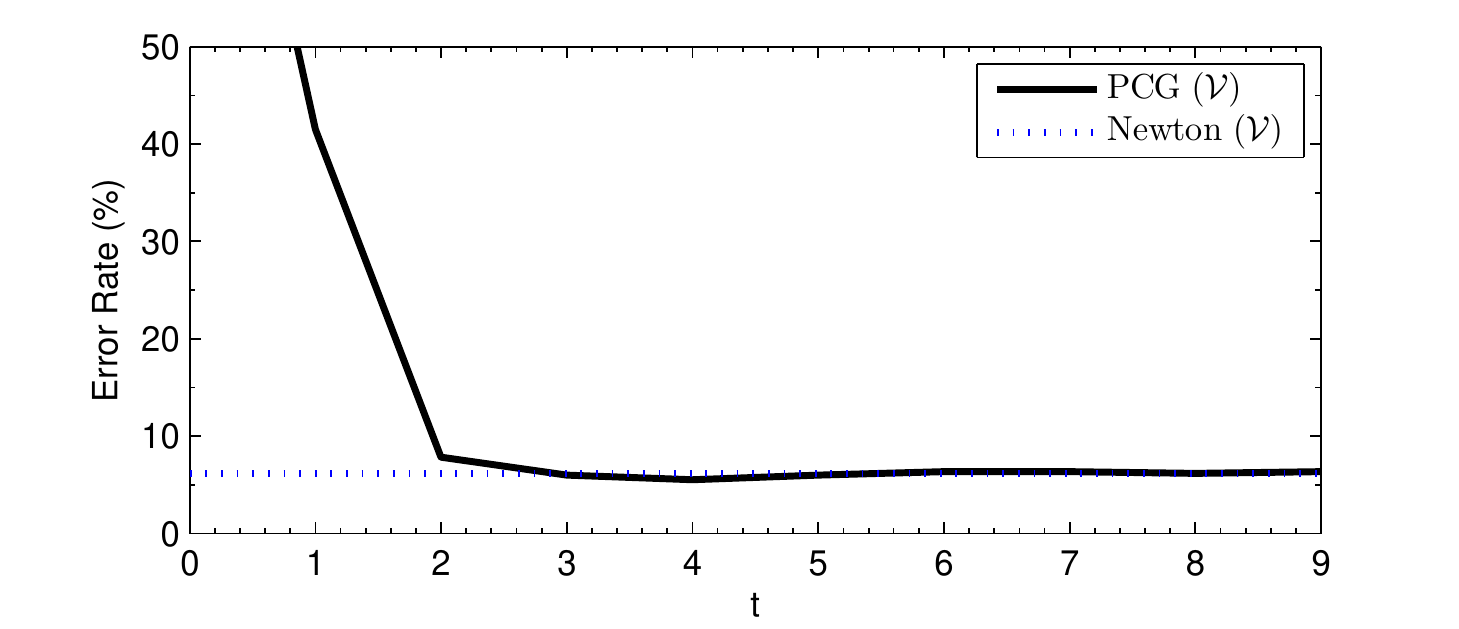}
		\end{minipage}
		\hspace{7mm}
		\begin{minipage}{0.45\textwidth}	
	  \centering
		\includegraphics[width=1.1\textwidth]{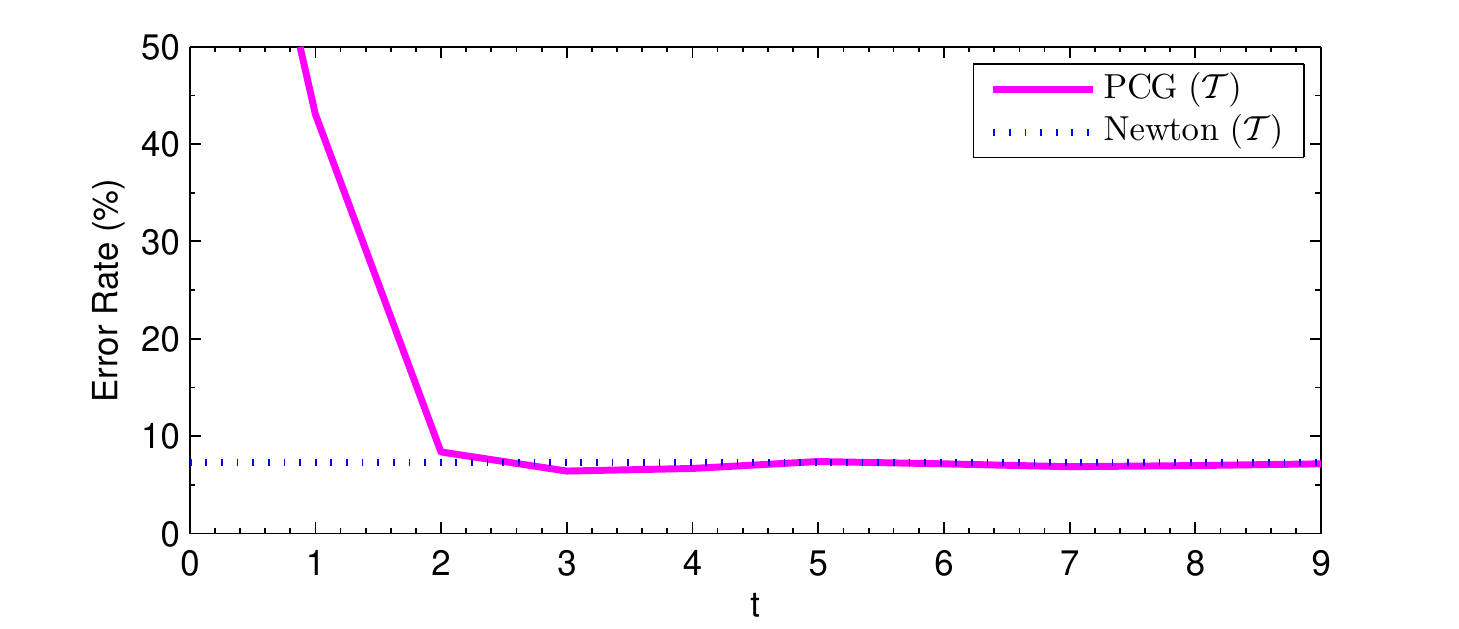}
		\end{minipage}\\
		\vspace{-2mm}				
	\caption{G50C dataset: error rate on $\mathcal{L}$, $\mathcal{U}$, $\mathcal{V}$, $\mathcal{T}$ of the Laplacian SVM classifier trained in the primal by preconditioned conjugate gradient (PCG), with respect to the number of gradient steps $t$. The error rate of the primal solution computed by means of Newton's method is reported as a horizontal line.}	
	\vspace{-2mm}
	\label{figtcg1}	
\end{figure}

\begin{figure}[ht!]
	\centering
	\hspace{-7mm}	
		\begin{minipage}{0.45\textwidth}	
	  \centering
		\includegraphics[width=1.1\textwidth]{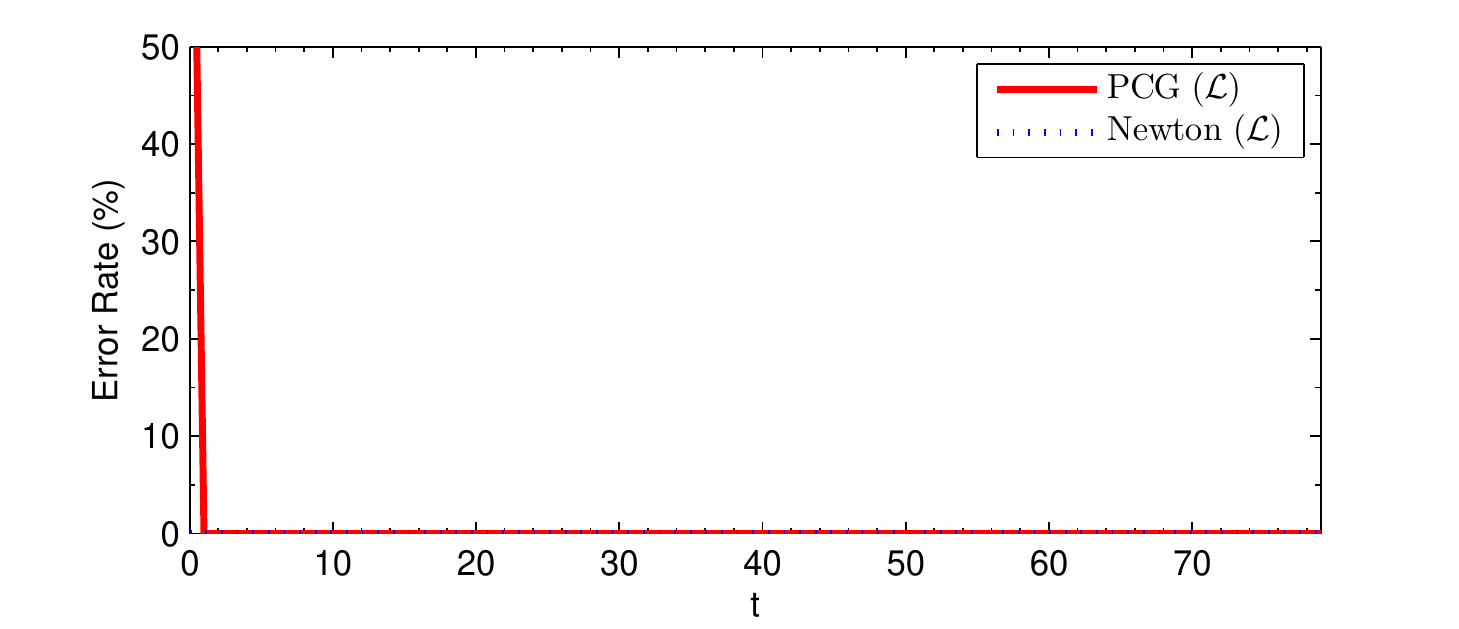}
		\end{minipage}
		\hspace{7mm}
		\begin{minipage}{0.45\textwidth}	
	  \centering
		\includegraphics[width=1.1\textwidth]{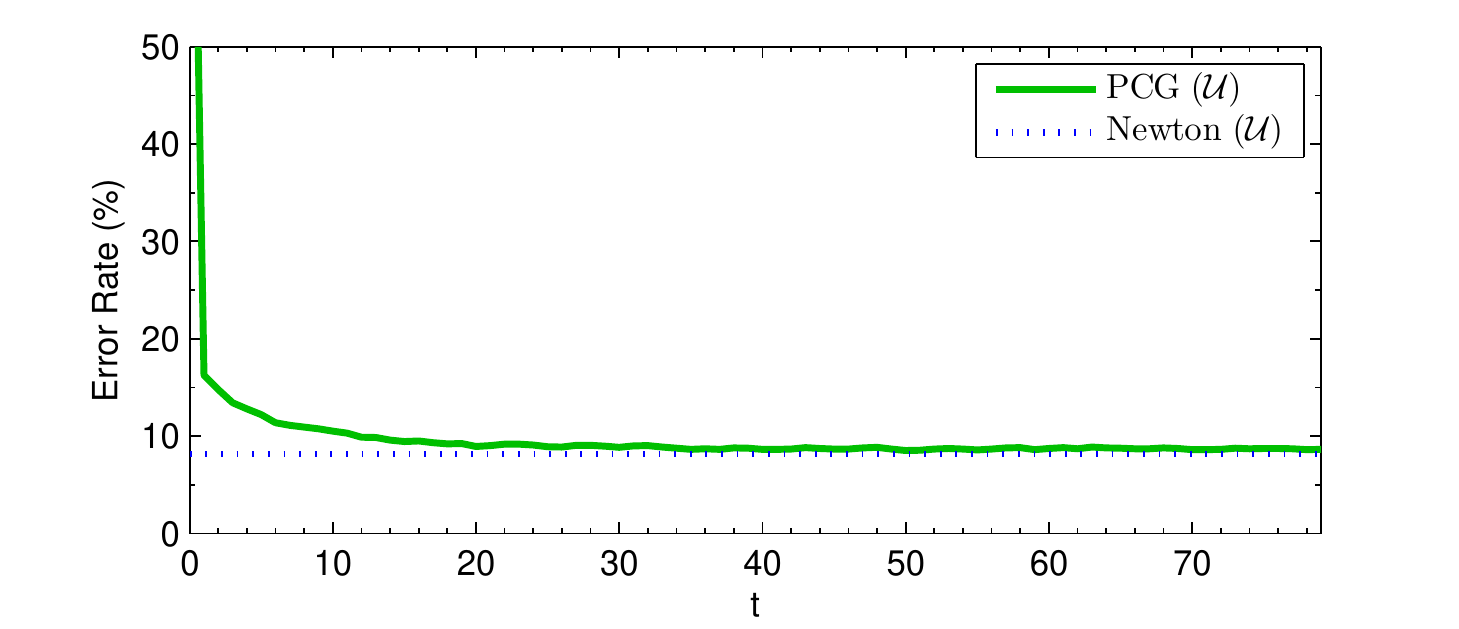}
		\end{minipage}\\		
    \hspace{-7mm}		
		\begin{minipage}{0.45\textwidth}	
	  \centering
		\includegraphics[width=1.1\textwidth]{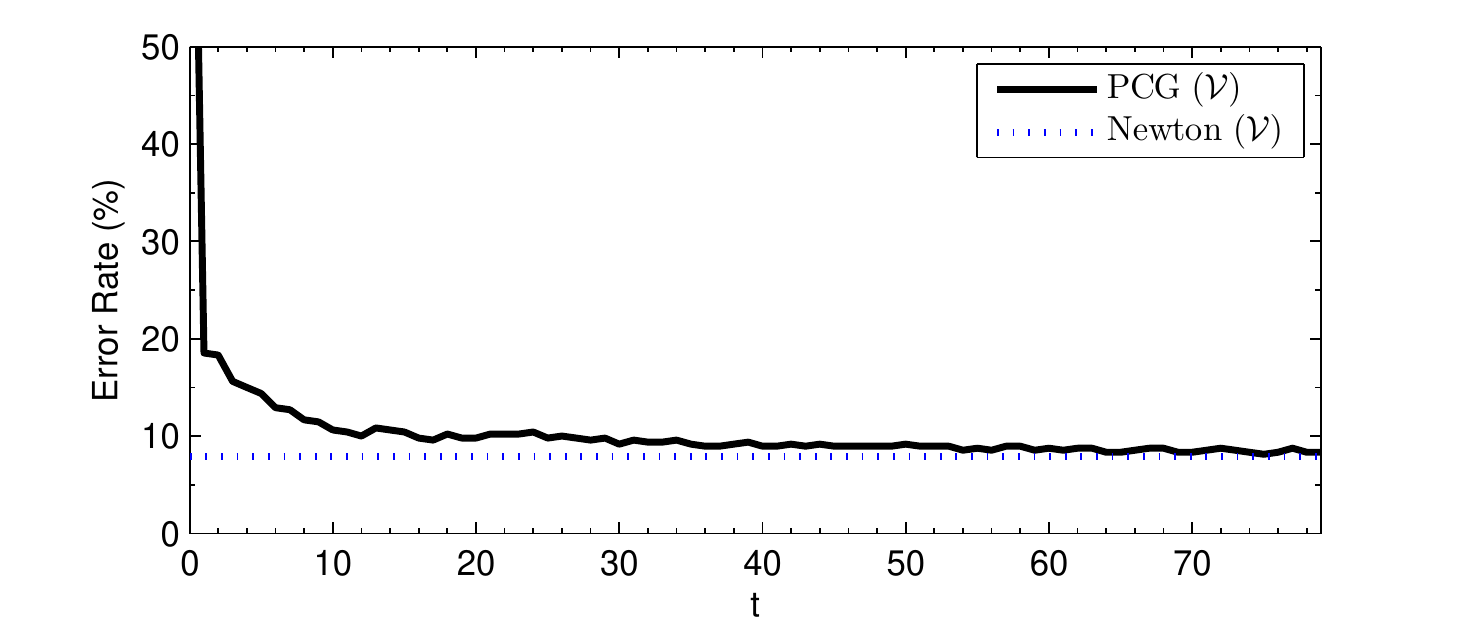}
		\end{minipage}
		\hspace{7mm}
		\begin{minipage}{0.45\textwidth}	
	  \centering
		\includegraphics[width=1.1\textwidth]{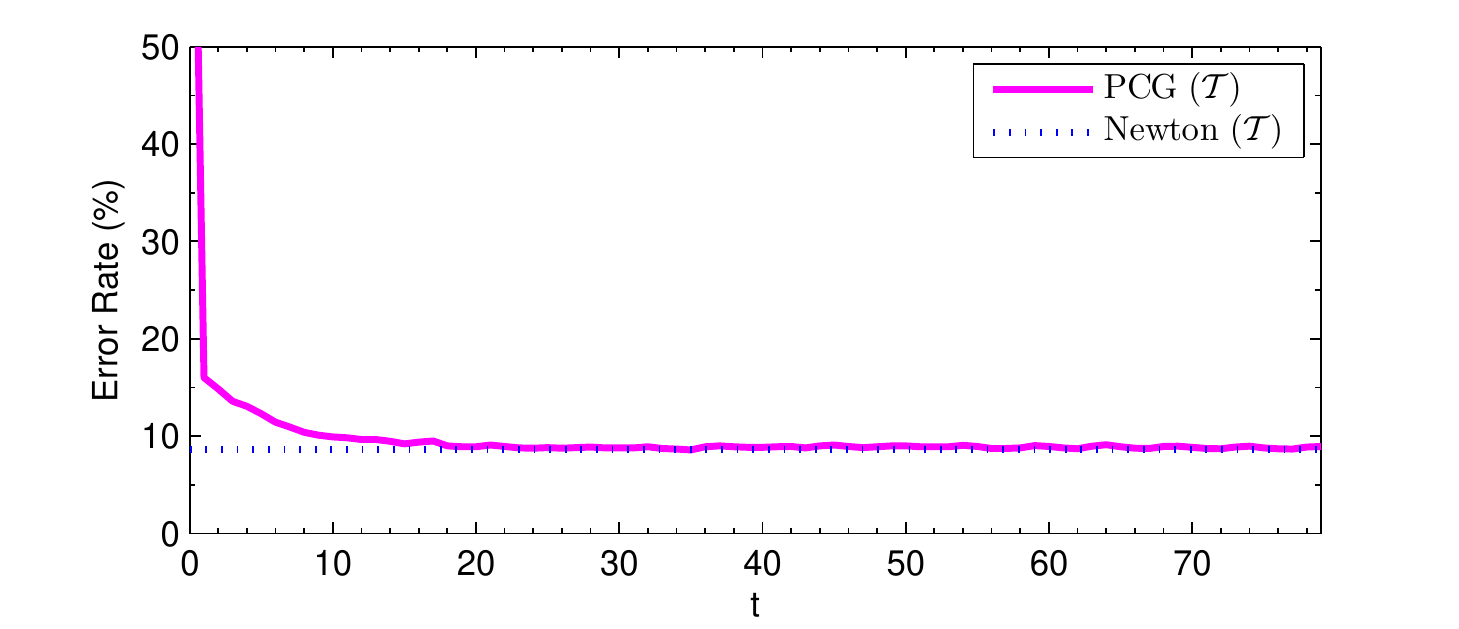}
		\end{minipage}\\	
		\vspace{-2mm}				
	\caption{COIL20(B) dataset: error rate on $\mathcal{L}$, $\mathcal{U}$, $\mathcal{V}$, $\mathcal{T}$ of the Laplacian SVM classifier trained in the primal by preconditioned conjugate gradient (PCG), with respect to the number of gradient steps $t$. The error rate of the primal solution computed by means of Newton's method is reported as a horizontal line.}	
	 \vspace{-2mm}
	\label{figtcg6}	
\end{figure}

\begin{figure}[ht!]
	\centering
	\hspace{-7mm}	
		\begin{minipage}{0.45\textwidth}	
	  \centering
		\includegraphics[width=1.1\textwidth]{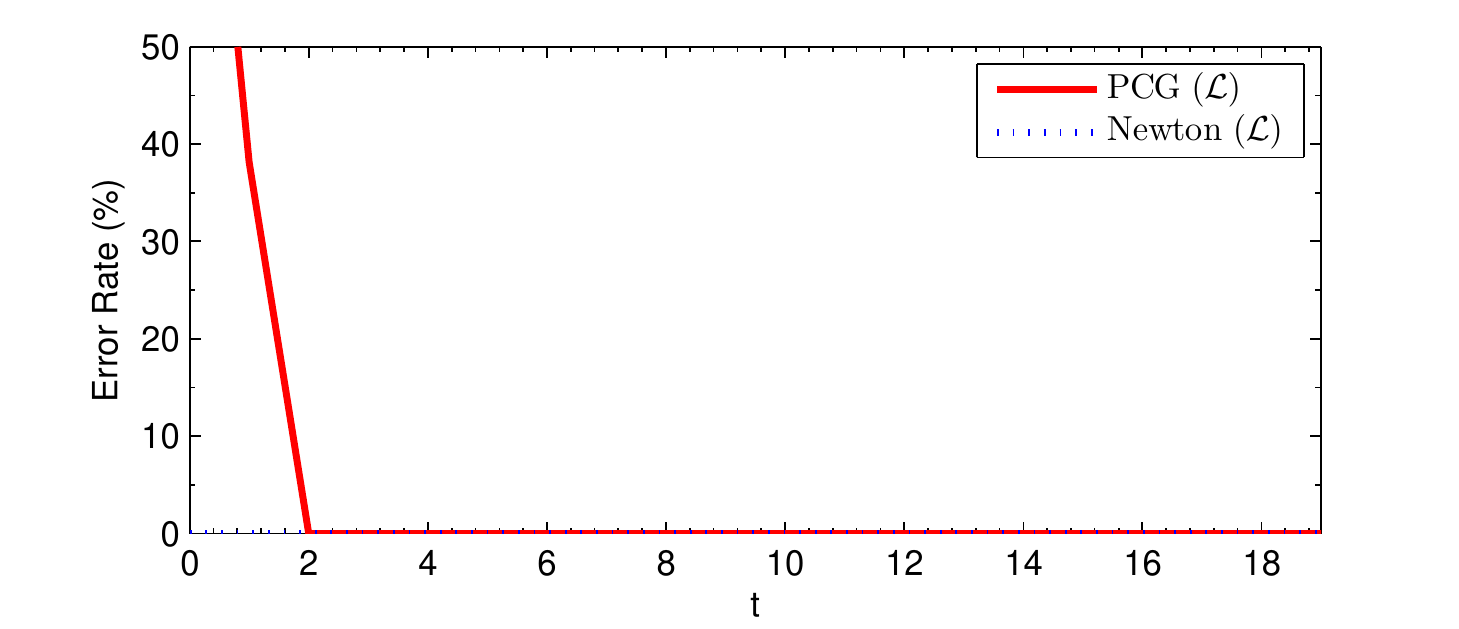}
		\end{minipage}
		\hspace{7mm}
		\begin{minipage}{0.45\textwidth}	
	  \centering
		\includegraphics[width=1.1\textwidth]{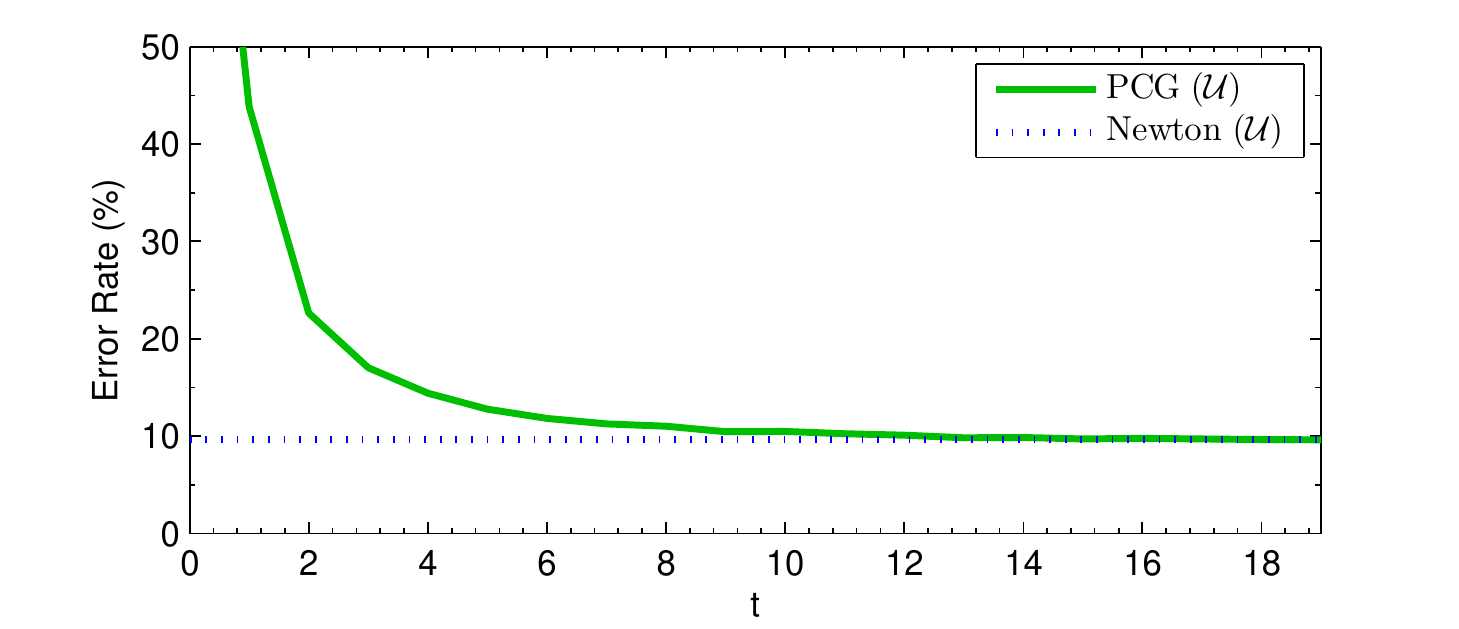}
		\end{minipage}\\	
		\hspace{-7mm}			
		\begin{minipage}{0.45\textwidth}	
	  \centering
		\includegraphics[width=1.1\textwidth]{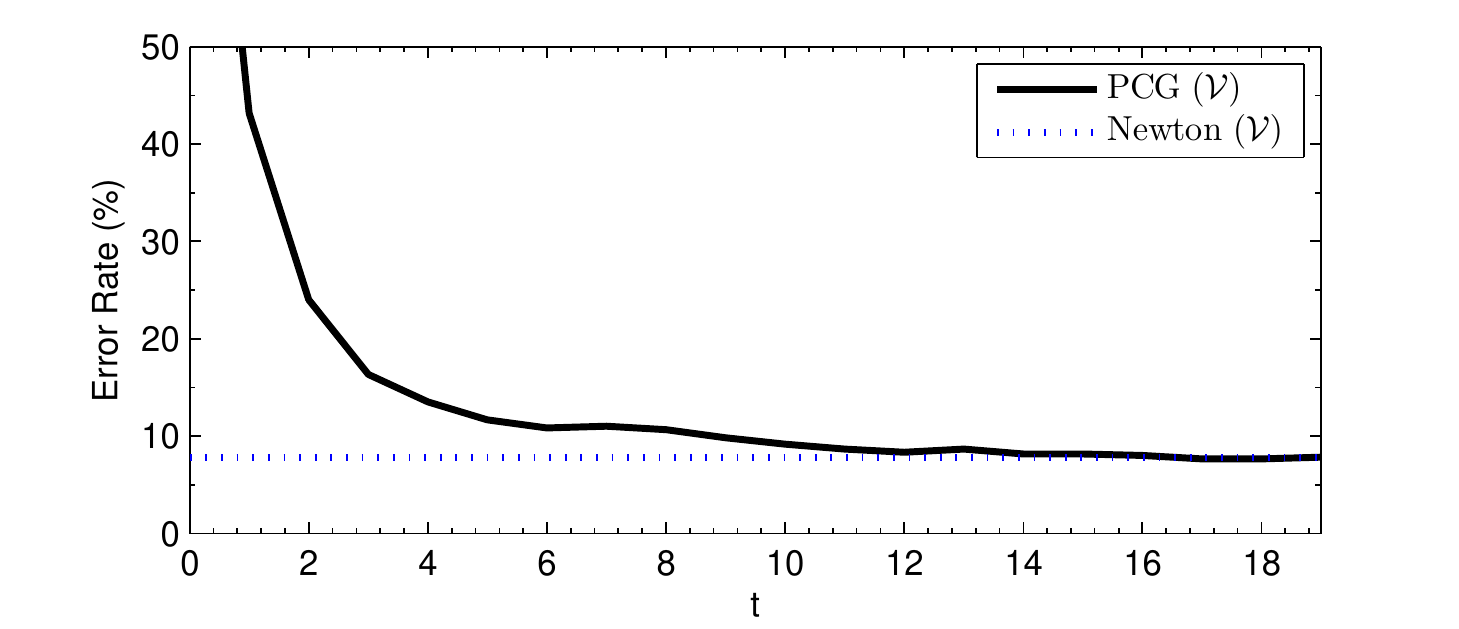}
		\end{minipage}
		\hspace{7mm}
		\begin{minipage}{0.45\textwidth}	
	  \centering
		\includegraphics[width=1.1\textwidth]{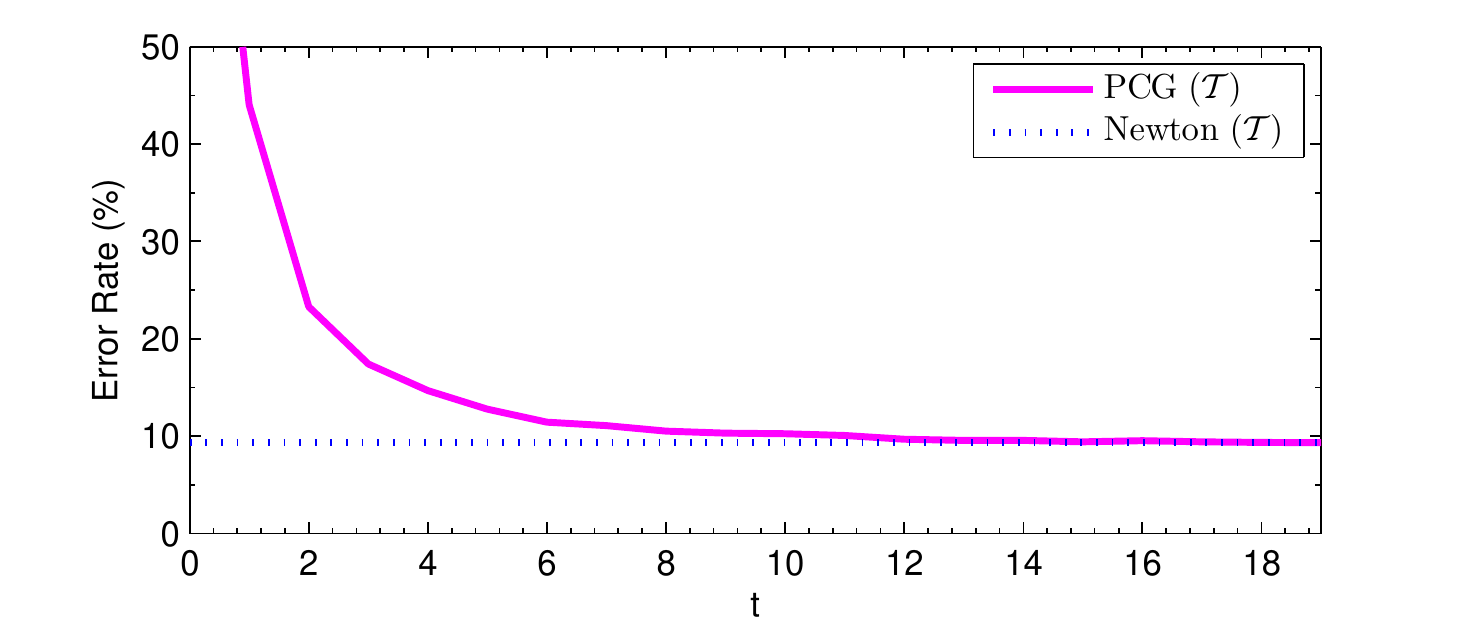}
		\end{minipage}\\
	\vspace{-2mm}				
	\caption{PCMAC dataset: error rate on $\mathcal{L}$, $\mathcal{U}$, $\mathcal{V}$, $\mathcal{T}$ of the Laplacian SVM classifier trained in the primal by preconditioned conjugate gradient (PCG), with respect to the number of gradient steps $t$. The error rate of the primal solution computed by means of Newton's method is reported as a horizontal line.}	
	\vspace{-2mm}
	\label{figtcg4}	
\end{figure}

\begin{figure}[ht!]
	\centering
	\hspace{-7mm}	
		\begin{minipage}{0.45\textwidth}	
	  \centering
		\includegraphics[width=1.1\textwidth]{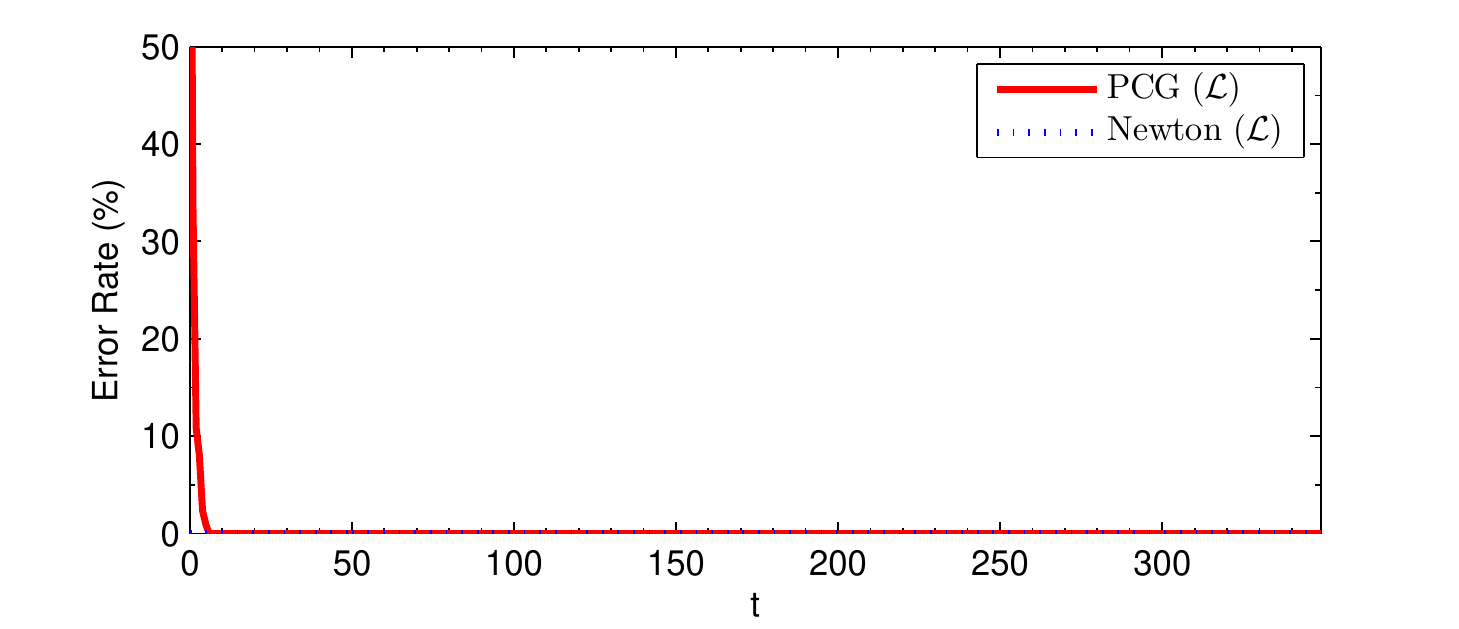}
		\end{minipage}
		\hspace{7mm}
		\begin{minipage}{0.45\textwidth}	
	  \centering
		\includegraphics[width=1.1\textwidth]{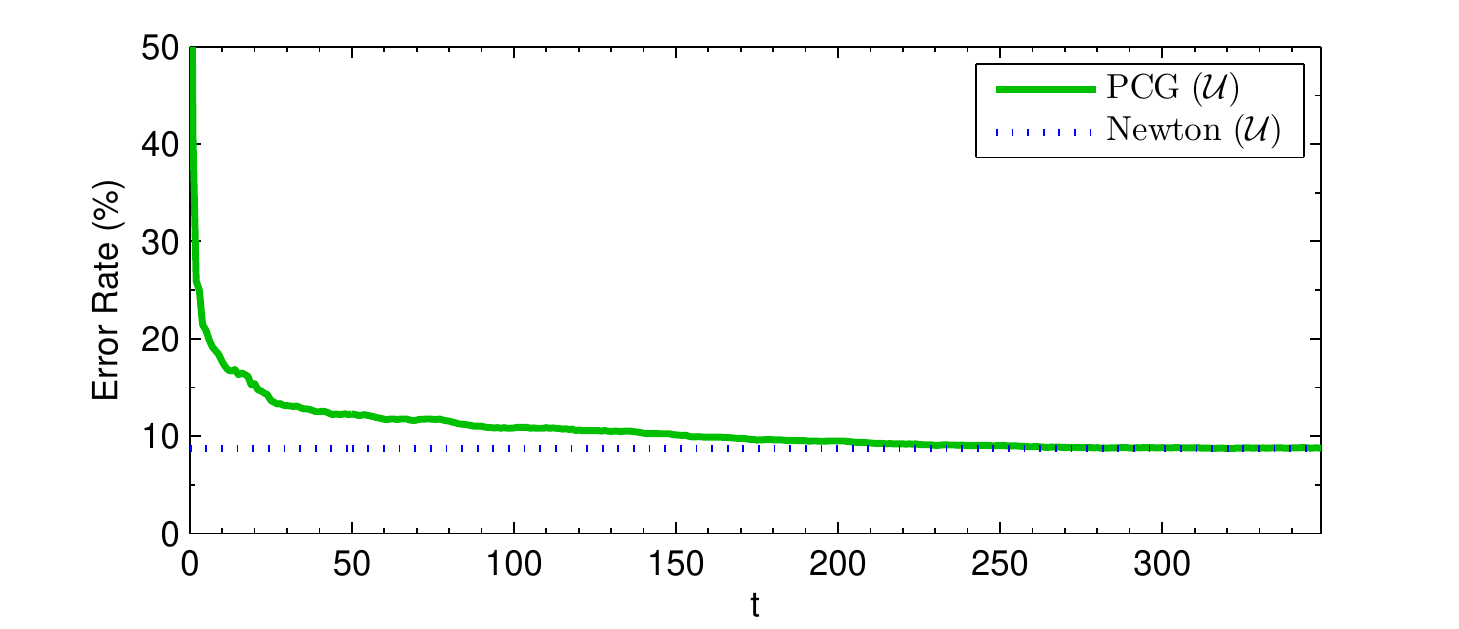}
		\end{minipage}\\				
		\hspace{-7mm}		
		\begin{minipage}{0.45\textwidth}	
	  \centering
		\includegraphics[width=1.1\textwidth]{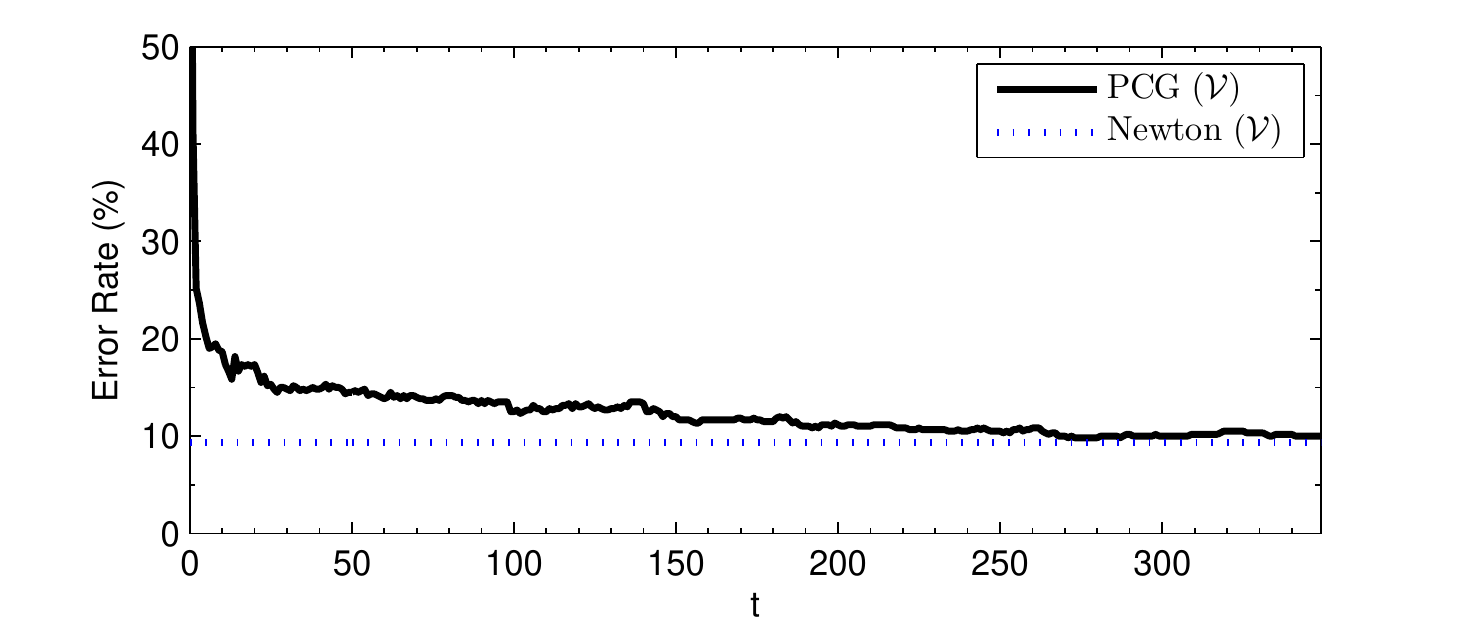}
		\end{minipage}
		\hspace{7mm}
		\begin{minipage}{0.45\textwidth}	
	  \centering
		\includegraphics[width=1.1\textwidth]{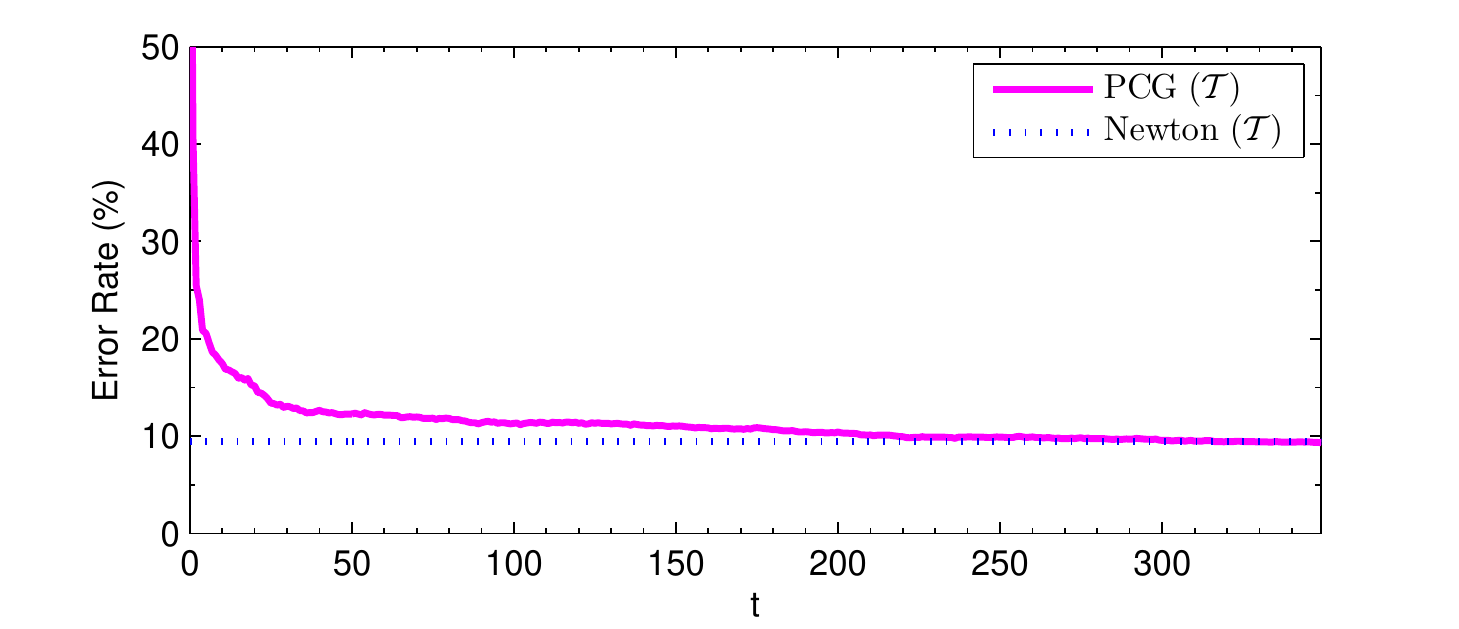}
		\end{minipage}\\
		\vspace{-2mm}							
	\caption{USPST(B) dataset: error rate on $\mathcal{L}$, $\mathcal{U}$, $\mathcal{V}$, $\mathcal{T}$ of the Laplacian SVM classifier trained in the primal by preconditioned conjugate gradient (PCG), with respect to the number of gradient steps $t$. The error rate of the primal solution computed by means of Newton's method is reported as a horizontal line.}	
	\vspace{-2mm}	
	\label{figtcg3}	
\end{figure}

\begin{figure}[ht!]
	\centering
	\hspace{-7mm}	
		\begin{minipage}{0.45\textwidth}	
	  \centering
		\includegraphics[width=1.1\textwidth]{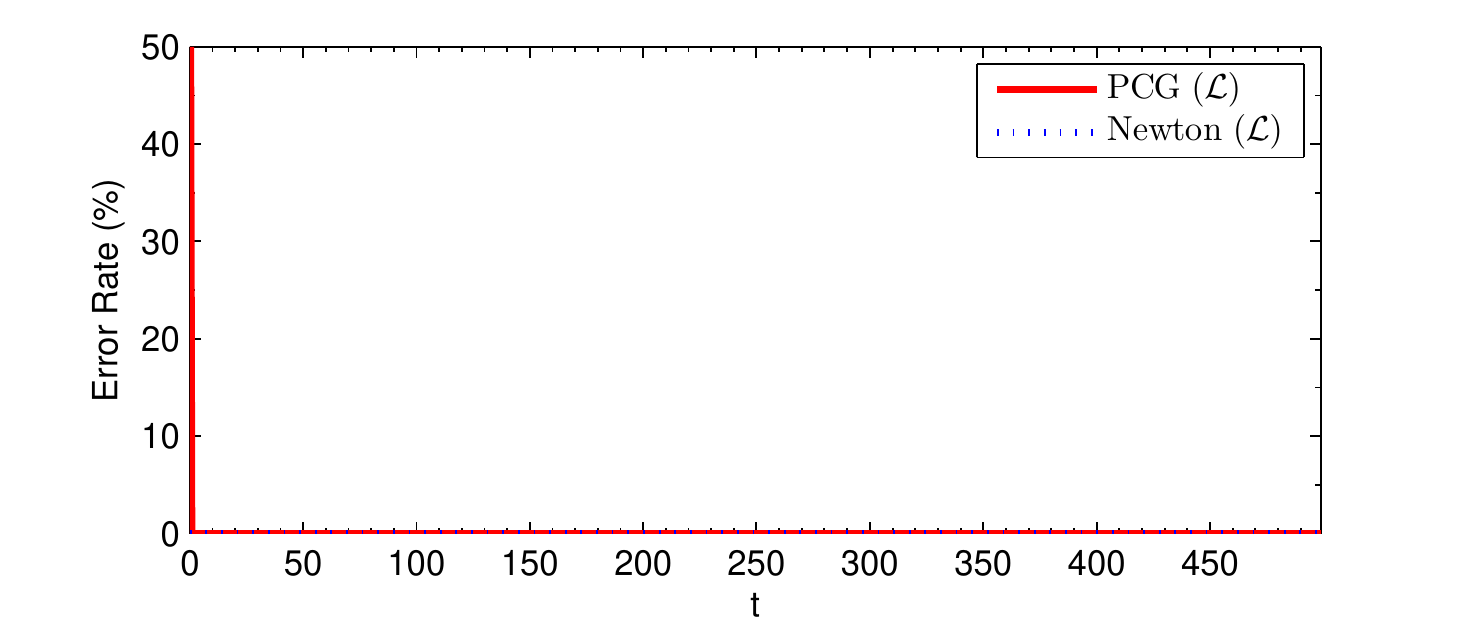}
		\end{minipage}
		\hspace{7mm}
		\begin{minipage}{0.45\textwidth}	
	  \centering
		\includegraphics[width=1.1\textwidth]{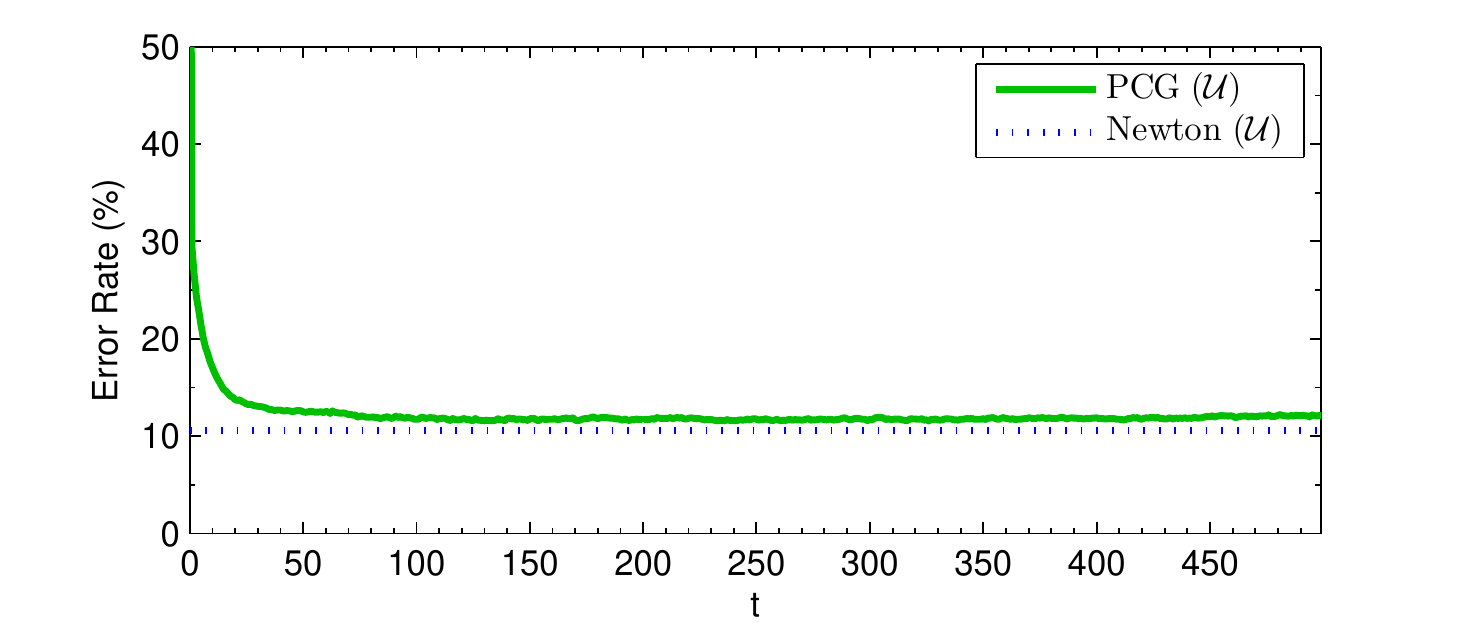}
		\end{minipage}\\		
		\hspace{-7mm}		
		\begin{minipage}{0.45\textwidth}	
	  \centering
		\includegraphics[width=1.1\textwidth]{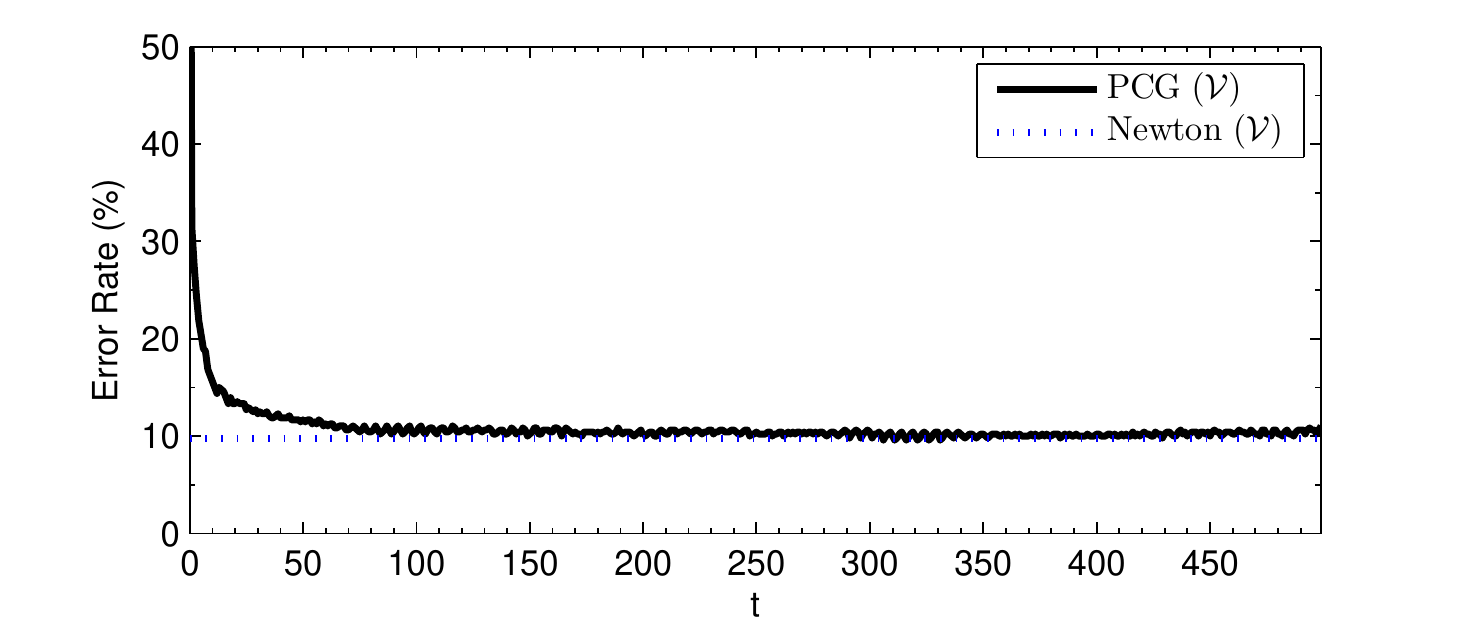}
		\end{minipage}
		\hspace{7mm}
		\begin{minipage}{0.45\textwidth}	
	  \centering
		\includegraphics[width=1.1\textwidth]{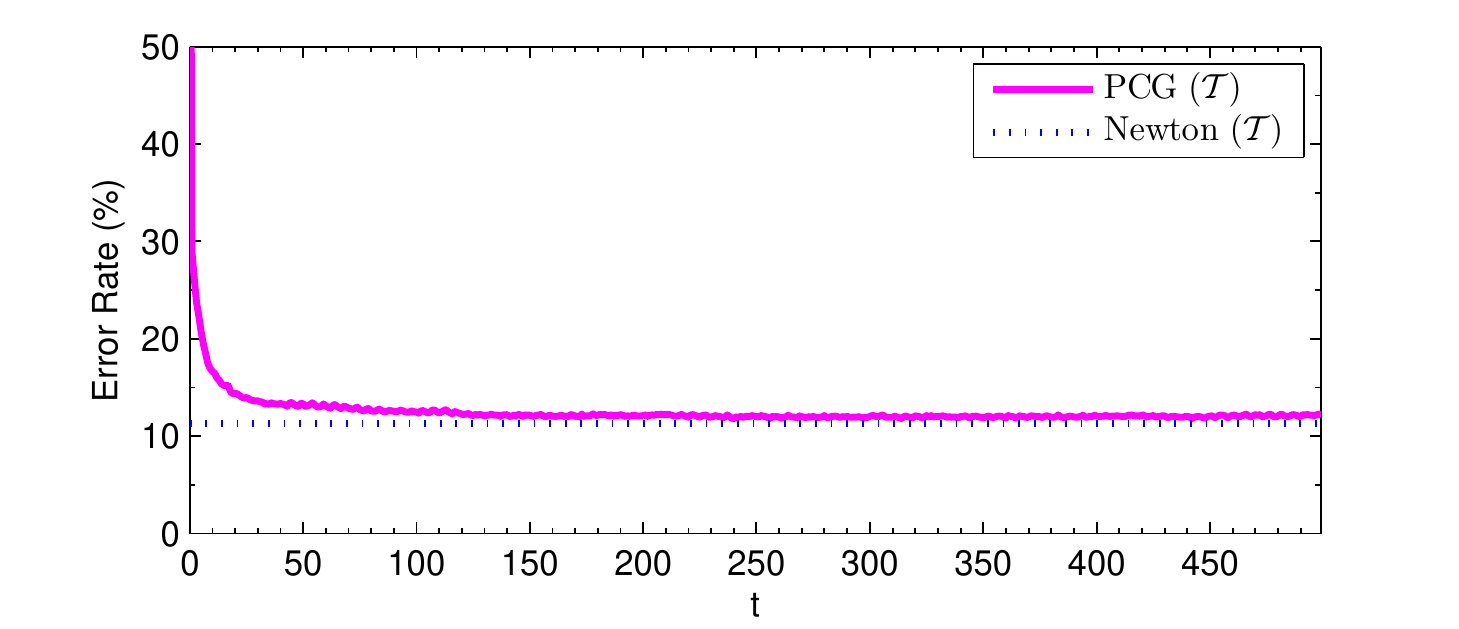}
		\end{minipage}\\
		\vspace{-2mm}						
	\caption{COIL20 dataset: error rate on $\mathcal{L}$, $\mathcal{U}$, $\mathcal{V}$, $\mathcal{T}$ of the Laplacian SVM classifier trained in the primal by preconditioned conjugate gradient (PCG), with respect to the number of gradient steps $t$. The error rate of the primal solution computed by means of Newton's method is reported as a horizontal line.}	
	\vspace{-2mm}
	\label{figtcg5}	
\end{figure}

\begin{figure}[ht!]
	\centering
	\hspace{-7mm}	
		\begin{minipage}{0.45\textwidth}	
	  \centering
		\includegraphics[width=1.1\textwidth]{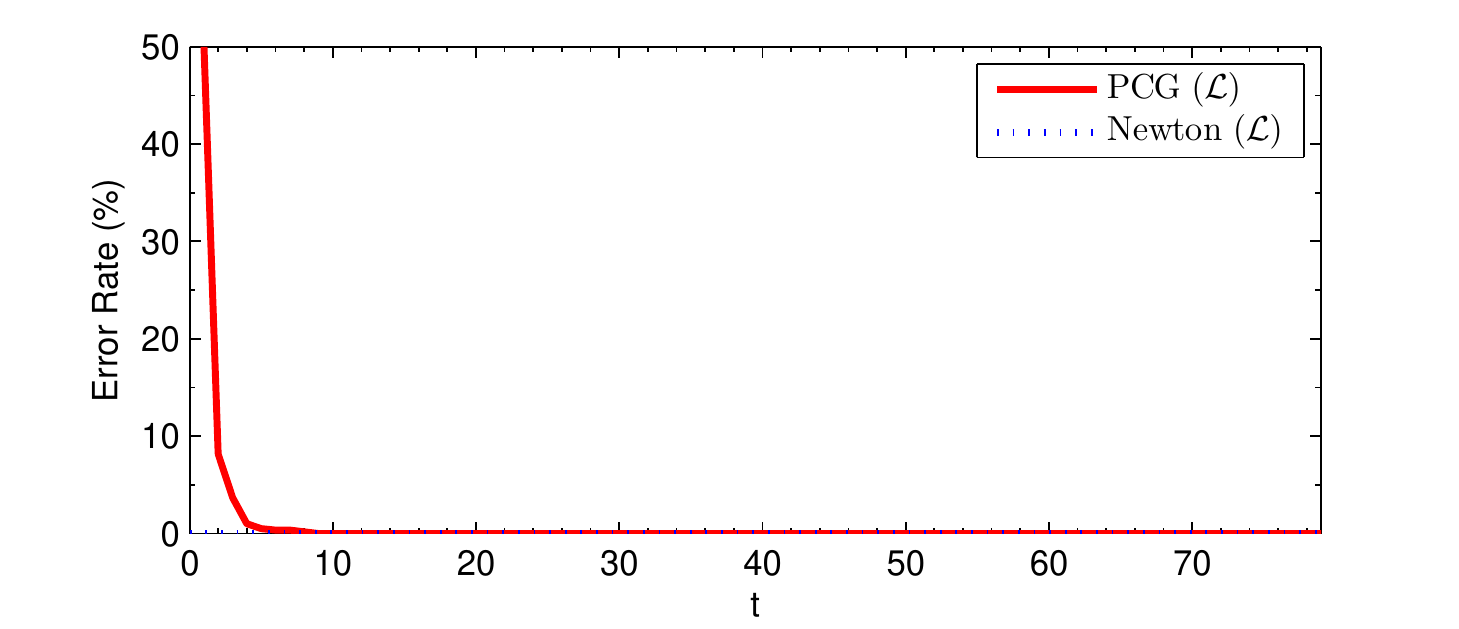}
		\end{minipage}
		\hspace{7mm}
		\begin{minipage}{0.45\textwidth}	
	  \centering
		\includegraphics[width=1.1\textwidth]{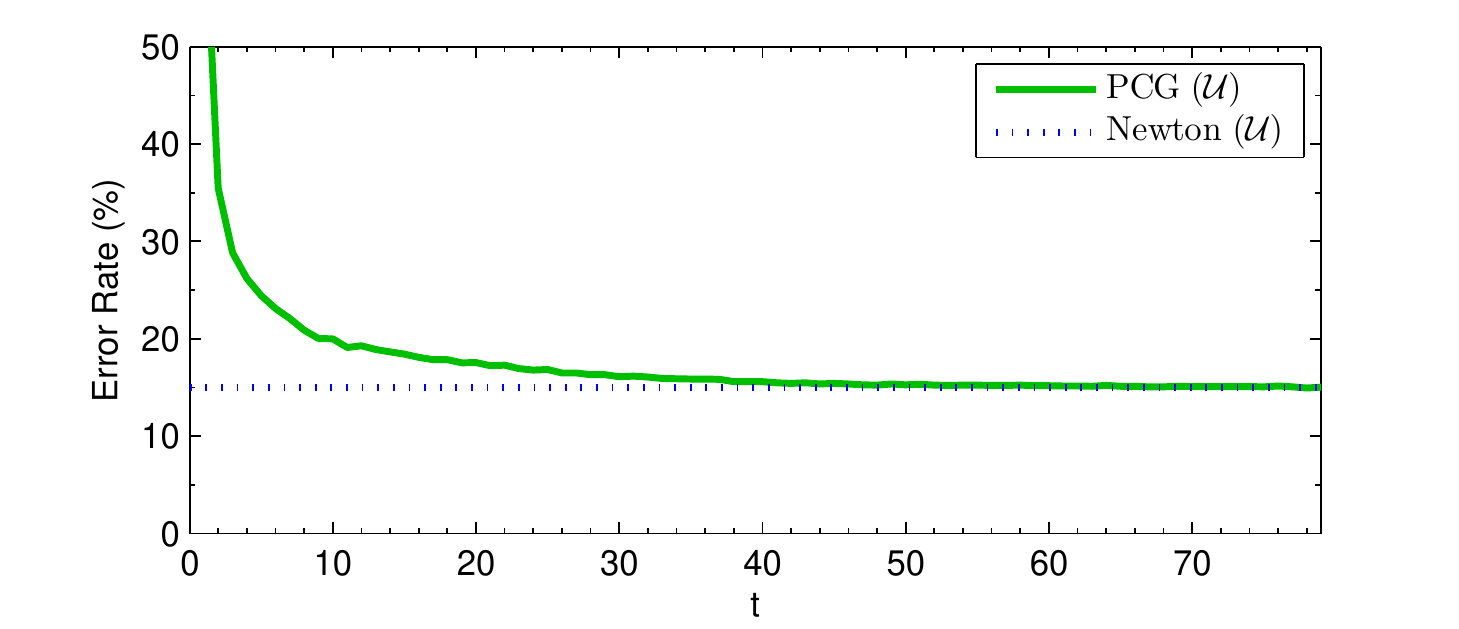}
		\end{minipage}\\			
		\hspace{-7mm}		
		\begin{minipage}{0.45\textwidth}	
	  \centering
		\includegraphics[width=1.1\textwidth]{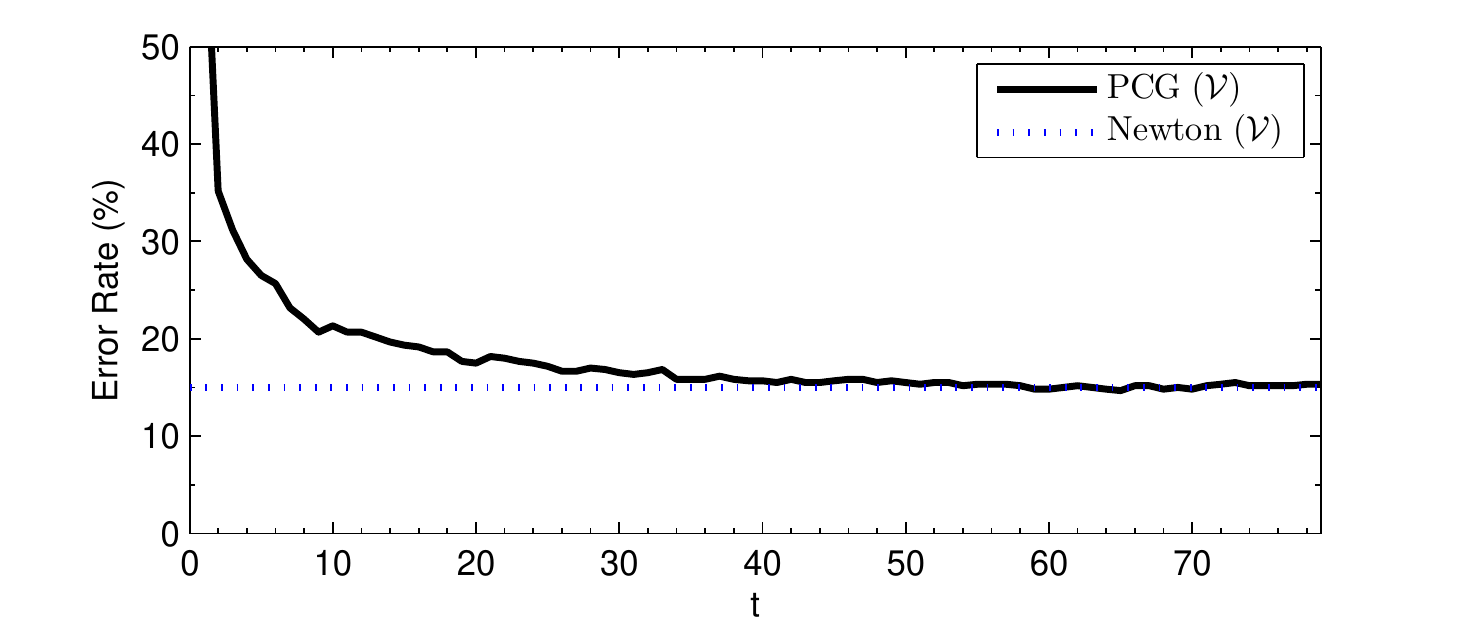}
		\end{minipage}
		\hspace{7mm}
		\begin{minipage}{0.45\textwidth}	
	  \centering
		\includegraphics[width=1.1\textwidth]{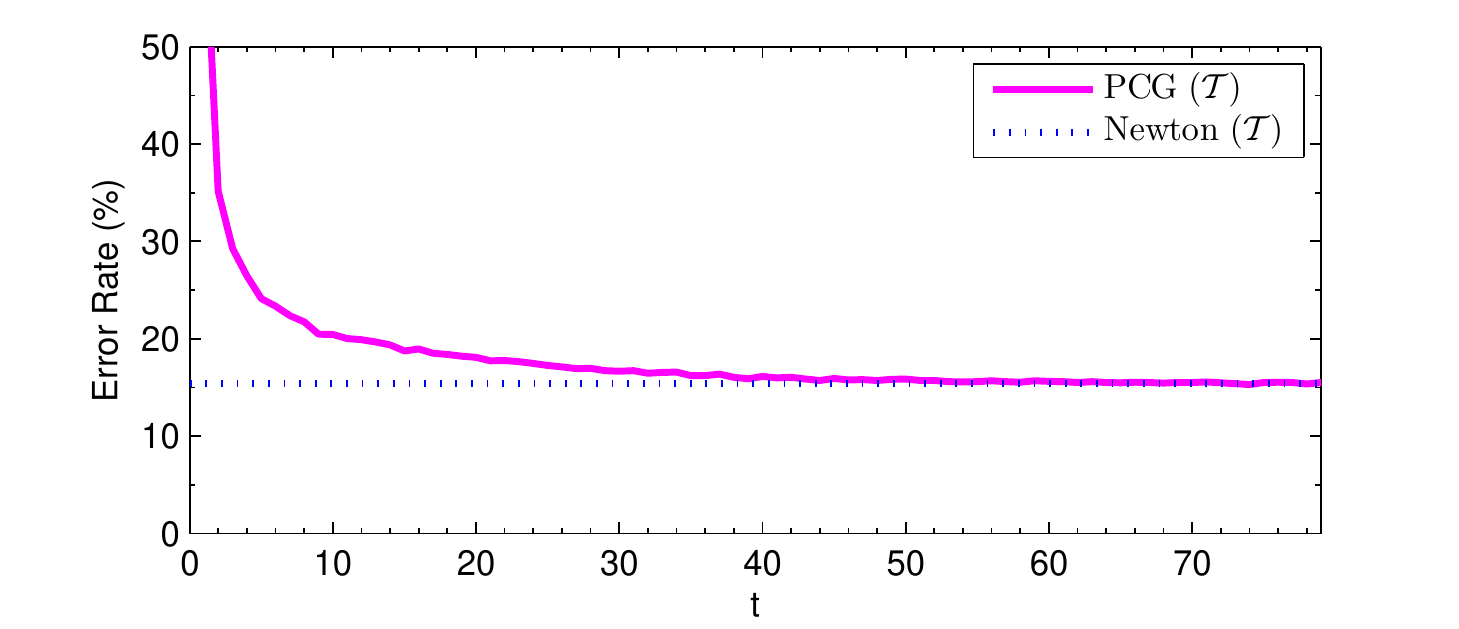}
		\end{minipage}\\	
		\vspace{-2mm}				
	\caption{USPST dataset: error rate on $\mathcal{L}$, $\mathcal{U}$, $\mathcal{V}$, $\mathcal{T}$ of the Laplacian SVM classifier trained in the primal by preconditioned conjugate gradient (PCG), with respect to the number of gradient steps $t$. The error rate of the primal solution computed by means of Newton's method is reported as a horizontal line.}	
	\vspace{-2mm}
	\label{figtcg2}	
\end{figure}

\begin{figure}[ht!]
	\centering
	\hspace{-7mm}	
		\begin{minipage}{0.45\textwidth}	
	  \centering
		\includegraphics[width=1.1\textwidth]{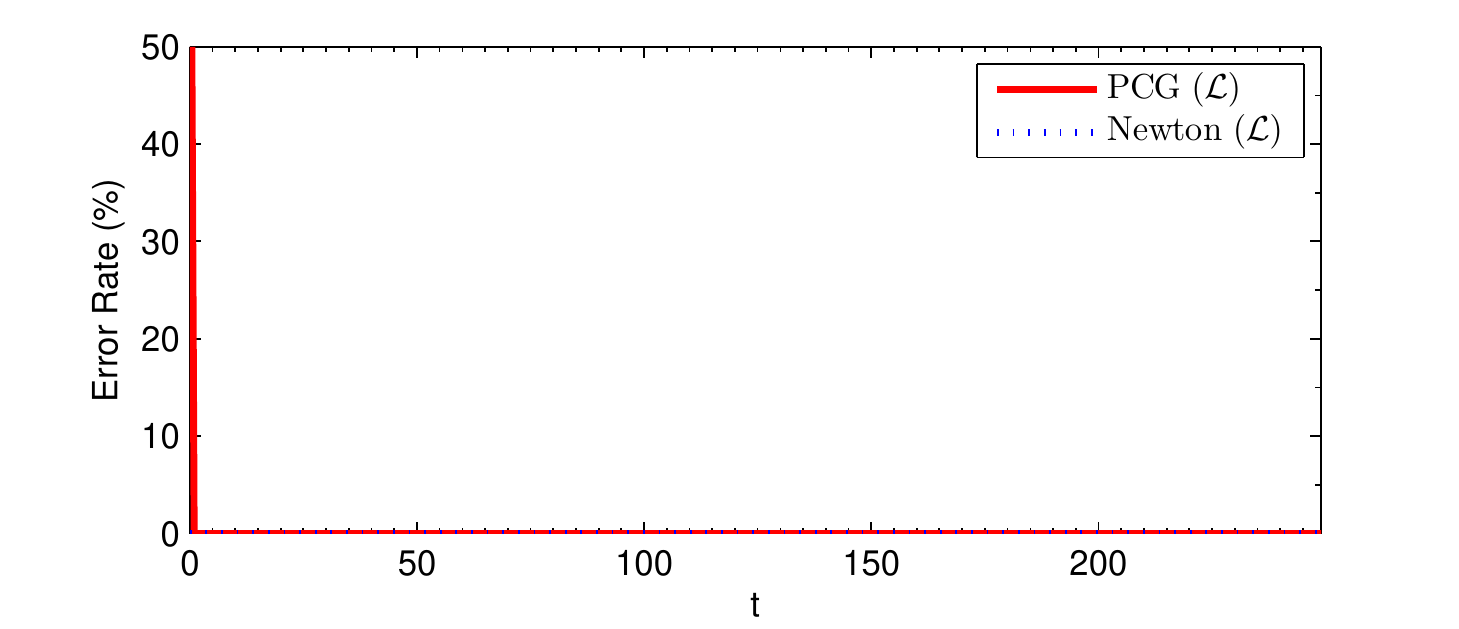}
		\end{minipage}
		\hspace{7mm}
		\begin{minipage}{0.45\textwidth}	
	  \centering
		\includegraphics[width=1.1\textwidth]{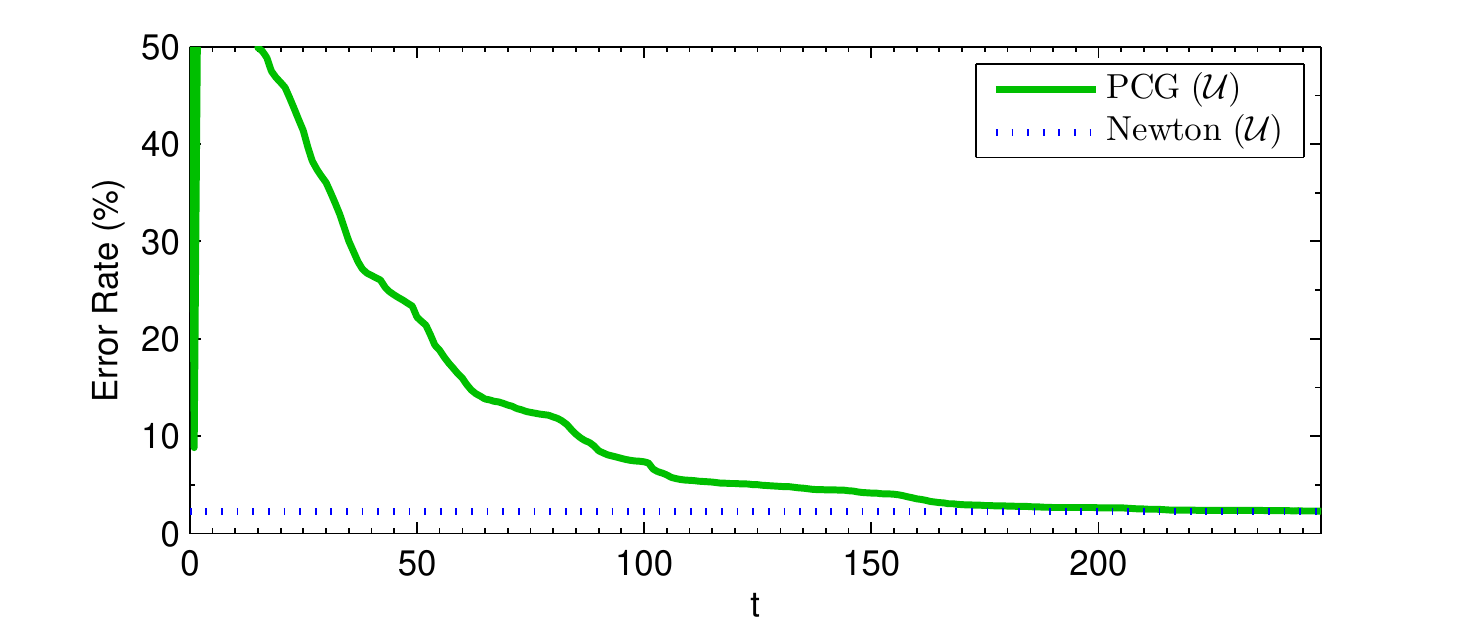}
		\end{minipage}\\			
		\hspace{-7mm}		
		\begin{minipage}{0.45\textwidth}	
	  \centering
		\includegraphics[width=1.1\textwidth]{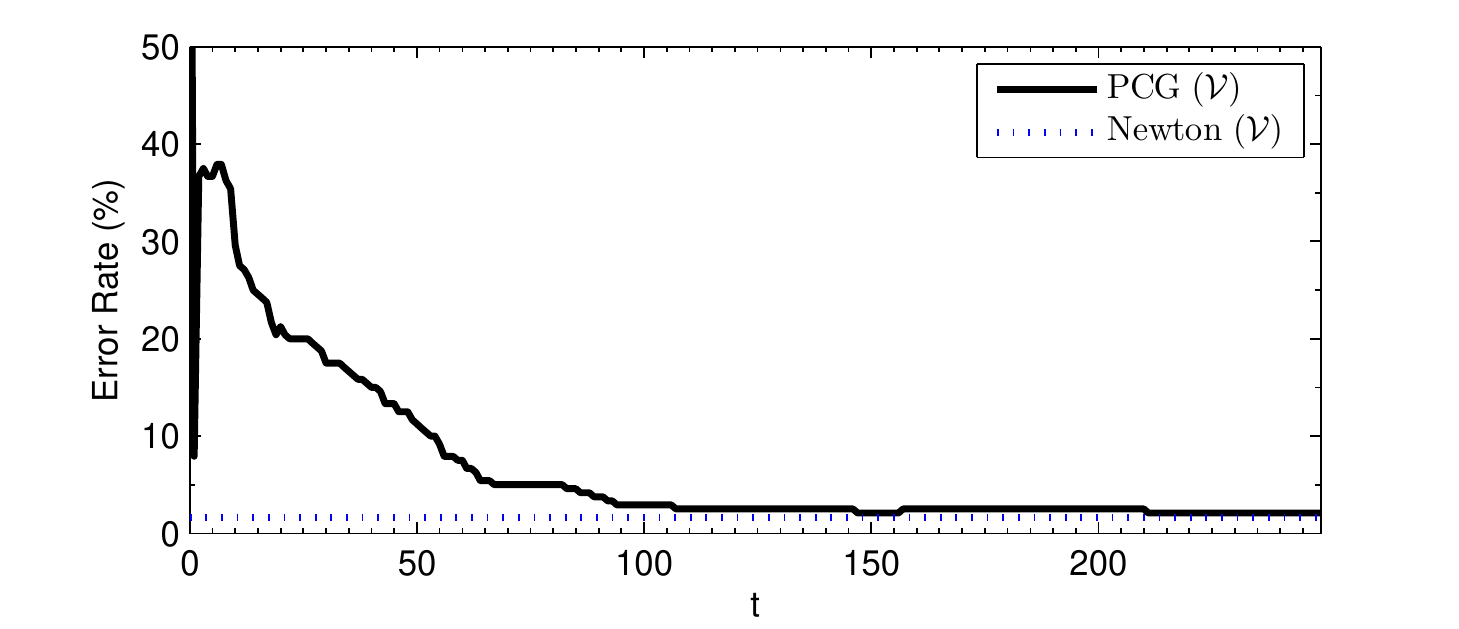}
		\end{minipage}
		\hspace{7mm}
		\begin{minipage}{0.45\textwidth}	
	  \centering
		\includegraphics[width=1.1\textwidth]{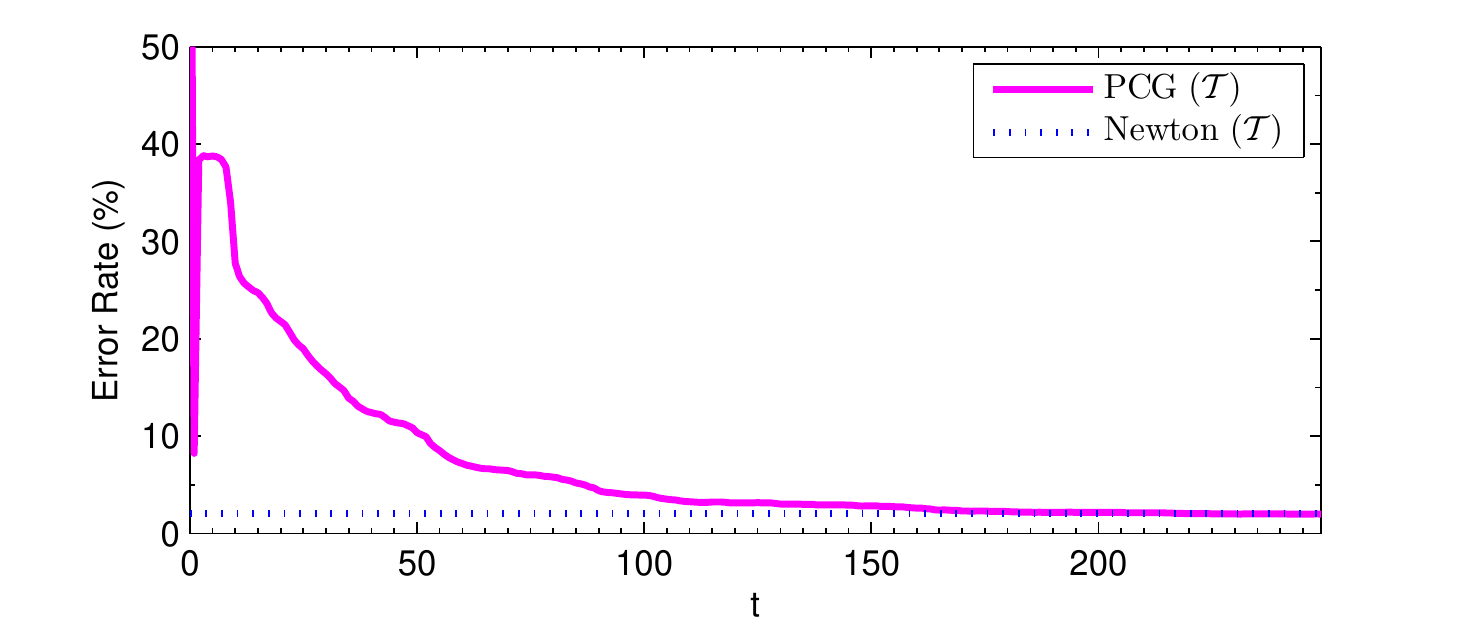}
		\end{minipage}\\
	\vspace{-2mm}					
	\caption{MNIST3VS8 dataset: error rate on $\mathcal{L}$, $\mathcal{U}$, $\mathcal{V}$, $\mathcal{T}$ of the Laplacian SVM classifier trained in the primal by preconditioned conjugate gradient (PCG), with respect to the number of gradient steps $t$. The error rate of the primal solution computed by means of Newton's method is reported as a horizontal line.}	
	\vspace{-2mm}
	\label{figtcg7}	
\end{figure}

\begin{figure}[ht!]
	\centering
	\hspace{-7mm}	
		\begin{minipage}{0.45\textwidth}	
	  \centering
		\includegraphics[width=1.1\textwidth]{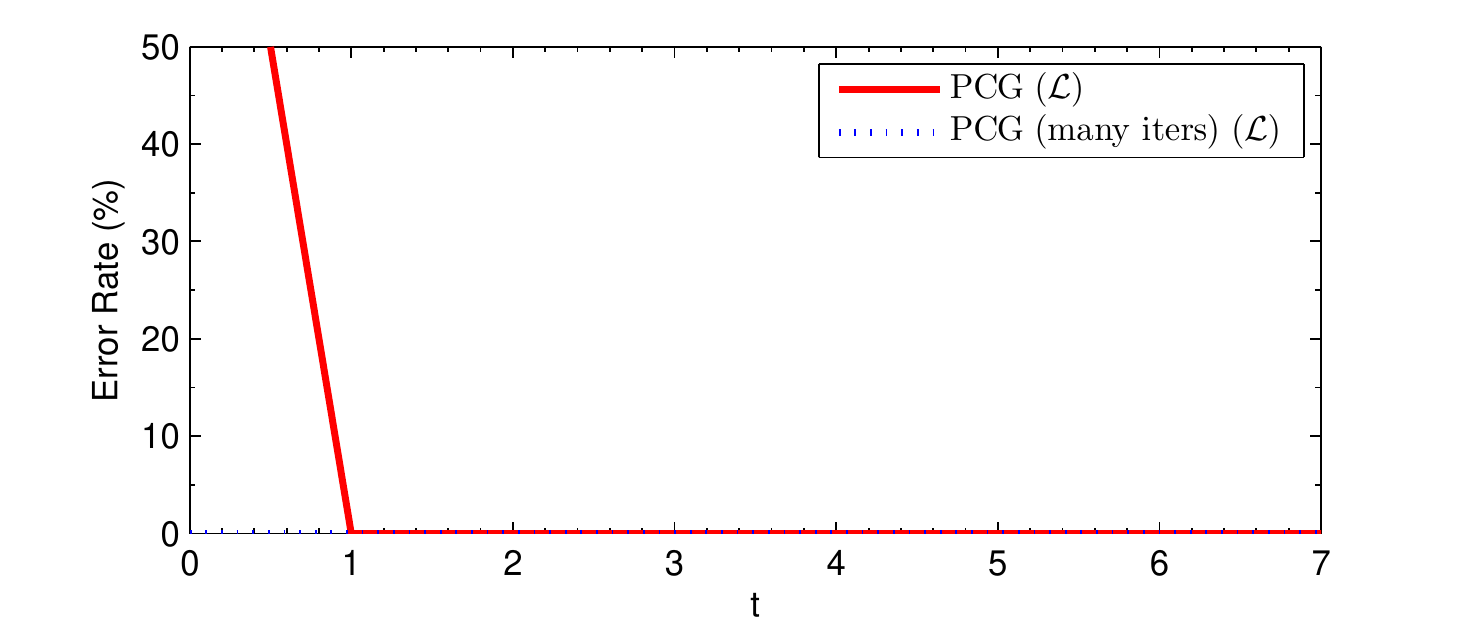}
		\end{minipage}
		\hspace{7mm}
		\begin{minipage}{0.45\textwidth}	
	  \centering
		\includegraphics[width=1.1\textwidth]{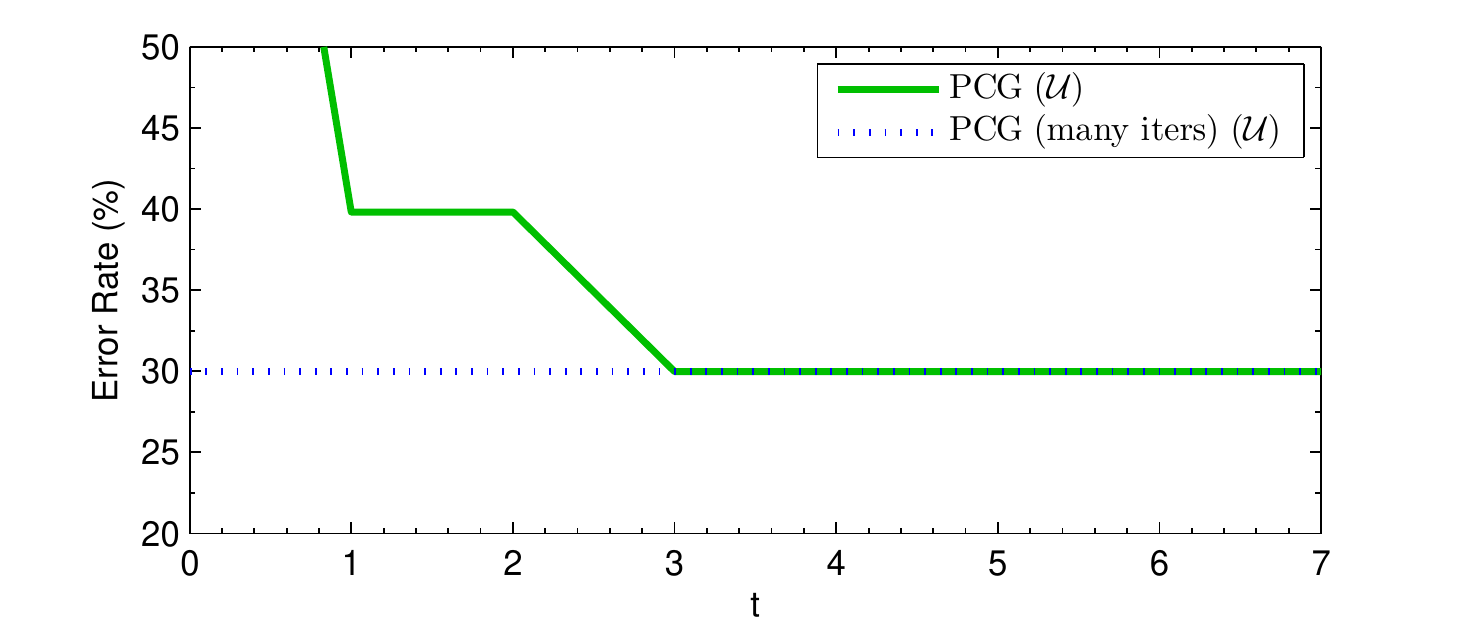}
		\end{minipage}\\			
		\hspace{-7mm}		
		\begin{minipage}{0.45\textwidth}	
	  \centering
		\includegraphics[width=1.1\textwidth]{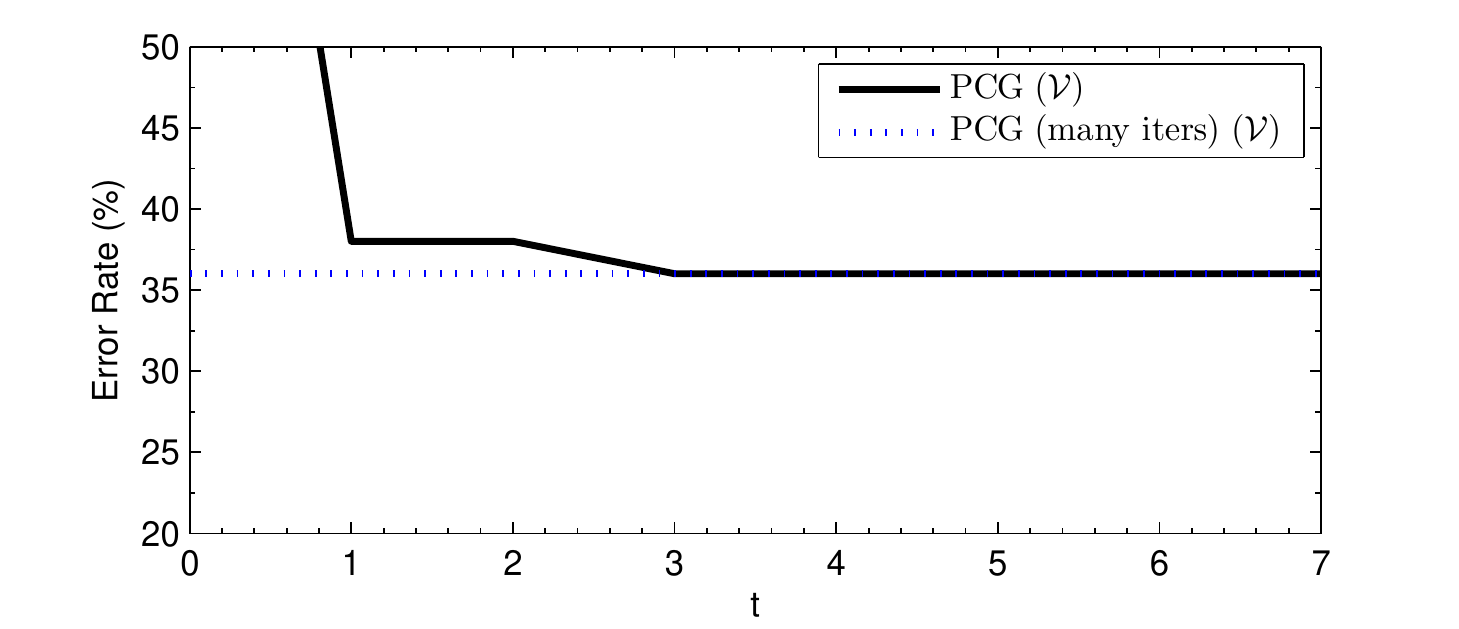}
		\end{minipage}
		\hspace{7mm}
		\begin{minipage}{0.45\textwidth}	
	  \centering
		\includegraphics[width=1.1\textwidth]{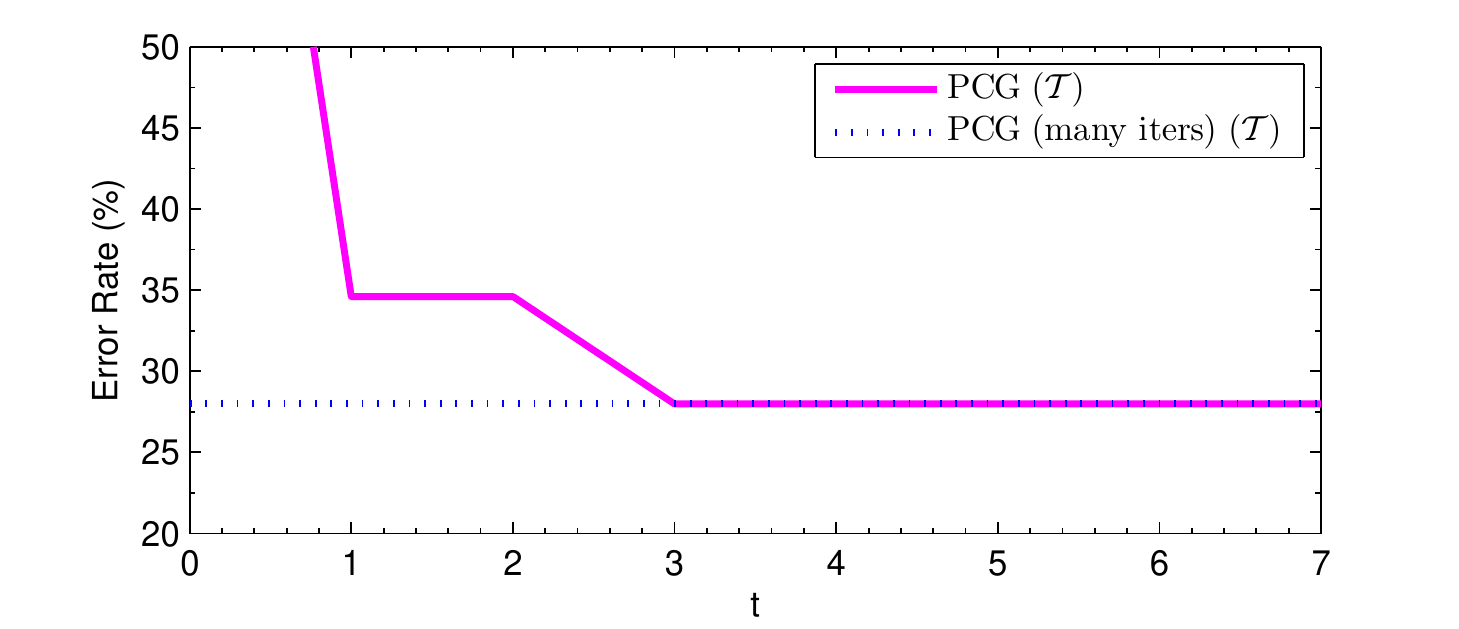}
		\end{minipage}\\
	\vspace{-2mm}					
	\caption{FACEMIT dataset: error rate on $\mathcal{L}$, $\mathcal{U}$, $\mathcal{V}$, $\mathcal{T}$ of the Laplacian SVM classifier trained in the primal by preconditioned conjugate gradient (PCG), with respect to the number of gradient steps $t$. The error rate of the primal solution computed by means of a very large set of PCG iterations is reported as a horizontal line.}	
	\vspace{-2mm}
	\label{figtcg8}	
\end{figure}

In \figurename~\ref{figtcg1}-\ref{figtcg8} we compared the error rates of LapSVMs trained in the primal by Newton's method with ones of PCG training, in function of the number of gradient steps $t$. For this comparison, $\gamma_A$ and $\gamma_I$ were selected by cross--validating with the former (see Appendix A).
The horizontal line on each graph represents the error rate of the optimal solution computed with the Newton's method. The number of iterations required to converge to a solution with the same accuracy of the optimal one is sensibly smaller than $n$. Convergence is achieved really fast, and only in the COIL20 dataset we experienced a relatively slower rate with respect to the other datasets. The error surface of each binary classifier is quite flat around optimum with the selected $\gamma_A$ and $\gamma_I$, leading to some round--off errors in gradient descent based techniques, stressed by the large number of classes and the one--against--all approach. Moreover labeled training examples are highly unbalanced. As a matter of fact, in the COIL20(B) dataset we did not experience this behavior. Finally, in the FACEMIT dataset the algorithm perfectly converges in a few iterations, showing that in this dataset the most of information is contained in the labeled data (even if it is very small), and the intrinsic constraint is easily fulfilled.

\begin{figure}[!ht]
\small
	\centering
	  \hspace{-5mm}	
		\begin{minipage}{0.45\textwidth}	
	  \centering
	  G50C\\ \vspace{1mm} \includegraphics[width=1.0\textwidth]{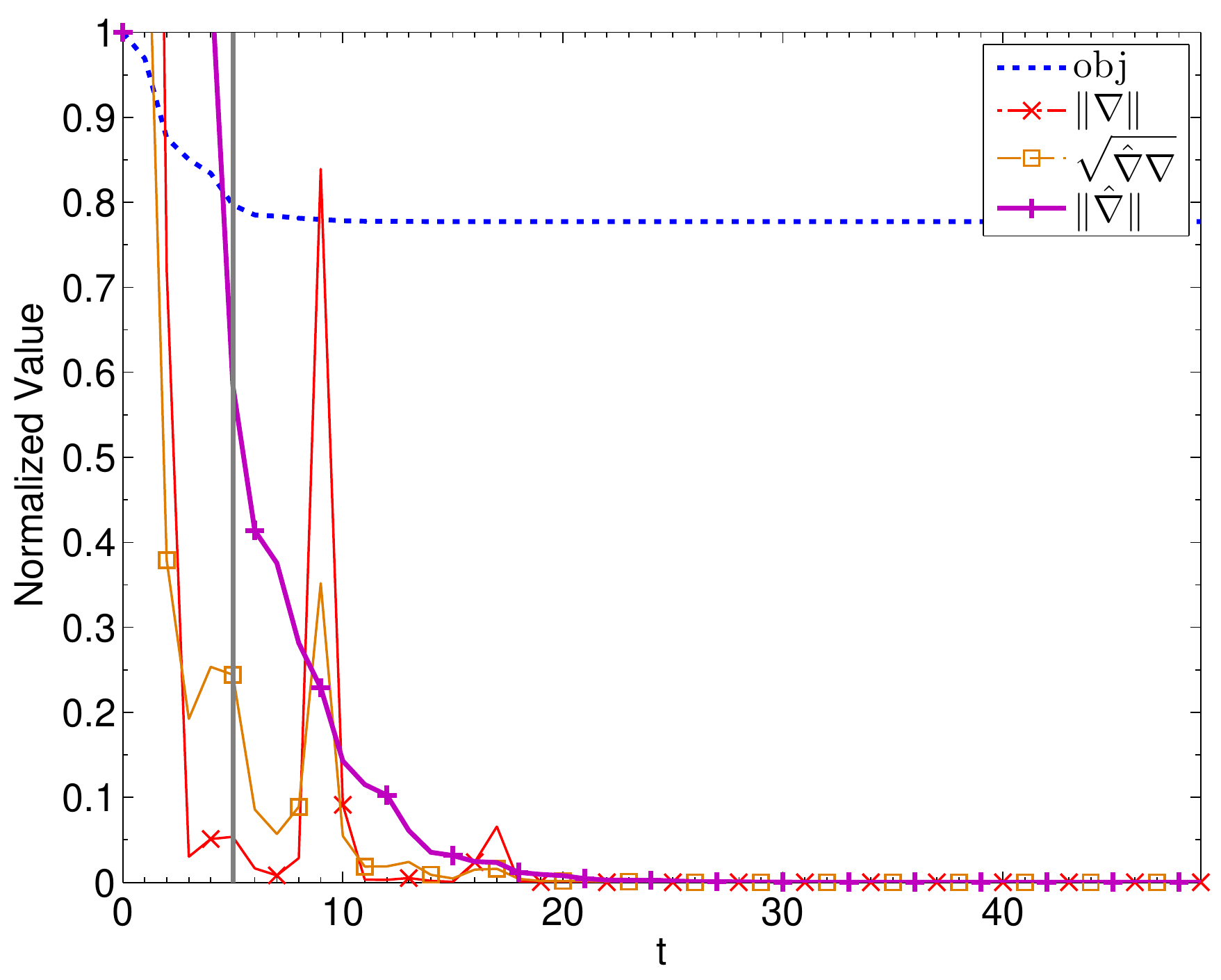}
		\end{minipage}
		\hspace{5mm}
		\begin{minipage}{0.45\textwidth}	
	  \centering
	  \vspace{-1mm}
		COIL20(B)\\ \includegraphics[width=1.0\textwidth]{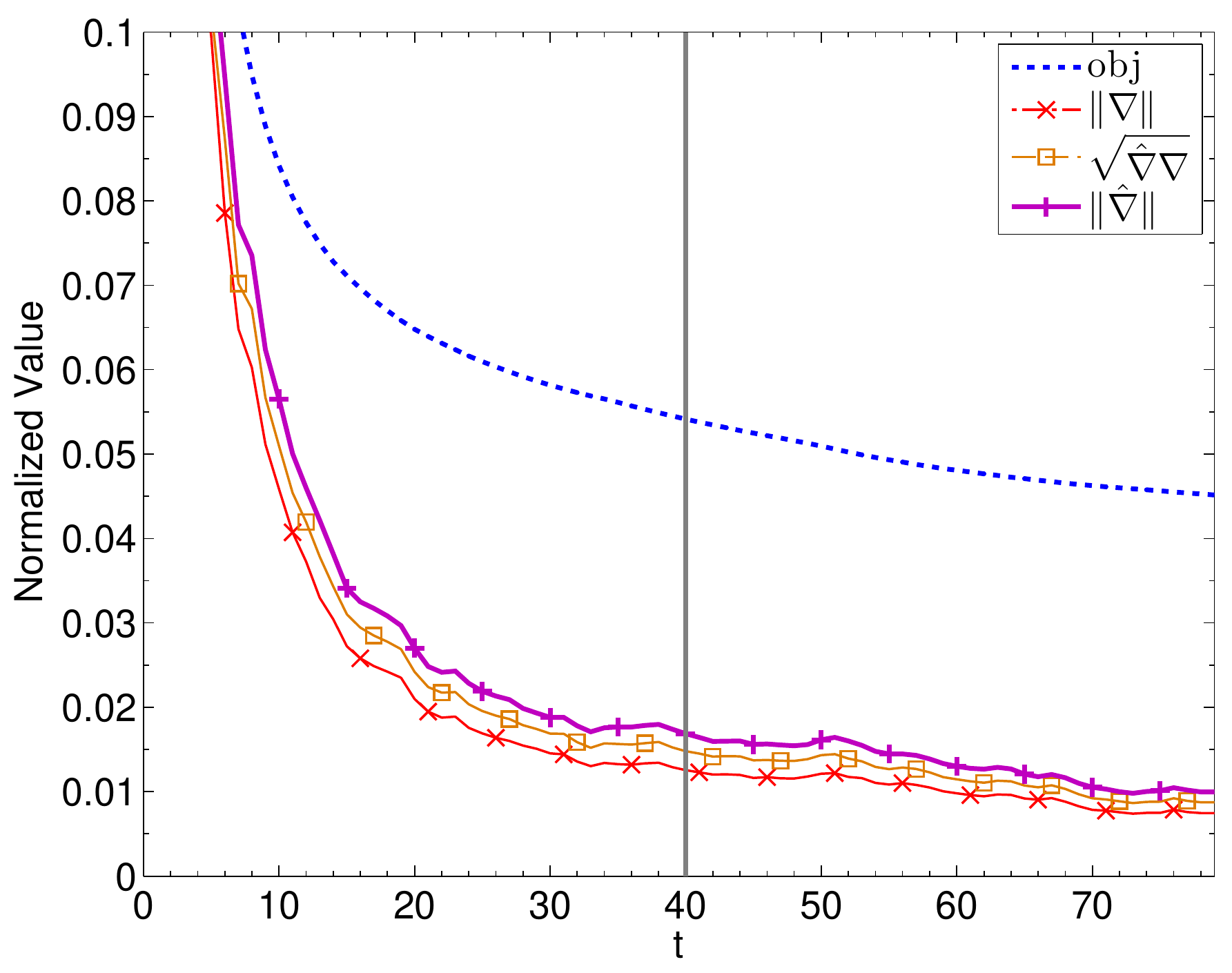} 
		\end{minipage}\\
%

	\vspace{3mm}

	\centering
		 \hspace{-5mm}
		\begin{minipage}{0.45\textwidth}			
	  \centering
		PCMAC\\ \vspace{1mm} \includegraphics[width=1.0\textwidth]{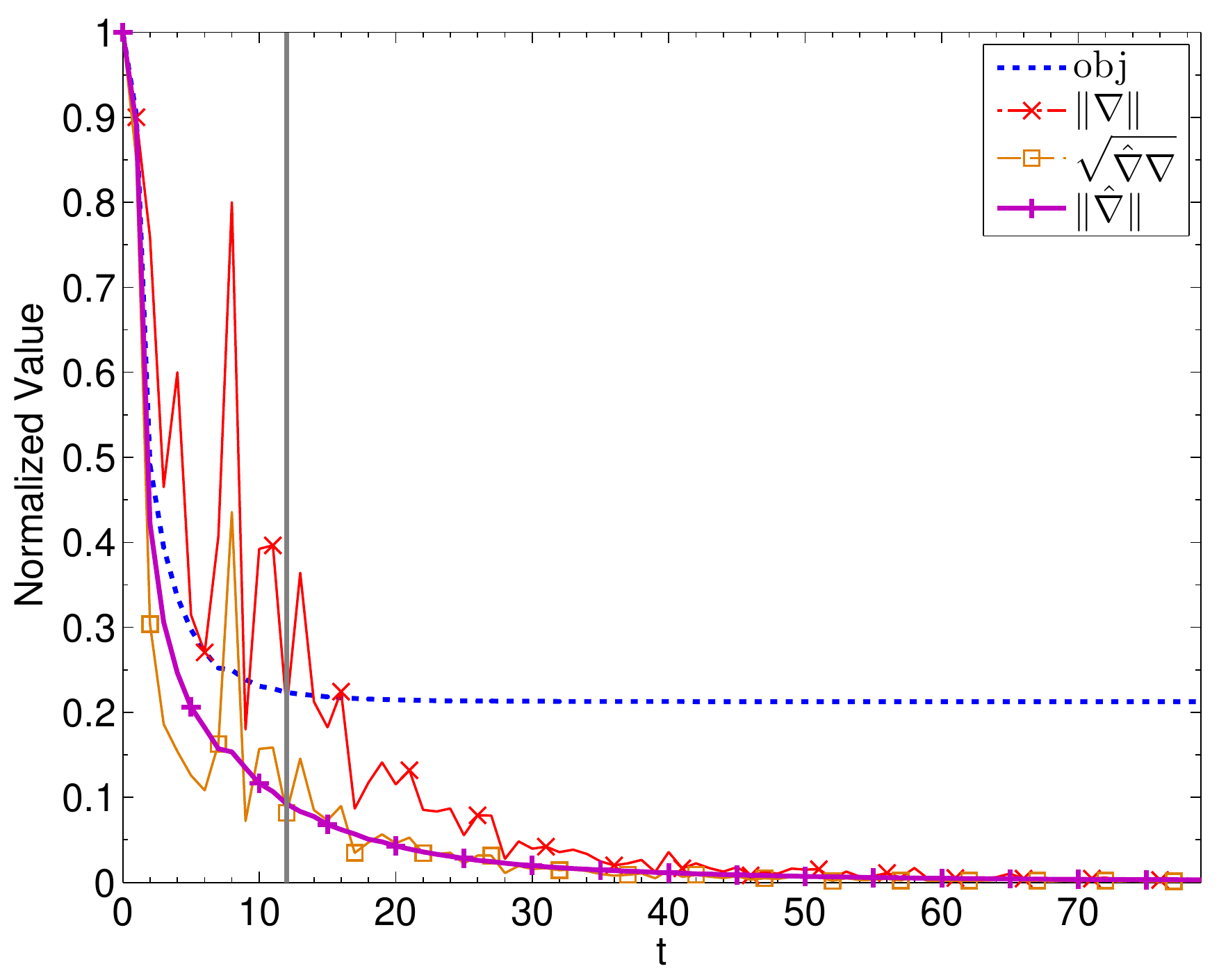}
		\end{minipage}
		\hspace{5mm}		
		\begin{minipage}{0.45\textwidth}	
	  \centering
	  \vspace{-1mm}
		USPST(B)\\ \includegraphics[width=1.01\textwidth]{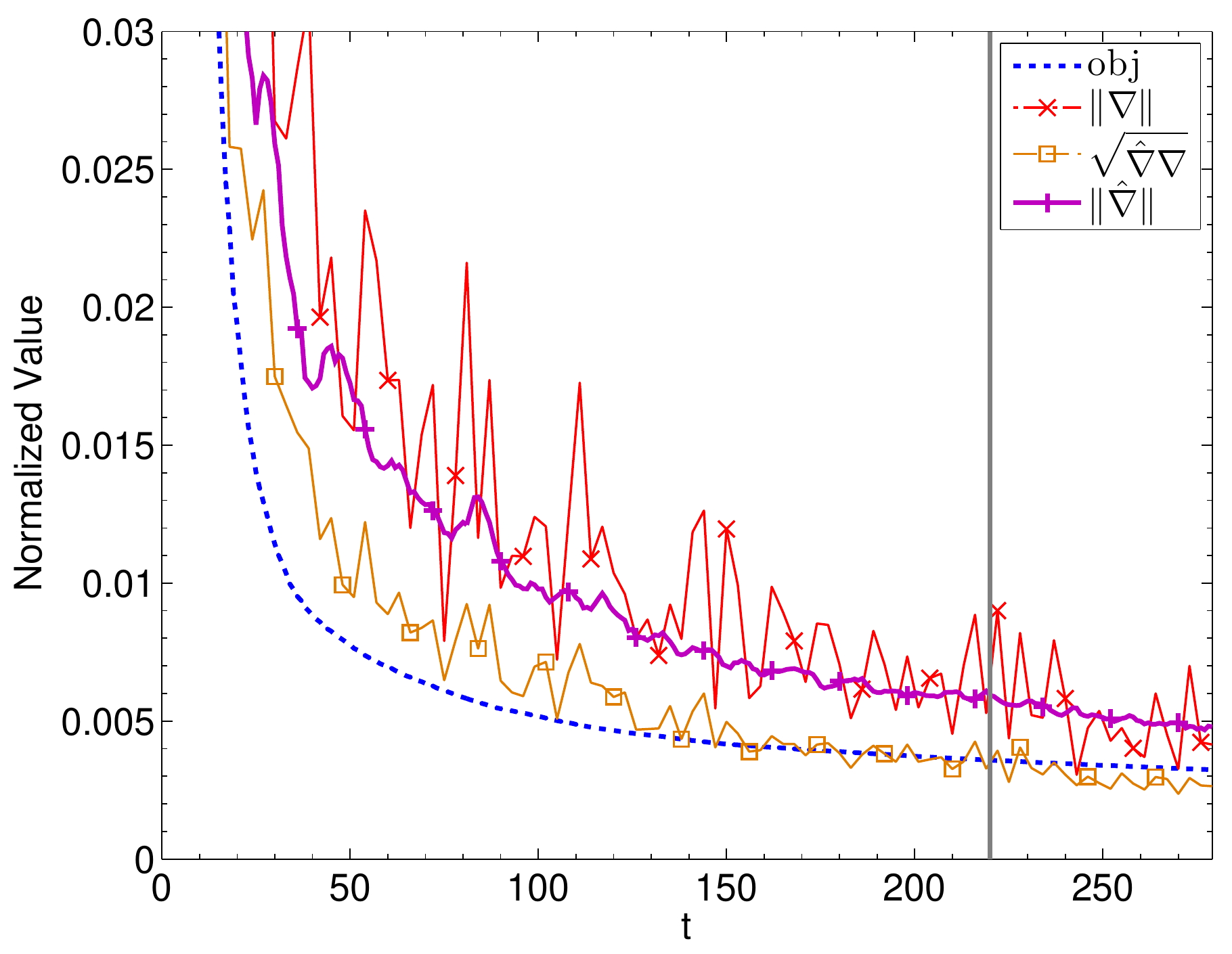}
		\end{minipage}
		\vspace{-2mm}
\caption{Details of each PCG iteration. The value of the objective function $obj$, of the gradient norm $\| \nabla \|$, of the preconditioned gradient norm $\| \hat{\nabla} \|$, and of the mixed product $\sqrt{\hat{\nabla}^T\nabla}$ are displayed in function of the number of PCG iterations ($t$). The vertical line represents the number of iterations after which the error rate is roughly the same of the one at the optimal solution.}
\label{fig:tcgnorm2}
\end{figure}

\begin{figure}[!ht]
\small
	\centering
		 \hspace{-5mm}
		\begin{minipage}{0.45\textwidth}	
	  \centering
		COIL20\\ \vspace{1mm} \includegraphics[width=1.022\textwidth]{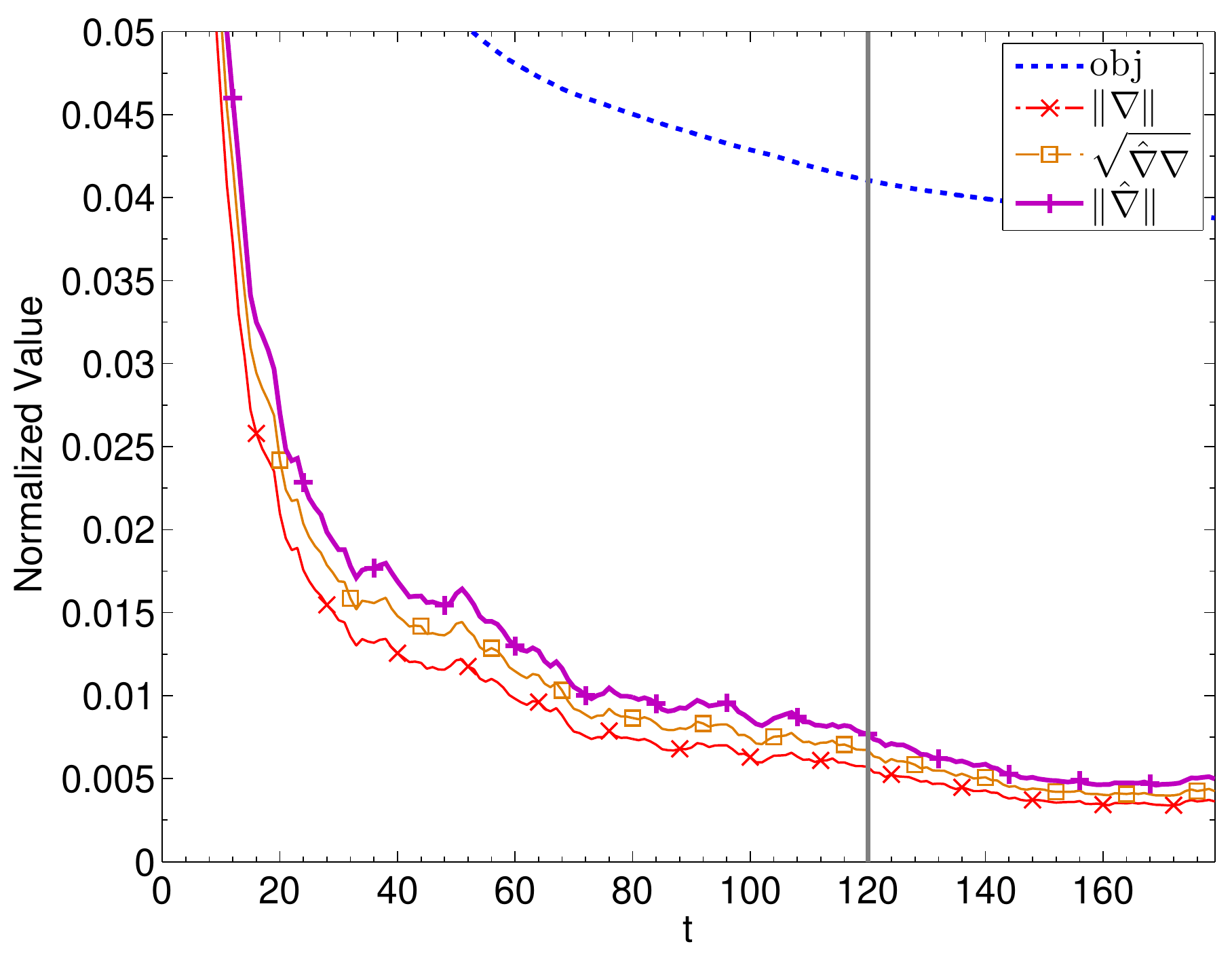}
		\end{minipage}
		\hspace{5mm}
		\begin{minipage}{0.45\textwidth}	
	  \centering
	  \vspace{-1mm}
		USPST\\ \includegraphics[width=1.0\textwidth]{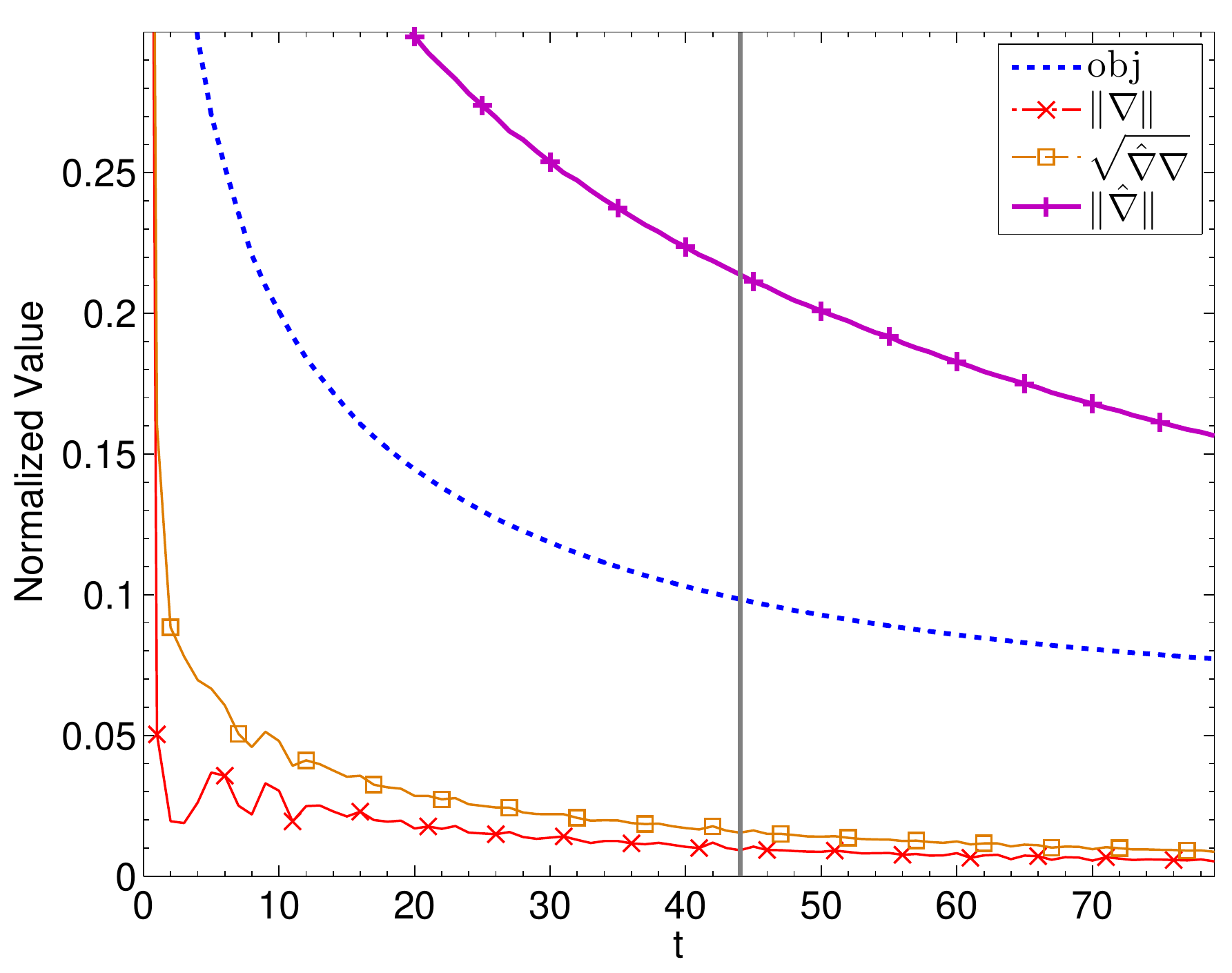}
		\end{minipage}\\
%

	\vspace{3mm}
	
	\centering
		 \hspace{-1mm}
		\begin{minipage}{0.45\textwidth}
	  \centering
		MNIST3VS8\\ \vspace{1.5mm} \includegraphics[width=1.022\textwidth]{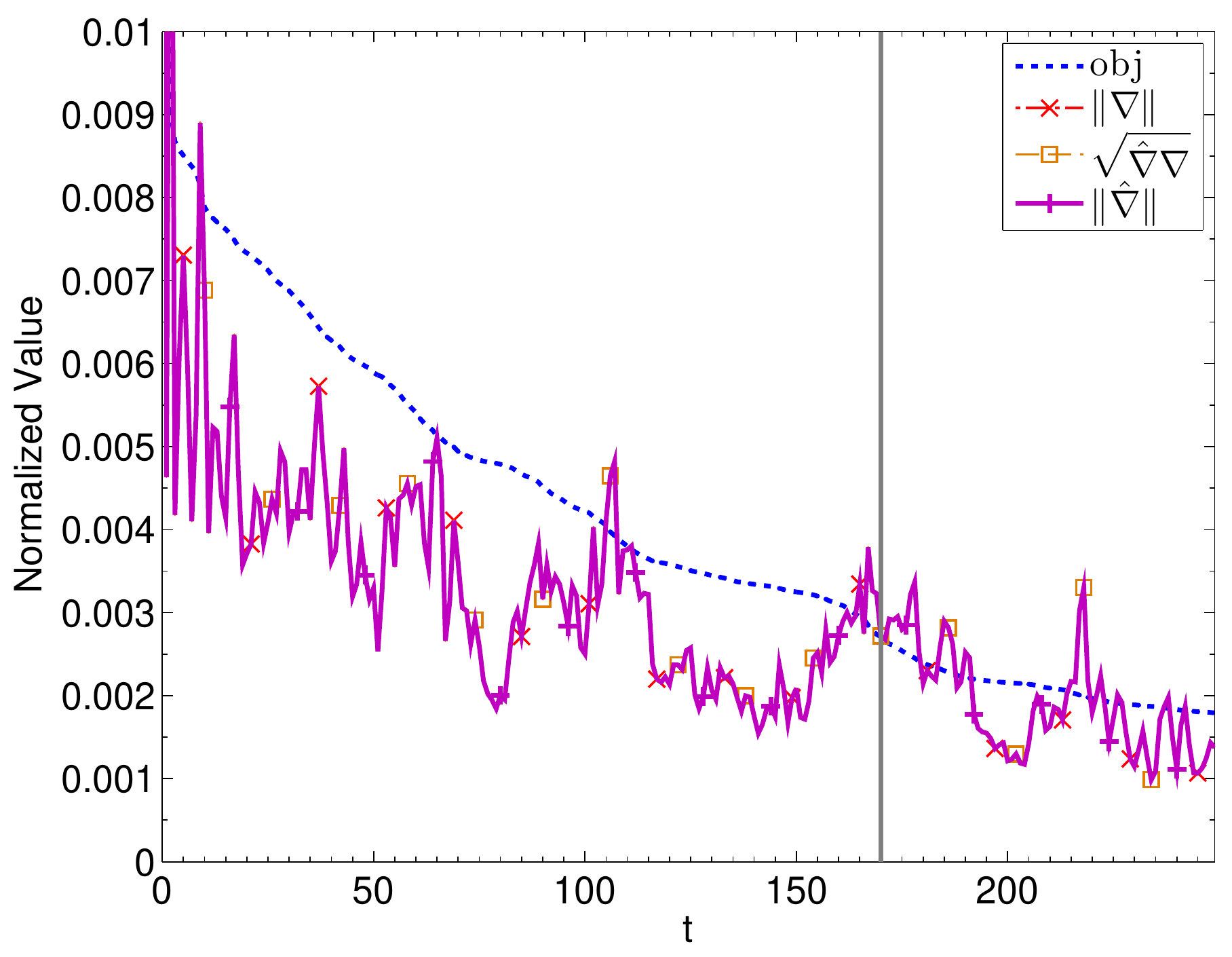}
		\end{minipage}
		\hspace{6mm}
		\begin{minipage}{0.45\textwidth}	
	  \centering
	  \vspace{-1mm}
	  FACEMIT\\ \includegraphics[width=1.0\textwidth]{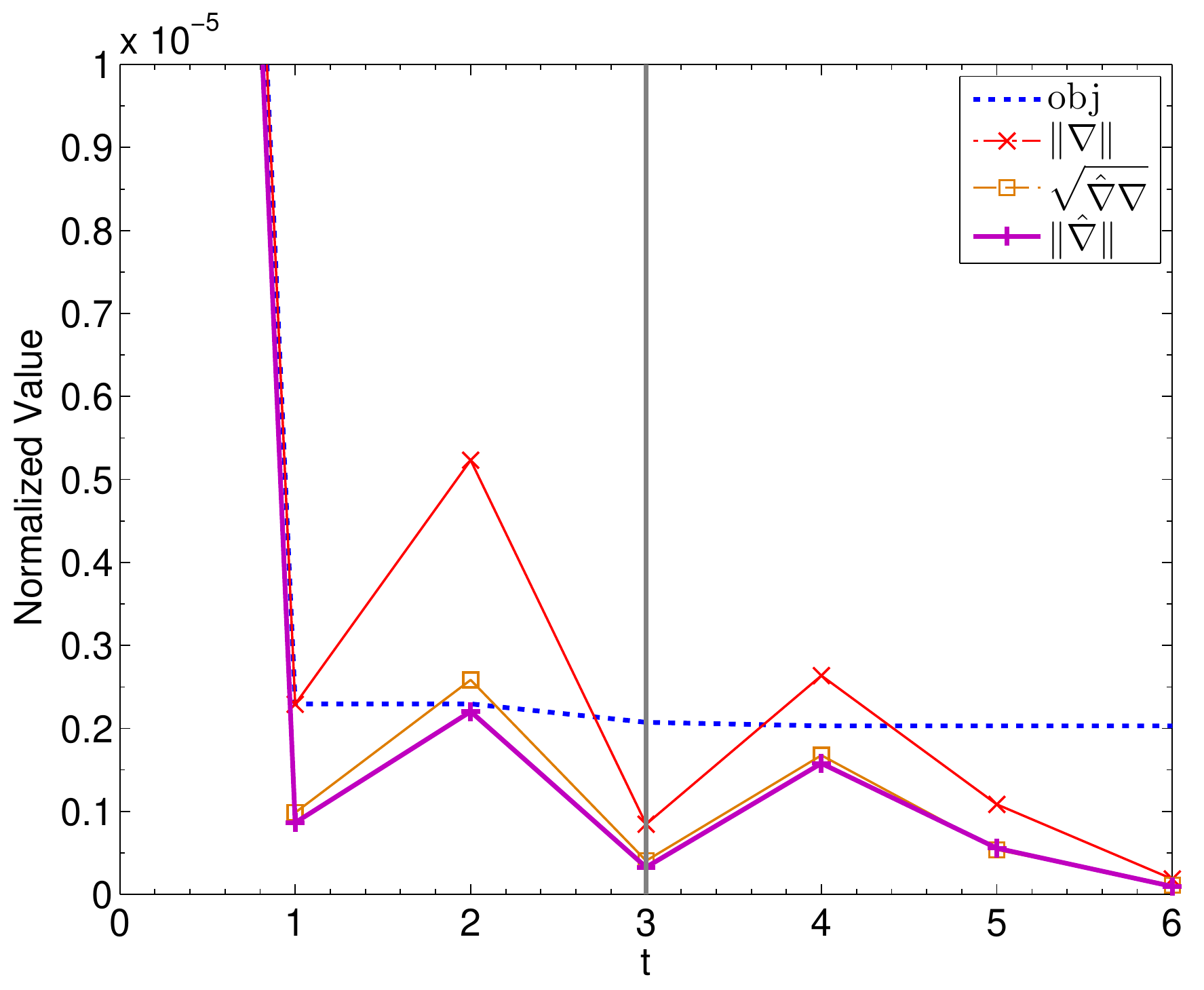} 
		\end{minipage}
		\vspace{-2mm}		 
\caption{Details of each PCG iteration. The value of the objective function $obj$, of the gradient norm $\| \nabla \|$, of the preconditioned gradient norm $\| \hat{\nabla} \|$, and of the mixed product $\sqrt{\hat{\nabla}^T\nabla}$ are displayed in function of the number of PCG iterations ($t$). The vertical line represents the number of iterations after which the error rate is roughly the same of the one at the optimal solution.}
\label{fig:tcgnorm4}	
\end{figure}

In \figurename~\ref{fig:tcgnorm2}-\ref{fig:tcgnorm4} we collected the values of the gradient norm $\| \nabla \|$, of the preconditioned gradient norm $\| \hat{\nabla} \|$, of the mixed product $\sqrt{\hat{\nabla}^T\nabla}$, and of the objective function $obj$ for each dataset, normalized by their respective values at $t=0$. The vertical line is an indicative index of the number of iterations after which the error rate on all partitions ($\mathcal{L}$, $\mathcal{U}$, $\mathcal{V}$, $\mathcal{T}$) becomes equal to the one at the optimal solution.
The curves generally keep sensibly decreasing even after such line, without reflecting real improvements in the classifier accuracy, and they differ by orders of magnitude among the considered dataset, showing their strong problem dependency (differently from our proposed conditions). As described in \sectionname~\ref{sec:approx}, we can see how it is clearly impossible to define a generic threshold on them to appropriately stop the PCG descent (i.e. to find a good trade--off between number of iterations and accuracy). Moreover, altering the values of the classifier parameters can sensibly change the shape of the error function, requiring a different threshold every time. In those datasets where points keep entering and leaving the $\mathcal{E}$ set as $t$ increases (mainly during the first steps) the norm of the gradient can show an instable behavior between consecutive iterations, due to the piecewise nature of the problem, making the threshold selection task ulteriorly complex. This is the case of the PCMAC and USPST(B) dataset. In the MNIST data, the elements of kernel matrix non belonging to the main diagonal are very small due to the high degree of the polynomial kernel, so that the gradient and the preconditioned gradient are close. 

%

Using the proposed PCG goal conditions (\sectionname~\ref{sec:approx}), we cross--validated the primal LapSVM classifier trained by PCG, and the selected parameters are reported in \tablename~\ref{tab:parampcg} of Appendix A. In the USPST(B), COIL20(B), and MNIST3VS8 datasets, larger values for $\ga$ or $\gi$ are selected by the validation process, since the convergence speed of PCG is enhanced. In the other datasets, parameter values remain substantially the same of the ones selected by solving with the Newton's method, suggesting that a reliable and fast cross--validation can be performed with PCG and the proposed early stopping heuristics. 

In \tablename~\ref{tabtimepcg1} the training times, the number of PCG and line search iterations are collected, whereas in \tablename~\ref{tabrespcg2} the corresponding classification error rates are reported, for a comparison with the optimal solution computed using Newton's method. As already stressed, the training times appreciably drop down when training a LapSVM in the primal using PCG and our goal conditions, independently by the dataset. Early stopping allows us to obtain results comparable to the Newton's method or to the original two step dual formulation, showing a direct correlation between the proposed goal conditions and the quality of the classifier. Moreover, our conditions are the same for each problem or dataset, overcoming all the issues of the previously described ones. In the COIL20 dataset we can observe performances less close to the one of the solution computed with Newton's method. This is due to the already addressed motivations, and it also suggests that the stopping condition should probably be checked while training in parallel the 20 binary classifiers, instead of separately checking it on each of them. A better tuning of the goal conditions or a different formulation of them can move the accuracy closer to the one of primal LapSVM trained with Newton's method, but it goes beyond to the scope of this paper.

The number of PCG iterations is noticeably smaller than $n$. Obviously it is function of the gap between each checking of a stopping criterion, that we set to $\sqrt{n}/2$. The number of iterations from the stability check is sometimes larger that the one from the validation check (COIL20(B), USPST, COIL20). As a matter of fact, labeled validation data is more informative than a stable, but unknown, decision on the unlabeled one. On the other hand validation data could not represent test data enough accurately. Using a mixed strategy makes sense in those cases, as can be observed in the COIL20 dataset. In our experiments the mixed criterion has generally the same behavior of the most strict of the two heuristics for each specific set of data. In the FACEMIT dataset complete convergence is achieved in just a few iterations, independently by the heuristics. The number of line search iterations is usually very small and negligible with respect to the computational cost of the training algorithm.

\begin{table}
\footnotesize
\centering
\begin{tabular}{lllll}
\hline\\[-3.8mm]
\textbf{Dataset}&\textbf{Laplacian SVM}&\textbf{Training Time}&\textbf{PCG Iters}&\textbf{LS Iters}\\
\hline\\[-3.4mm]
\multirow{5}{*}{G50C} 
 &Dual&0.155 (0.004)&-&-\\
 &Newton&0.134 (0.006)&-&-\\ 
 & \multirow{1}{*}{PCG [Stability Check]}
&\textbf{0.044} (0.006)&20 (0)& 1 (0)\\ 
 & \multirow{1}{*}{PCG [Validation Check]}
&\textbf{0.043} (0.006)&20.83 (2.89)& 1 (0)\\ 
 & \multirow{1}{*}{PCG [Mixed Check]}
&\textbf{0.044} (0.006)&20.83 (2.89)& 1 (0)\\ 
\hline\\[-3.1mm]
\multirow{5}{*}{COIL20(B)}
 &Dual&0.311 (0.012)&-&-\\
 &Newton&0.367 (0.097)&-&-\\
 & \multirow{1}{*}{PCG [Stability Check]}
&\textbf{0.198} (0.074)&74.67 (28.4)& 2.41 (1.83)\\ 

& \multirow{1}{*}{PCG [Validation Check]}
&\textbf{0.097} (0.026)&37.33 (10.42)& 1 (0)\\ 
 & \multirow{1}{*}{PCG [Mixed Check]}
&\textbf{0.206} (0.089)&78.67 (34.42)& 2.38 (1.79)\\ 
\hline\\[-3.1mm]
\multirow{5}{*}{PCMAC} 
 &Dual&14.8203 (0.104)&-&-\\
 &Newton&15.756 (0.285)&-&-\\ 
 & \multirow{1}{*}{PCG [Stability Check]}
&\textbf{1.897} (0.040)&38.00 (0)& 1.16 (0.45)\\ 
 & \multirow{1}{*}{PCG [Validation Check]}
&\textbf{1.967} (0.269)&39.58 (5.48)& 1.15 (0.44)\\ 
 & \multirow{1}{*}{PCG [Mixed Check]}
&\textbf{4.610} (1.602)&91.83 (32.24)& 3.70 (3.09)\\  
\hline\\[-3.1mm]
\multirow{5}{*}{USPST(B)} 
 &Dual&1.196 (0.015)&-&-\\
 &Newton&1.4727 (0.2033)&-&-\\
 & \multirow{1}{*}{PCG [Stability Check]}
&\textbf{0.300} (0.030)&58.58 (5.48)& 1.74 (0.90)\\ 
 & \multirow{1}{*}{PCG [Validation Check]}
&\textbf{0.281} (0.086)&55.42 (17.11)& 1.68 (0.90)\\ 
 & \multirow{1}{*}{PCG [Mixed Check]}
&\textbf{0.324} (0.059)&63.33 (12.38)& 1.70 (0.89)\\ 
\hline\\[-3.1mm]
\multirow{5}{*}{COIL20} 
 &Dual&6.321 (0.441)&-&-\\
 &Newton&7.26 (1.921)&-&-\\
 & \multirow{1}{*}{PCG [Stability Check]}
&\textbf{3.297} (1.471)&65.47 (30.35)& 2.53 (1.90)\\ 
 & \multirow{1}{*}{PCG [Validation Check]}
&\textbf{1.769} (0.299)&34.07 (6.12)& 3.37 (2.22)\\ 
 & \multirow{1}{*}{PCG [Mixed Check]}
&\textbf{3.487} (1.734)&69.53 (35.86)& 2.48 (1.87)\\ 
\hline\\[-3.1mm]
\multirow{5}{*}{USPST} 
 &Dual&12.25 (0.2)&-&-\\
 &Newton&17.74 (2.44)&-&-\\   
 & \multirow{1}{*}{PCG [Stability Check]}
&\textbf{1.953} (0.403)&41.17 (8.65)& 3.11 (1.73)\\ 
 & \multirow{1}{*}{PCG [Validation Check]}
&\textbf{2.032} (0.434)&42.91 (9.38)& 3.13 (1.73)\\ 
 & \multirow{1}{*}{PCG [Mixed Check]}
&\textbf{2.158} (0.535)&45.60 (11.66)& 3.12 (1.72)\\ 
\hline\\[-3.1mm]
\multirow{5}{*}{MNIST3VS8} 
 &Dual&2064.18 (3.1)&-&-\\
 &Newton&2824.174 (105.07)&-&-\\   
 & \multirow{1}{*}{PCG [Stability Check]}
&\textbf{114.441} (0.235)&110 (0)& 5.58 (2.79)\\
 & \multirow{1}{*}{PCG [Validation Check]}
&\textbf{124.69} (0.335)&110 (0)& 5.58 (2.79)\\ 
 & \multirow{1}{*}{PCG [Mixed Check]}
 &\textbf{124.974} (0.414)&110 (0)& 5.58 (2.79)\\
\hline\\[-3.1mm]
\multirow{3}{*}{FACEMIT}  
 & \multirow{1}{*}{PCG [Stability Check]}
&\textbf{35.728} (0.868)&3 (0)& 1 (0)\\
 & \multirow{1}{*}{PCG [Validation Check]}
&\textbf{35.728} (0.868)&3 (0)& 1 (0)\\
 & \multirow{1}{*}{PCG [Mixed Check]}
&\textbf{35.728} (0.868)&3 (0)& 1 (0)\\
\hline
\end{tabular}
\caption{Training time comparison among the Laplacian SVMs trained in the dual (Dual), LapSVM trained in the primal by means of Newton's method (Newton) and by means of preconditioned conjugate gradient (PCG) with the proposed early stopping conditions (in square brackets). Average training times (in seconds) and their standard deviations, the number of PCG iterations, and of Line Search (LS) iterations (per each PCG one) are reported.}
\label{tabtimepcg1}
\end{table}

\begin{table}
\footnotesize
\centering
\begin{tabular}{lllll}
\hline\\[-3.8mm]
\textbf{Dataset}&\textbf{Laplacian SVM}&\textbf{$\mathcal{U}$}&\textbf{$\mathcal{V}$}&\textbf{$\mathcal{T}$}\\
\hline\\[-3.4mm]
\multirow{4}{*}{G50C} 
 &Newton&6.16 (1.48)&6.17 (3.46)&7.27 (2.87)\\ 
 & \multirow{1}{*}{PCG [Stability Check]}
& 6.13 (1.46)& 6.17 (3.46)& 7.27 (2.87)\\ 
 & \multirow{1}{*}{PCG [Validation Check]}
& 6.16 (1.48)& 6.17 (3.46)& 7.27 (2.87)\\ 
 & \multirow{1}{*}{PCG [Mixed Check]}
& 6.16 (1.48)& 6.17 (3.46)& 7.27 (2.87)\\ 
\hline\\[-3.1mm]
\multirow{4}{*}{COIL20(B)}
 &Newton&8.16 (2.04)&7.92 (3.96)&8.56 (1.9)\\ 
 & \multirow{1}{*}{PCG [Stability Check]}
& 8.81 (2.23)& 8.13 (3.71)& 8.84 (1.93)\\ 
& \multirow{1}{*}{PCG [Validation Check]}
& 8.32 (2.28)& 8.96 (4.05)& 8.45 (1.58)\\ 
 & \multirow{1}{*}{PCG [Mixed Check]}
& 8.84 (2.28)& 8.13 (3.71)& 8.84 (1.96)\\ 
\hline\\[-3.1mm]
\multirow{4}{*}{PCMAC} 
 &Newton&9.68 (0.77)&7.83 (4.04)&9.37 (1.51)\\ 
 & \multirow{1}{*}{PCG [Stability Check]}
& 9.65 (0.78)& 7.83 (4.04)& 9.42 (1.50)\\ 
 & \multirow{1}{*}{PCG [Validation Check]}
& 9.67 (0.76)& 7.83 (4.04)& 9.40 (1.50)\\ 
 & \multirow{1}{*}{PCG [Mixed Check]}
& 9.79 (0.72)& 7.67 (3.80)& 9.42 (1.28)\\ 
\hline\\[-3.1mm]
\multirow{4}{*}{USPST(B)} 
 &Newton&8.72 (2.15)&9.33 (3.85)&9.42 (2.34)\\
 & \multirow{1}{*}{PCG [Stability Check]}
& 9.11 (2.14)&10.50 (4.36)& 9.70 (2.55)\\ 
 & \multirow{1}{*}{PCG [Validation Check]}
& 9.10 (2.17)&10.50 (4.36)& 9.75 (2.59)\\ 
 & \multirow{1}{*}{PCG [Mixed Check]}
& 9.09 (2.17)&10.50 (4.36)& 9.70 (2.55)\\  
\hline\\[-3.1mm]
\multirow{4}{*}{COIL20} 
 &Newton&10.54 (2.03)&9.79 (4.94)&11.32 (2.19)\\ 
 & \multirow{1}{*}{PCG [Stability Check]}
&12.42 (2.68)&10.63 (4.66)&12.92 (2.14)\\ 
 & \multirow{1}{*}{PCG [Validation Check]}
&13.07 (2.73)&12.08 (4.75)&13.52 (2.12)\\ 
 & \multirow{1}{*}{PCG [Mixed Check]}
&12.43 (2.69)&10.42 (4.63)&12.87 (2.20)\\ 
\hline\\[-3.1mm]
\multirow{4}{*}{USPST} 
 &Newton&14.98 (2.88)&15 (3.57)&15.38 (3.55)\\ 
 & \multirow{1}{*}{PCG [Stability Check]}
&15.60 (3.45)&15.67 (3.60)&16.11 (3.95)\\ 
 & \multirow{1}{*}{PCG [Validation Check]}
&15.40 (3.38)&15.67 (3.98)&15.94 (4.04)\\ 
 & \multirow{1}{*}{PCG [Mixed Check]}
&15.45 (3.53)&15.50 (3.92)&15.94 (4.08)\\ 
\hline\\[-3.1mm]
\multirow{4}{*}{MNIST3VS8} 
 &Newton&2.2 (0.14)&1.67 (1.44)&2.02 (0.22)\\ 
 & \multirow{1}{*}{PCG [Stability Check]}
&2.11 (0.06)&1.67 (1.44)&1.93 (0.2)\\ 
& \multirow{1}{*}{PCG [Validation Check]}
&2.11 (0.06)&1.67 (1.44)&1.93 (0.2)\\  
 & \multirow{1}{*}{PCG [Mixed Check]}
&2.11 (0.06)&1.67 (1.44)&1.93 (0.2)\\    
\hline\\[-3.1mm]
\multirow{3}{*}{FACEMIT} 
 & \multirow{1}{*}{PCG [Stability Check]}
& 29.97 (2.51)& 36 (3.46)& 27.97 (5.38)\\
& \multirow{1}{*}{PCG [Validation Check]}
& 29.97 (2.51)& 36 (3.46)& 27.97 (5.38)\\
 & \multirow{1}{*}{PCG [Mixed Check]}
& 29.97 (2.51)& 36 (3.46)& 27.97 (5.38)\\
\hline
\end{tabular}
\caption{Average classification error (standard deviation is reported brackets) of Laplacian SVMs trained in the primal by means of Newton's method (Newton) and of preconditioned conjugate gradient (PCG) with the proposed early stopping conditions (in square brackets).
$\mathcal{U}$ is the set of unlabeled examples used to train the classifiers. $\mathcal{V}$ is the labeled set for cross--validating parameters whereas $\mathcal{T}$ is the out--of--sample test set. Results on the labeled training set $\mathcal{L}$ are omitted since all algorithms correctly classify such a few labeled training points.}
\label{tabrespcg2}
\end{table}

\section{Conclusions and future work}
\label{sec:concl}
In this paper we described investigated in detail two strategies for solving the optimization problem of Laplacian Support Vector Machines (LapSVMs) in the primal. A very fast solution can be achieved using preconditioned conjugate gradient coupled with an early stopping criterion based on the stability of the classifier decision. Detailed experimental results on real world data show the validity of such strategy. The computational cost for solving the problem reduces from $O(n^3)$ to $O(n^2)$, where $n$ is the total number of training points, both labeled and unlabeled, without the need of storing in memory the Hessian matrix and its inverse. Training times are significantly reduced on all selected benchmarks, in particular, as the amount of training data increases. This solution can be a useful starting point for applying greedy techniques for incremental classifier building or for studying the effects of a sparser kernel expansion of the classification function, that we will address in future work.


\newpage

\appendix
\section*{Appendix A.}
\label{appp}
This Appendix collects all the parameters selected using our experimental protocol, for reproducibility of the experiments (\tablename~\ref{tabparam} and \tablename~\ref{tab:parampcg}). Details of the cross--validation procedure are described in \sectionname~\ref{sec:results}.

In the most of the datasets, parameter values selected using the PCG solution remain substantially the same of the ones selected by solving the primal problem with the Newton's method, suggesting that a reliable and fast cross--validation can be performed with PCG and the proposed early stopping heuristics. In the USPST(B), COIL20(B), and MNIST3VS8 datasets, larger values for $\ga$ or $\gi$ are selected when using PCG, since the convergence speed of gradient descent is enhanced.

To emphasize this behavior, the training times and the resulting error rates of the PCG solution computed using $\ga$ and $\gi$ tuned by means of the Newton's method (instead of the ones computed by PCG with each specific goal condition) are reported in \tablename~\ref{tabtimepcg2a} and in \tablename~\ref{tabrespcg2a}. Comparing these results with the ones presented in \sectionname~\ref{sec:results}, it can be appreciated that both the convergence speed (\tablename~\ref{tabtimepcg1}) and the accuracy of the PCG solution (\tablename~\ref{tabrespcg2}) benefit from an appropriate parameter selection.

\begin{table}
\footnotesize
\centering
\begin{tabular}{llccccc}
\hline\\[-3.8mm]
\textbf{Dataset}&\textbf{Classifier}&\textbf{$\sigma$}&\textbf{$nn$}&\textbf{$p$}&\textbf{$\ga$}&\textbf{$\gi$}\\
\hline\\[-3.4mm]
\multirow{5}{*}{\ G50C} &\ SVM&17.5&-&-&$10^{-1}$&-\\
 &\ RLSC&17.5&-&-&1&-\\
 &\ LapRLSC&17.5&50&5&$10^{-6}$&$10^{-2}$\\ 
 &\ LapSVM Dual (Original)&17.5&50&5&1&10\\
 &\ LapSVM Primal (Newton)&17.5&50&5&$10^{-1}$&10\\
\hline\\[-3.1mm]
\multirow{5}{*}{\ COIL20(B)} &\ SVM&0.6&-&-&$10^{-6}$&-\\
 &\ RLSC&0.6&-&-&$10^{-6}$&-\\
 &\ LapRLSC&0.6&2&1&$10^{-6}$&1\\ 
 &\ LapSVM Dual (Original)&0.6&2&1&$10^{-2}$&100\\
 &\ LapSVM Primal (Newton)&0.6&2&1&$10^{-6}$&1\\ 
 \hline\\[-3.1mm]
 \multirow{5}{*}{\ PCMAC} &\ SVM&2.7&-&-&$10^{-6}$&-\\
 &\ RLSC&2.7&-&-&$10^{-6}$&-\\
 &\ LapRLSC&2.7&50&5&$10^{-6}$&$10^{-2}$\\ 
 &\ LapSVM Dual (Original)&2.7&50&5&$10^{-6}$&$10^{-4}$\\
 &\ LapSVM Primal (Newton)&2.7&50&5&$10^{-6}$&1\\ 
\hline\\[-3.1mm]
\multirow{5}{*}{\ USPST(B)} &\ SVM&9.4&-&-&$10^{-6}$&-\\
 &\ RLSC&9.4&-&-&$10^{-1}$&-\\
 &\ LapRLSC&9.4&10&2&$10^{-4}$&$10^{-1}$\\ 
 &\ LapSVM Dual (Original)&9.4&10&2&$10^{-6}$&$10^{-2}$\\
 &\ LapSVM Primal (Newton)&9.4&10&2&$10^{-6}$&$10^{-2}$\\  
\hline\\[-3.1mm]
\multirow{5}{*}{\ COIL20} &\ SVM&0.6&-&-&$10^{-6}$&-\\
 &\ RLSC&0.6&-&-&$10^{-6}$&-\\
 &\ LapRLSC&0.6&2&1&$10^{-6}$&1\\ 
 &\ LapSVM Dual (Original)&0.6&2&1&$10^{-6}$&10\\
 &\ LapSVM Primal (Newton)&0.6&2&1&$10^{-6}$&1\\ 
\hline\\[-3.1mm]
\multirow{5}{*}{\ USPST} &\ SVM&9.4&-&-&$10^{-1}$&-\\
 &\ RLSC&9.4&-&-&$10^{-6}$&-\\
 &\ LapRLSC&9.4&10&2&$10^{-6}$&$10^{-1}$\\ 
 &\ LapSVM Dual (Original)&9.4&10&2&$10^{-6}$&$10^{-2}$\\
 &\ LapSVM Primal (Newton)&9.4&10&2&$10^{-4}$&1\\ 
\hline\\[-3.1mm]
\multirow{5}{*}{\ MNIST3VS8} &\ SVM&9&-&-&$10^{-6}$&-\\
 &\ RLSC&9&-&-&$10^{-6}$&-\\
 &\ LapRLSC&9&20&3&$10^{-6}$&$10^{-2}$\\ 
 &\ LapSVM Dual (Original)&9&20&3&$10^{-6}$&$10^{-2}$\\
 &\ LapSVM Primal (Newton)&9&20&3&$10^{-6}$&$10^{-2}$\\ 
 \hline\\[-3.1mm]
 \multirow{3}{*}{\ FACEMIT} &\ SVM&4.3&-&-&$10^{-6}$&-\\
 &\ RLSC&4.3&-&-&$10^{-6}$&-\\
 &\ LapSVM Primal (PCG)&4.3&6&1&$10^{-6}$&$10^{-8}$\\ 
\hline
\end{tabular}
\caption{Parameters selected by cross--validation for supervised algorithms (SVM, RLSC) and semi--supervised ones based on manifold regularization, using different loss functions (LapRLSC, LapSVM trained in the dual formulation and in the primal one by means of Newton's method). The parameter $\sigma$ is the bandwidth of the Gaussian kernel or, in the MNIST3VS8, the degree of the polynomial one.}
\label{tabparam}
\end{table}

\begin{table}
\footnotesize
\centering
\begin{tabular}{llcc}
\hline\\[-3.8mm]
\textbf{Dataset}&\textbf{Laplacian SVM}&\textbf{$\ga$}&\textbf{$\gi$}\\
\hline\\[-3.4mm]
\multirow{4}{*}{\ G50C} 
 &Newton&$10^{-1}$&10\\
 &PCG [Stability Check]&$10^{-1}$&10\\ 
 &PCG [Validation Check]&$10^{-1}$&10\\ 
 &PCG [Mixed Check]&$10^{-1}$&10\\ 
\hline\\[-3.1mm]
\multirow{4}{*}{\ COIL20(B)}
 &Newton&$10^{-6}$&1\\ 
 &PCG [Stability Check]&$10^{-6}$&1\\
 &PCG [Validation Check]&1&100\\
 &PCG [Mixed Check]&$10^{-6}$&1\\
\hline\\[-3.1mm]
\multirow{4}{*}{\ PCMAC}
 &Newton&$10^{-6}$&1\\ 
 &PCG [Stability Check]&$10^{-4}$&1\\
 &PCG [Validation Check]&$10^{-4}$&1\\
 &PCG [Mixed Check]&$10^{-6}$&$10^{-1}$\\
\hline\\[-3.1mm]
\multirow{4}{*}{\ USPST(B)}
 &Newton&$10^{-6}$&$10^{-2}$\\  
 &PCG [Stability Check]&$10^{-6}$&1\\ 
 &PCG [Validation Check]&$10^{-6}$&1\\ 
 &PCG [Mixed Check]&$10^{-6}$&1\\ 
\hline\\[-3.1mm]
\multirow{4}{*}{\ COIL20}
 &Newton&$10^{-6}$&1\\ 
 &PCG [Stability Check]&$10^{-6}$&1\\ 
 &PCG [Validation Check]&$10^{-6}$&1\\ 
 &PCG [Mixed Check]&$10^{-6}$&1\\
\hline\\[-3.1mm]
\multirow{4}{*}{\ USPST} 
 &Newton&$10^{-4}$&1\\ 
 &PCG [Stability Check]&$10^{-4}$&1\\ 
 &PCG [Validation Check]&$10^{-4}$&1\\ 
 &PCG [Mixed Check]&$10^{-4}$&1\\ 
\hline\\[-3.1mm] 
\multirow{4}{*}{\ MNIST3VS8} 
 &Newton&$10^{-6}$&$10^{-2}$\\ 
 &PCG [Stability Check]&$10^{-6}$&$10^{-1}$\\ 
 &PCG [Validation Check]&$10^{-6}$&$10^{-1}$\\ 
 &PCG [Mixed Check]&$10^{-6}$&$10^{-1}$\\  
\hline\\[-3.1mm]
\multirow{3}{*}{\ FACEMIT} 
 &PCG [Stability Check]&$10^{-6}$&$10^{-8}$\\ 
 &PCG [Validation Check]&$10^{-6}$&$10^{-8}$\\ 
 &PCG [Mixed Check]&$10^{-6}$&$10^{-8}$\\ 
\hline
\end{tabular}
\caption{A comparison of the parameters selected by cross--validation for Laplacian SVMs trained in the primal by means of Newton's method (Newton) and preconditioned conjugate gradient (PCG) with the proposed early stopping conditions (in square brackets).}
\label{tab:parampcg}
\end{table}

\begin{table}
\footnotesize
\centering
\begin{tabular}{lllll}
\hline\\[-3.8mm]
\textbf{Dataset}&\textbf{Laplacian SVM}&\textbf{Training Time}&\textbf{PCG Iters}&\textbf{LS Iters}\\
\hline\\[-3.4mm]
\multirow{5}{*}{G50C} 
 &Dual&0.155 (0.004)&-&-\\
 &Newton&0.134 (0.006)&-&-\\ 
 & \multirow{1}{*}{PCG [Stability Check]}
&0.044 (0.006)&20 (0)& 1 (0)\\ 
 & \multirow{1}{*}{PCG [Validation Check]}
&0.043 (0.006)&20.83 (2.89)& 1 (0)\\ 
 & \multirow{1}{*}{PCG [Mixed Check]}
&0.044 (0.006)&20.83 (2.89)& 1 (0)\\ 
\hline\\[-3.1mm]
\multirow{5}{*}{COIL20(B)}
 &Dual&0.311 (0.012)&-&-\\
 &Newton&0.367 (0.097)&-&-\\
 & \multirow{1}{*}{PCG [Stability Check]}
&0.198 (0.074)&74.67 (28.4)& 2.41 (1.83)\\ 

& \multirow{1}{*}{PCG [Validation Check]}
&0.095 (0.018)&36 (7.24)& 3.26 (2.21)\\ 
 & \multirow{1}{*}{PCG [Mixed Check]}
&0.206 (0.089)&78.67 (34.42)& 2.38 (1.79)\\ 
\hline\\[-3.1mm]
\multirow{5}{*}{PCMAC} 
 &Dual&14.8203 (0.104)&-&-\\
 &Newton&15.756 (0.285)&-&-\\ 
 & \multirow{1}{*}{PCG [Stability Check]}
&1.901 (0.022)&38.00 (0)& 1.18 (0.45)\\ 
 & \multirow{1}{*}{PCG [Validation Check]}
&1.970 (0.265)&39.58 (5.48)& 1.18 (0.44)\\ 
 & \multirow{1}{*}{PCG [Mixed Check]}
&1.969 (0.268)&39.58 (5.48)& 1.18 (0.44)\\ 
\hline\\[-3.1mm]
\multirow{5}{*}{USPST(B)} 
 &Dual&1.196 (0.015)&-&-\\
 &Newton&1.4727 (0.2033)&-&-\\
 & \multirow{1}{*}{PCG [Stability Check]}
&0.496 (0.172)&95.00 (33.40)& 6.56 (3.18)\\ 
 & \multirow{1}{*}{PCG [Validation Check]}
&0.279 (0.096)&52.25 (18.34)& 6.83 (3.44)\\ 
 & \multirow{1}{*}{PCG [Mixed Check]}
&0.567 (0.226)&107.67 (43.88)& 6.49 (3.15)\\ 
\hline\\[-3.1mm]
\multirow{5}{*}{COIL20} 
 &Dual&6.321 (0.441)&-&-\\
 &Newton&7.26 (1.921)&-&-\\
 & \multirow{1}{*}{PCG [Stability Check]}
&3.297 (1.471)&65.47 (30.35)& 2.53 (1.90)\\ 
 & \multirow{1}{*}{PCG [Validation Check]}
&1.769 (0.299)&34.07 (6.12)& 3.37 (2.22)\\ 
 & \multirow{1}{*}{PCG [Mixed Check]}
&3.487 (1.734)&69.53 (35.86)& 2.48 (1.87)\\ 
\hline\\[-3.1mm]
\multirow{5}{*}{USPST} 
 &Dual&12.25 (0.2)&-&-\\
 &Newton&17.74 (2.44)&-&-\\   
 & \multirow{1}{*}{PCG [Stability Check]}
&1.953 (0.403)&41.17 (8.65)& 3.11 (1.73)\\ 
 & \multirow{1}{*}{PCG [Validation Check]}
&2.032 (0.434)&42.91 (9.38)& 3.13 (1.73)\\ 
 & \multirow{1}{*}{PCG [Mixed Check]}
&2.158 (0.535)&45.60 (11.66)& 3.12 (1.72)\\ 
\hline\\[-3.1mm]
\multirow{5}{*}{MNIST3VS8} 
 &Dual&2064.18 (3.1)&-&-\\
 &Newton&2824.174 (105.07)&-&-\\   
 & \multirow{1}{*}{PCG [Stability Check]}
&188.775 (0.237)&165 (0)& 6.78 (3.65)\\
 & \multirow{1}{*}{PCG [Validation Check]}
&207.986 (35.330)&183.33 (31.75)& 6.65 (3.57)\\
 & \multirow{1}{*}{PCG [Mixed Check]}
&207.915 (35.438)&183.33 (31.75)& 6.65 (3.57)\\
\hline\\[-3.1mm]
\multirow{3}{*}{FACEMIT}  
 & \multirow{1}{*}{PCG [Stability Check]}
&35.728 (0.868)&3 (0)& 1 (0)\\
 & \multirow{1}{*}{PCG [Validation Check]}
&35.728 (0.868)&3 (0)& 1 (0)\\
 & \multirow{1}{*}{PCG [Mixed Check]}
&35.728 (0.868)&3 (0)& 1 (0)\\
\hline
\end{tabular}
\caption{Training time comparison among the Laplacian SVMs trained in the dual (Dual), LapSVM trained in the primal by means of Newton's method (Newton) and by means of preconditioned conjugate gradient (PCG) with the proposed early stopping conditions (in square brackets). \textit{Parameters of the classifiers were tuned using the Newton's method}. Average training times (in seconds) and their standard deviations, the number of PCG iterations, and of Line Search (LS) iterations (per each PCG one) are reported.}
\label{tabtimepcg2a}
\end{table}

\begin{table}
\footnotesize
\centering
\begin{tabular}{lllll}
\hline\\[-3.8mm]
\textbf{Dataset}&\textbf{Laplacian SVM}&\textbf{$\mathcal{U}$}&\textbf{$\mathcal{V}$}&\textbf{$\mathcal{T}$}\\
\hline\\[-3.4mm]
\multirow{4}{*}{G50C} 
 &Newton&6.16 (1.48)&6.17 (3.46)&7.27 (2.87)\\ 
 & \multirow{1}{*}{PCG [Stability Check]}
& 6.13 (1.46)& 6.17 (3.46)& 7.27 (2.87)\\ 
 & \multirow{1}{*}{PCG [Validation Check]}
& 6.16 (1.48)& 6.17 (3.46)& 7.27 (2.87)\\ 
 & \multirow{1}{*}{PCG [Mixed Check]}
& 6.16 (1.48)& 6.17 (3.46)& 7.27 (2.87)\\ 
\hline\\[-3.1mm]
\multirow{4}{*}{COIL20(B)}
 &Newton&8.16 (2.04)&7.92 (3.96)&8.56 (1.9)\\ 
 & \multirow{1}{*}{PCG [Stability Check]}
& 8.81 (2.23)& 8.13 (3.71)& 8.84 (1.93)\\ 
& \multirow{1}{*}{PCG [Validation Check]}
& 8.97 (2.32)& 9.17 (3.74)& 8.96 (1.64)\\ 
 & \multirow{1}{*}{PCG [Mixed Check]}
& 8.84 (2.28)& 8.13 (3.71)& 8.84 (1.96)\\ 
\hline\\[-3.1mm]
\multirow{4}{*}{PCMAC} 
 &Newton&9.68 (0.77)&7.83 (4.04)&9.37 (1.51)\\ 
 & \multirow{1}{*}{PCG [Stability Check]}
& 9.65 (0.76)& 7.83 (4.04)& 9.42 (1.43)\\ 
 & \multirow{1}{*}{PCG [Validation Check]}
& 9.65 (0.76)& 7.83 (4.04)& 9.40 (1.43)\\ 
 & \multirow{1}{*}{PCG [Mixed Check]}
& 9.65 (0.76)& 7.83 (4.04)& 9.40 (1.43)\\ 
\hline\\[-3.1mm]
\multirow{4}{*}{USPST(B)} 
 &Newton&8.72 (2.15)&9.33 (3.85)&9.42 (2.34)\\
 & \multirow{1}{*}{PCG [Stability Check]}
&11.07 (2.27)&13.33 (4.21)&11.49 (2.55)\\ 
 & \multirow{1}{*}{PCG [Validation Check]}
&12.02 (2.22)&14.67 (2.99)&12.01 (2.14)\\ 
 & \multirow{1}{*}{PCG [Mixed Check]}
&10.81 (2.39)&12.83 (4.78)&11.31 (2.71)\\ 
\hline\\[-3.1mm]
\multirow{4}{*}{COIL20} 
 &Newton&10.54 (2.03)&9.79 (4.94)&11.32 (2.19)\\ 
 & \multirow{1}{*}{PCG [Stability Check]}
&12.42 (2.68)&10.63 (4.66)&12.92 (2.14)\\ 
 & \multirow{1}{*}{PCG [Validation Check]}
&13.07 (2.73)&12.08 (4.75)&13.52 (2.12)\\ 
 & \multirow{1}{*}{PCG [Mixed Check]}
&12.43 (2.69)&10.42 (4.63)&12.87 (2.20)\\ 
\hline\\[-3.1mm]
\multirow{4}{*}{USPST} 
 &Newton&14.98 (2.88)&15 (3.57)&15.38 (3.55)\\ 
 & \multirow{1}{*}{PCG [Stability Check]}
&15.60 (3.45)&15.67 (3.60)&16.11 (3.95)\\ 
 & \multirow{1}{*}{PCG [Validation Check]}
&15.40 (3.38)&15.67 (3.98)&15.94 (4.04)\\ 
 & \multirow{1}{*}{PCG [Mixed Check]}
&15.45 (3.53)&15.50 (3.92)&15.94 (4.08)\\ 
\hline\\[-3.1mm]
\multirow{4}{*}{MNIST3VS8} 
 &Newton&2.2 (0.14)&1.67 (1.44)&2.02 (0.22)\\ 
 & \multirow{1}{*}{PCG [Stability Check]}
& 3.16 (0.15)& 2.5 (1.25)& 2.4 (0.38)\\
& \multirow{1}{*}{PCG [Validation Check]}
& 2.89 (0.62)& 2.50 (1.25)& 2.37 (0.44)\\
 & \multirow{1}{*}{PCG [Mixed Check]}
& 2.89 (0.62)& 2.5 (1.25)& 2.37 (0.44)\\
\hline\\[-3.1mm]
\multirow{3}{*}{FACEMIT} 
 & \multirow{1}{*}{PCG [Stability Check]}
& 29.97 (2.51)& 36 (3.46)& 27.97 (5.38)\\
& \multirow{1}{*}{PCG [Validation Check]}
& 29.97 (2.51)& 36 (3.46)& 27.97 (5.38)\\
 & \multirow{1}{*}{PCG [Mixed Check]}
& 29.97 (2.51)& 36 (3.46)& 27.97 (5.38)\\
\hline
\end{tabular}
\caption{Average classification error (standard deviation is reported brackets) of Laplacian SVMs trained in the primal by means of Newton's method and of preconditioned conjugate gradient (PCG) with the proposed early stopping conditions (in square brackets). \textit{Parameters of the classifiers were tuned using the Newton's method}. $\mathcal{U}$ is the set of unlabeled examples used to train the classifiers. $\mathcal{V}$ is the labeled set for cross--validating parameters whereas $\mathcal{T}$ is the out--of--sample test set. Results on the labeled training set $\mathcal{L}$ are omitted since all classifiers perfectly fit such few labeled training points.}
\label{tabrespcg2a}
\end{table}

\end{document}